%% file: arxiv_v2/arxiv.tex
\newif\ifisTR

\isTRtrue

\documentclass{article} 

\usepackage{fullpage}

\usepackage[T1]{fontenc}
\usepackage{microtype}
\usepackage{graphicx}
\usepackage{booktabs} 
\usepackage{placeins}
\usepackage{authblk}
\usepackage{subcaption}
\usepackage{tcolorbox}
\usepackage{nicefrac}  
\usepackage{enumitem}
\usepackage{hyperref}
\usepackage{multicol}
\usepackage{multirow}
\usepackage{colortbl}
\usepackage{xspace}
\usepackage{wrapfig}
\input{math_commands.tex}

\usepackage{amsmath}
\usepackage{amssymb}
\usepackage{mathtools}
\usepackage[export]{adjustbox}
\usepackage{amsthm}
\usepackage{xcolor}
\usepackage{makecell}
\usepackage{amsmath}
\usepackage{tikz}

\usepackage[capitalize,noabbrev]{cleveref}

\crefformat{equation}{(#2#1#3)}
\Crefformat{equation}{(#2#1#3)}
\crefformat{subequation}{(#2#1#3)}
\Crefformat{subequation}{(#2#1#3)}

\makeatletter

\let\algorithmicinput\relax
\let\algorithmicoutput\relax

\makeatother

\usepackage{algorithm}
\usepackage{algpseudocode}

\algnewcommand\algorithmicinput{\textbf{Input:}}
\algnewcommand\Input{\item[\algorithmicinput]}
\algnewcommand\algorithmicoutput{\textbf{Output:}}
\algnewcommand\Output{\item[\algorithmicoutput]}
\algnewcommand\algorithmicparam{\textbf{Parameters:}}
\algnewcommand\Param{\item[\algorithmicparam]}
\algnewcommand\algorithmicconfig{\textbf{Configurations:}}
\algnewcommand\Config{\item[\algorithmicconfig]}

\theoremstyle{plain}
\newtheorem{theorem}{Theorem}[section]
\newtheorem{proposition}[theorem]{Proposition}

\theoremstyle{definition}

\theoremstyle{remark}

\definecolor{mygray}{gray}{0.85}
\definecolor{LightBlue}{cmyk}{0.06, 0.03, 0.01, 0.0}

\usepackage[textsize=tiny]{todonotes}

\usepackage[round,semicolon,sort]{natbib}

\newcommand{\objective}{$\rho_{\log}$\xspace}
\newcommand{\framework}{\textsc{AutoSpec}\xspace}

\renewcommand{\cite}[1]{\citep{#1}}

\newif\ifshowcomments
\showcommentsfalse

\ifshowcomments
\newcommand {\michael}[1]{{\color{red}\sf{[Michael: #1]}}}
\newcommand {\addressedmichael}[1]{{\color{blue}\sf{[(Addressed) Michael: #1]}}}
\newcommand {\yaoqing}[1]{{\color{teal}\sf{[Yaoqing: #1]}}}
\newcommand {\addressedyaoqing}[1]{{\color{cyan}\sf{[(Addressed) Yaoqing: #1]}}}
\newcommand{\zihang}[1]{{\color{purple}\sf{[Zihang: #1]}}}

\else

\newcommand {\michael}[1]{}
\newcommand {\addressedmichael}[1]{}
\newcommand {\yaoqing}[1]{}
\newcommand {\addressedyaoqing}[1]{}
\newcommand{\zihang}[1]{}

\fi

\begin{document}

\title{Learning to Discover Iterative Spectral Algorithms}

\author{
  Zihang Liu$^{*1,2}$,
  Oleg Balabanov$^{*1,2,4}$,
  Yaoqing Yang$^{3}$, and
  Michael W. Mahoney$^{1,2,4}$\\
  $^1$International Computer Science Institute\\
  $^2$University of California, Berkeley\\
  $^3$Dartmouth College\\
  $^4$Lawrence Berkeley National Laboratory\\
  \vspace{3mm}
  \normalsize{
  \texttt{\{zihang.liu, obalaban\}@berkeley.edu}\\
  \texttt{yaoqing.yang@dartmouth.edu}\\
  \texttt{mmahoney@stat.berkeley.edu}
  }
  \vspace{-10mm}
}

\date{}
\maketitle

\begingroup
\renewcommand{\thefootnote}{*}
\footnotetext{These authors contributed equally.}
\endgroup

\input{arxiv_v2/sections/abstract}

\input{arxiv_v2/sections/introduction}
\input{arxiv_v2/sections/related_works}

\input{arxiv_v2/sections/problem_setup}

\input{arxiv_v2/sections/methodology_arxiv}
\input{arxiv_v2/sections/experiments_arxiv}

\input{arxiv_v2/sections/discussions}
\input{arxiv_v2/sections/conclusions}

\vspace{5mm}
\paragraph{Acknowledgements.}
We would like to acknowledge the NSF and the DARPA DIAL and DARPA AIQ programs for partial support of this work.

\clearpage
\bibliographystyle{plainnat}
\bibliography{example_paper}

\newpage
\appendix
\onecolumn
\input{arxiv_v2/sections/appendix}

\end{document}

%% file: math_commands.tex
\usepackage{amsmath,amsfonts,bm}

\def\eqref#1{equation~(\ref{#1})}

\def\1{\bm{1}}

\DeclareMathAlphabet{\mathsfit}{\encodingdefault}{\sfdefault}{m}{sl}
\SetMathAlphabet{\mathsfit}{bold}{\encodingdefault}{\sfdefault}{bx}{n}



%% file: arxiv_v2/sections/abstract.tex
\begin{abstract}
We introduce \framework, a neural network framework for discovering iterative spectral algorithms for large-scale numerical linear algebra and numerical optimization. 
Our self-supervised models adapt to input operators using coarse spectral information (e.g., eigenvalue estimates and residual norms), and predict recurrence coefficients for computing or applying a matrix polynomial tailored to a downstream task.
The effectiveness of \framework relies on three ingredients: an architecture whose inference pass implements short, executable numerical linear algebra recurrences; efficient training on small synthetic problems with transfer to large-scale real-world operators; and task-defined objectives that enforce the desired approximation or preconditioning behavior across the range of spectral profiles represented in the training set.  
We apply \framework to discovering algorithms for representative tasks on spd matrices: accelerating matrix function approximation; accelerating sparse linear solvers; and spectral filtering/preconditioning for eigenvalue computations. 
On real-world matrices, the learned procedures deliver up to order-of-magnitude improvements in accuracy and/or reductions in iteration count, relative to spectrum-agnostic baselines. We find clear connections to classical theory: the induced polynomials may exhibit equioscillation behavior characteristic of Chebyshev polynomial approximation. The code is available at: \href{https://github.com/zihanghliu/AutoSpec}{https://github.com/zihanghliu/AutoSpec}.

\end{abstract}

%% file: arxiv_v2/sections/introduction.tex
\section{Introduction}
Machine learning (ML) has shown growing potential to scale scientific discovery by automating parts of the algorithm design pipeline~\cite{mankowitz2023alphadev,fawzi_discovering_2022,real2020automlzero,ellis2021dreamcoder,prism_26_tr}. 
However, many successful discovery frameworks search over discrete hypothesis classes (e.g., programs or pseudocode), which does not transfer cleanly to numerical linear algebra (NLA) and numerical optimization, where effective methods are typically continuous objects that must adapt to operator-specific (often spectral)~structure.

In this paper, we propose \framework, a neural network framework for discovering a broad class of iterative spectral methods for NLA and numerical optimization. Our models learn mappings from spectral probes---e.g., a small set of coarse eigenvalue approximations with associated residual norms---to iteration coefficients that define an operator polynomial (or, more generally, an operator function) tailored to a downstream task. At inference time, given a new operator, we (i) extract a probe using a short warm-start eigensolver run, (ii) feed this probe to the trained model to predict iteration coefficients, and (iii) execute the resulting iterative process. This ML workflow mirrors standard NLA practice, where inexpensive spectral information enables methods to adaptively exploit and shape the operator’s spectrum.

Our algorithm discovery approach is fully self-supervised: we train end-to-end by minimizing NLA task-specific objectives, using only task performance as supervision. As a result, the learned methods are optimized to outperform fixed baselines through task feedback, rather than by imitating existing algorithms. Our design offers three advantages: (i) a unified parameterization spanning classical iterative spectral algorithmic structures, enabling discovery across multiple NLA tasks; (ii) training on small synthetic spectral instances with transfer to large-scale operators, producing a compact and deployable learning engine; and (iii) reliance only on coarse spectral information at inference time, improving applicability when spectral access is limited or computational budgets are tight.

 Moreover, we observe clear connections between the discovered algorithms and classical polynomial approximation theory for matrix functions. For instance, some learned polynomials exhibit near-minimax, equiripple behavior, consistent with the Chebyshev equioscillation theorem for minimax approximation. At the same time, our approach is not limited to approximating a prescribed function: it can optimize task-level objectives, such as residual decay or conditioning of a preconditioned operator and the preconditioner. This flexibility can be beneficial for more complex NLA tasks.

We apply \framework to discovering matrix-free iterative algorithms\footnote{Using an operator only through its action on vectors (matvecs).} for several tasks on positive-definite matrices: solving linear systems; computing eigenvalues; and approximating matrix-function actions. 
Across a suite of real-world sparse matrices, the learned methods significantly improve accuracy and/or reduce iteration counts, in some cases by orders of magnitude relative to basic baseline algorithms that do not exploit spectral information. We summarize our main contributions as follows:

\begin{itemize}[topsep=2pt, itemsep=1pt, parsep=0pt, partopsep=0pt]

    \item We propose \framework, a general framework for discovering iterative spectral algorithms with neural networks. Unlike symbolic algorithm discovery methods, our approach learns continuous algorithmic updates suited to NLA and related numerical tasks.

    
    \item \framework is demonstrated to discover nontrivial recurrence structures already when using a simple neural architecture and without supervision. We identify connections to classical approximation theory by noticing that some discovered recurrences exhibit behavior consistent with Chebyshev equioscillation theorem.

    \item We apply \framework to three classical NLA tasks on symmetric positive-definite matrices and show
    that it learns to exploit spectral information from the input, yielding up to order-of-magnitude gains in accuracy and/or iteration count.
\end{itemize}

%% file: arxiv_v2/sections/related_works.tex
\section{Related Works}

\begin{figure*}[htb!]
    \centering
    \begin{subfigure}[t]{0.5\linewidth}
      \centering
      \includegraphics[width=0.75\linewidth]{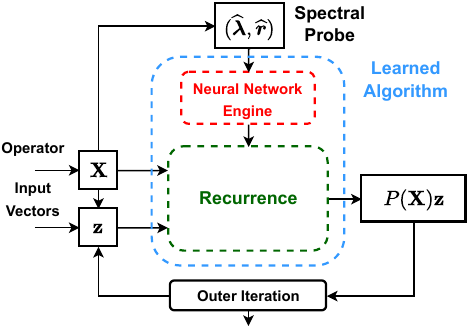}
      \caption{Inference in NLA tasks.}
    \end{subfigure}%
    \begin{subfigure}[t]{0.5\linewidth}
      \centering
      \includegraphics[width=0.9\linewidth]{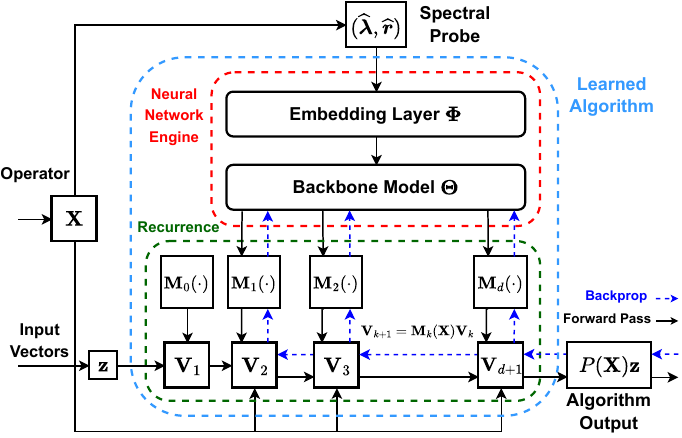}
      \caption{Algorithm structure.}
    \end{subfigure}
    \caption{\textsc{AutoSpec} framework.
    (a) A trained neural network engine, paired with an executable recurrence, defines a discovered numerical algorithm for downstream NLA tasks.
    (b) An end-to-end differentiable algorithmic structure: given operator $\mathbf{X}$, a spectral probe $(\widehat{\boldsymbol{\lambda}}, \widehat{\boldsymbol{r}})$ consisting of eigenvalue estimates and residual norms, is extracted and fed to the neural engine to produce the coefficients of a degree-$d$ polynomial $P(\cdot)$. The polynomial is implemented in a matrix-free manner via a short recurrence: starting from $\mathbf{V}_0 = \mathbf{z}$, we iterate $\mathbf{V}_{k+1} = \mathbf{M}_k(\mathbf{X}) \mathbf{V}_k$, where each $\mathbf{M}_k$ is parameterized by the engine. The terminal state returns the action $\mathbf{V}_{d+1} = P(\mathbf{X}) \mathbf{z}$ of the polynomial to inputs $\mathbf{z}$, which allows self-supervised training by backpropagating task-defined NLA losses on $\mathbf{V}_{d+1}$.}
    \label{fig:framework_teaser} 
    \vspace{-5mm}
\end{figure*}

\paragraph{Algorithm Discovery with ML.}

ML has become a key driver of scientific discovery by enabling automated algorithm design. This problem is naturally framed as a search over a high-dimensional combinatorial space of candidate operations. Many works adopt discrete program spaces, where symbolic regression remains a dominant paradigm, spanning classical evolutionary methods~\citep{koza_genetic_1994, doi:10.1126/science.1165893} and neural-guided approaches~\citep{petersen2019deep, hayes2025deep, kim2020integration, mundhenk2021symbolic, doi:10.1126/sciadv.aay2631, lample2019deep}. Moreover, by formulating the discovery process as a sequential decision-making task, reinforcement learning (RL) becomes a useful method for discovering more complex algorithms~\citep{fawzi_discovering_2022, mankowitz2023alphadev, oh2020discovering, oh_discovering_2025}. More recently, researchers have used the reasoning capabilities of large language models (LLMs) to discover novel algorithms and solve advanced math problems~\citep{romera-paredes_mathematical_2024, novikov2025alphaevolve, chervonyi2025gold}. In this work, we propose a framework for discovering algorithms that can be used in NLA and numerical optimization. For optimization, recent works have used deep unfolding~\citep{takabe2022convergence}, meta-learning~\citep{andrychowicz2016learning, ravi2017optimization, wichrowska2017learned, li2016learning}, and RL~\citep{bello2017neural, chen2023symbolic} to discover optimization algorithms.  For NLA, neural networks have been used to approximate the action of a matrix inverse on vectors, with the learned maps then applied as preconditioners within linear system solvers~\citep{li2023learning, hausner2023neural, lerer2024multigrid,kaneda2023deep, luo2024neural,trifonov2026learning, rudikov2024neural}. These methods are not directly comparable to \framework as they typically train a separate model for each input matrix, rather than discovering a transferable algorithmic procedure.  Notably, \framework could be applied as a second-level accelerator for the resulting preconditioned system solves.  

\paragraph{Spectral Approximation and Preconditioning Methods in NLA.} 
Polynomial accelerations and approximations are among the earliest and most widely used NLA tools for iterative solution of large, sparse linear systems and related tasks. A common strategy is to build low-degree polynomial transforms guided by coarse spectral bounds, that damp slowly converging error components and accelerate convergence~\citep{young1971,varga2000mia,axelsson1994iter,saad2003iter}. Krylov subspace methods provide a complementary paradigm and can be viewed as constructing analogous polynomial filters implicitly and adaptively, by optimizing residuals over progressively enriched subspaces~\citep{hestenesstiefel1952cg,saadschultz1986gmres,greenbaum1997,liesenstrakos2013,saad2011eig}. On modern architectures, Krylov performance is increasingly constrained by communication/synchronization (e.g., global reductions), renewing interest in polynomial preconditioners and filtering~\citep{saad2011eig,zhou2007chebdavidson,banerjee2016chefsidg,bergamaschi2021parallel}. Extensive work develops polynomial preconditioners and smoothers via truncated series, minimax, or least-squares designs, as matrix-free surrogates for inverses or spectrum-shaping operators, used both standalone and within multilevel solvers~\citep{johnson1983poly,ashby1991minimax,ashby1992ls,benzi2002survey,briggs2000multigrid,trottenberg2000multigrid}. Beyond first-order methods, second-order matrix iterations such as Newton methods yield rapidly convergent approximations for inverses, inverse roots, and matrix sign functions, with kernels dominated by matrix-matrix products that map well to modern parallel hardware~\citep{higham2008fm}. These loops now appear in large-scale optimizers and training systems (e.g., Shampoo; Muon; and variants such as PolarExpress)~\citep{gupta2018shampoo,jordan2024muon,amsel2025polarexpress,grishina2025cans,ahn2025dion}. Related iterative matrix (inverse) square-root primitives also underpin second-order vision layers and whitening/decorrelated normalization~\citep{song2021approximate,song2023fast,li2017universal,huang2018decorrelated}. However, these second-order iterations remain far less applicable in sparse and matrix-free regimes, where matvec-based kernels are essential. Finally, when shifted solves (or approximate shift-and-invert actions) are available, rational approximation extends these designs via rational filters, rational Krylov/subspace methods, and adaptive rational fitting such as AAA~\citep{higham2008fm,druskin1998extended,guettel2013rationalkrylov,nakatsukasa2018aaa}.

%% file: arxiv_v2/sections/problem_setup.tex
\section{\textsc{AutoSpec} Framework}
\label{sec:learning_framework}
In this section, we introduce the \textsc{AutoSpec} framework: an end-to-end differentiable approach built around a unified representation of iterative spectral algorithms, with modular components that can be learned by neural networks.

\paragraph{Framework Setup.} For any diagonalizable matrix $\mathbf{X} \in \mathbb{R}^{n \times n}$, we define a \textit{spectral probe} $(\widehat{\boldsymbol{\lambda}}(\mathbf{X}), \widehat{\boldsymbol{r}}(\mathbf{X}))$, consisting of a small set of approximate eigenvalues $\widehat{\boldsymbol{\lambda}}$ from a task-relevant portion of the spectrum (this could be the top eigenvalues, e.g., if one is interested in low-rank approximation, but for many applications it is not), optionally augmented with auxiliary quantities such as residual norms $\widehat{\boldsymbol{r}}$. This mirrors common NLA workflows, where inexpensive spectral estimates (e.g., Ritz values from short Lanczos or Arnoldi runs) are used to guide the construction of preconditioners and approximations.
In particular, our models in~\cref{sec:experiments} 
can take $(\widehat{\boldsymbol{\lambda}},\widehat{\boldsymbol{r}})$
to consist of a few extreme eigenvalue estimates and their associated residuals, obtained via a cheap eigensolver warm-start.

Then, we train a \textit{neural network engine},
\begin{equation} \label{eq:nn_engine}
(\widehat{\boldsymbol{\lambda}},\widehat{\boldsymbol{r}}) \;\mapsto\; P(\cdot),
\end{equation}
that maps an input spectral probe to an operator polynomial (or more generally, an operator function) designed for downstream tasks.  The designated objective may be classical, such as minimizing a matrix function approximation error $\|P(\mathbf{X}) - f(\mathbf{X})\|$, or non-standard, such as e.g. minimizing both the condition number of the preconditioned operator $P(\mathbf{X})\mathbf{X}$, which governs basic linear solver convergence, and the condition number of the preconditioner $P(\mathbf{X})$, which promotes robustness of Krylov solvers.

We parametrize the polynomials $P(\cdot)$ to admit an executable short recurrence that applies the matrix polynomial $P(\mathbf{X})$ efficiently to input vectors and matrices. In this setting, the neural engine predicts the recurrence coefficients and thereby can be realized to govern the convergence (and numerical stability) of the iterative process.

Here is the important point: the combination of a neural network engine and an executable recurrence defines a learned algorithm. This algorithm can then be deployed for NLA downstream tasks, such as preconditioning linear solver iterations.
See~\cref{fig:framework_teaser} for an illustration.

For robust training and for efficient evaluation at inference time, we can represent $P(\mathbf{X})$
implicitly via a state recurrence:
\begin{equation}\label{eq:general_update}
\mathbf{V}_{k+1} = M_k(\mathbf{V}_{k},\mathbf{X}),\qquad 0 \leq k \leq d,
\end{equation}
The initialization for matrix actions is $\mathbf{V}_0=\mathbf{z}$ so that
$\mathbf{V}_{d+1}=P(\mathbf{X})\mathbf{z}$; when $P(\mathbf{X})$ is needed to be formed explicitly (e.g., in small-scale training
or diagnostics) we take $\mathbf{V}_0=\mathbf{I}$, yielding $\mathbf{V}_{d+1}=P(\mathbf{X})$.
The operators $M_k(\cdot,\cdot)$ are specified modularly to realize short NLA-style recurrences.

\paragraph{Modular linear transitions.}
We primarily focus on transitions that are linear in the state,
\begin{equation}\label{eq:modular_lin_update}
M_k(\mathbf{V}_{k},\mathbf{X}) := \mathbf{M}_k(\mathbf{X})\,\mathbf{V}_{k},
\end{equation}
where $\mathbf{M}_k(\mathbf{X})$ is assembled from blocks
$\mathbf{M}_k^{(i,j)} \in \mathbb{R}^{n\times n}$, each given by a low-degree matrix polynomial
$M_k^{(i,j)}(\mathbf{X})$. The final transition $\mathbf{M}_d(\mathbf{X})$ is constrained to have a single block-row,
so the terminal state extracts the desired polynomial output.
With this parameterization, learning the neural engine~\cref{eq:nn_engine} reduces to learning maps
$(\widehat{\boldsymbol{\lambda}},\widehat{\boldsymbol{r}})\mapsto M_k^{(i,j)}(\cdot)$ that specify the block polynomials.
The same modular construction also supports richer transition classes; see Appendix~\ref{app:nonlinear_higher_order}
for extensions to higher-order (state-polynomial) and rational~recurrences.

\paragraph{Matrix-free regime.}
Our main target is the large-scale sparse setting, where constructing or storing $P(\mathbf{X})$ is infeasible.
Accordingly, we seek recurrences that compute the action $\mathbf{z}\mapsto P(\mathbf{X})\mathbf{z}$ using only a
small number of applications of $\mathbf{X}$. Under~\cref{eq:modular_lin_update}, the action is obtained by composing
the transitions,
\[
P(\mathbf{X})\mathbf{z}
= \mathbf{M}_d(\mathbf{X})\,\mathbf{M}_{d-1}(\mathbf{X})\hdots \mathbf{M}_0(\mathbf{X})\,\mathbf{z},
\]
without ever forming $P(\mathbf{X})$ (or any dense $n\times n$ block), provided that applying each block polynomial
$M_k^{(i,j)}(\mathbf{X})$ is implemented via a short sequence of $\mathbf{X}$ matvecs. To control per-step cost and improve robustness in sparse, matrix-free regimes, we can further restrict the transition matrices to use at most one
application of $\mathbf{X}$.

For many downstream tasks such as preconditioning, the polynomial action is only required up to a scale factor. The sequential state-transition representation allows normalization of intermediate states without affecting the result up to scaling, enabling control of numerical growth and improving stability during training and inference (see~\cref{sec:training}).

\subsection{State Transition View}
\label{sec:state_space_view}

In our empirical evaluation, we chose linear transitions 
with the following block parameterization: 
\begin{equation*}
\mathbf{M}_0 =
\begin{bmatrix}
\mathbf{I} &
\mathbf{I} &
\mathbf{X}^\mathrm{T}
\end{bmatrix}^\mathrm{T}~~~ \mathbf{M}_d =
\begin{bmatrix}
\gamma_d \mathbf{I} & \eta_d \mathbf{I} & \alpha_d \mathbf{I} + \beta_d \mathbf{X}
\end{bmatrix}
\end{equation*}
\begin{equation}\label{eq:three_term_recurrence}
\begin{split}
\mathbf{M}_k &=
\begin{bmatrix}
\mathbf{I} & \mathbf{0} & \rho_k \mathbf{I}\\
\mathbf{0} & \mathbf{0} & \mathbf{I}\\
\gamma_k \mathbf{I} & \eta_k \mathbf{I} & \alpha_k \mathbf{I} + \beta_k \mathbf{X}
\end{bmatrix}\qquad 1\le k<d .
\end{split}
\end{equation}
Under~\cref{eq:three_term_recurrence}, the neural engine outputs scalar coefficient sequences
$\{\rho_k,\gamma_k,\eta_k,\alpha_k,\beta_k\}$.

To interpret the blocks in~\cref{eq:three_term_recurrence}, write the state as three $n\times n$ blocks $\mathbf{V}_k \;=\; \begin{bmatrix}\mathbf{A}_k &\mathbf{B}_k &\mathbf{C}_k\end{bmatrix}^\mathrm{T}.$
Then for $1\le k<d$, the update $\mathbf{V}_{k+1}=\mathbf{M}_k\mathbf{V}_k$ is equivalent to (assuming that $\mathbf{X}$ is symmetric)
\begin{equation}\label{eq:block_updates}
\begin{split}
\mathbf{A}_{k+1} &= \mathbf{A}_k + \rho_k\,\mathbf{C}_k,\\
\mathbf{B}_{k+1} &= \mathbf{C}_k, \\
\mathbf{C}_{k+1} &= \gamma_k \mathbf{A}_k + \eta_k \mathbf{B}_k + (\alpha_k\mathbf{I}+\beta_k\mathbf{X})\,\mathbf{C}_k.
\end{split}
\end{equation}
Thus, $\mathbf{B}_k$ is a one-step delay storing the previous polynomial state, $\mathbf{C}_k$ stores the current polynomial state, and $\mathbf{A}_k$ is an accumulator that enables learned linear combinations of intermediate states.

The propositions below isolate two useful special cases:
(i) a direct affine three-term polynomial recurrence; and
(ii) basis generation followed by a learned expansion.
Together, they motivate~\cref{eq:three_term_recurrence} as a compact hypothesis class that subsumes many classical polynomial approximation templates, appropriate for use in an ML workflow. 
In this view, learning
$(\widehat{\boldsymbol{\lambda}},\widehat{\boldsymbol{r}})\mapsto P(\cdot)$ amounts to \emph{selecting and tuning}
a member of this classical design space from a spectral probe, rather than committing a priori to a fixed polynomial family
or coefficient~rule.

\paragraph{Scenario I: Affine Three-term Polynomial Recurrence.}
Set $\rho_k=0$, so $\mathbf{A}_k\equiv \mathbf{I}$. Then $\mathbf{C}_k$ follows an affine three-term recursion
with constant injection through $\gamma_k\mathbf{I}$.
\begin{proposition}[Affine three-term polynomial recurrence]\label{prop:affine_three_term}
Consider~\cref{eq:three_term_recurrence}. If $\rho_k=0$ for all $1\le k\le d$, then $\mathbf{C}_k$ satisfies
\cref{eq:affine_three_term} with $\mathbf{C}_0=\mathbf{I}$ and $\mathbf{C}_1=\mathbf{X}$, and for $k\ge 1$,
\begin{equation}\label{eq:affine_three_term}
\mathbf{C}_{k+1}
= \gamma_k \mathbf{I} + \eta_k \mathbf{C}_{k-1}+ \alpha_k \mathbf{C}_k + \beta_k \mathbf{X} \mathbf{C}_k.
\end{equation}
\end{proposition}
This regime covers classical polynomial iterations defined by orthogonal polynomial recurrences (e.g., Chebyshev on a
spectral interval)~\cite{young1971,manteuffel1977tchebychev,varga2000mia,saad2003iter}, Clenshaw evaluation of expansions in three-term polynomial bases~\cite{clenshaw1955summation}, and matrix-free polynomial
filters for eigenvalue computations~\cite{saad2011eig,zhou2007chebdavidson,banerjee2016chefsidg}.

\paragraph{Scenario II: Basis Generation Plus Learned Expansion.}
This specialization separates \emph{basis generation} from \emph{coefficient selection}. Setting $\gamma_k=0$ for $k<d$
removes the affine injection in~\cref{eq:three_term_recurrence}, so $\mathbf{C}_k$ is generated by a homogeneous three-term
recurrence. The accumulator $\mathbf{A}_k$ then forms a learned linear combination of these basis polynomials via $\rho_k$.
\begin{proposition}[Expansion in a learned three-term basis]\label{prop:basis_expansion}
Consider~\cref{eq:three_term_recurrence}. Assume $\gamma_k=0$ for all $1\le k<d$ and assume the readout uses
$\eta_d=\alpha_d=\beta_d=0$. Then $\mathbf{C}_k$ satisfies \cref{eq:H_three_term} with $\mathbf{C}_0=\mathbf{I}$,
$\mathbf{C}_1=\mathbf{X}$, and for $1\le k<d$,
\begin{equation}\label{eq:H_three_term}
\mathbf{C}_{k+1} = \eta_k \mathbf{C}_{k-1} + \alpha_k \mathbf{C}_k + \beta_k \mathbf{X} \mathbf{C}_k,
\end{equation}
and the output has the expansion form $\mathbf{C}_d
=\gamma_d\mathbf{I} + \gamma_d\sum_{k=1}^{d-1} \rho_k \mathbf{C}_k$
\end{proposition}

This regime represents a common polynomial approximation template in scientific computing: generate a numerically stable polynomial basis through a three-term recurrence (e.g., Chebyshev or Legendre polynomials), and then choose expansion weights to approximate a target function on a prescribed spectral domain using minimax or least-squares criteria~\cite{varga2000mia,saad2003iter,ashby1991minimax,ashby1992ls,dunham1982choice}. \textsc{AutoSpec} learns both the recurrence defining the basis and the expansion weights from the spectral probe, and can optimize objectives beyond direct function fitting, such as downstream residual reduction, eigengaps or conditioning.

\subsection{Learning Objective}
We propose a general learning objective for discovering different NLA algorithms with \textsc{AutoSpec}. 
For each NLA task, we adopt the metric $r$ representing convergence rate of the iteration, and we evaluate the ratio between the learned algorithm and commonly used (spectrum-agnostic) baseline methods on a log scale. 
We denote our learning objective as \objective:
\begin{equation}
\label{eq:objective}
    \text{\objective} = \frac{{\log r_{\text{model}}}}{\log r_{\text{baseline}}} .
\end{equation}
We choose the ratio in logarithmic scale to alleviate bias towards certain spectral profiles when performing training/evaluation on a large, diverse set of matrices, enforcing desired preconditioning behaviors uniformly on all $\mathbf{X}$ in the dataset. Details on the metric $r$ and baseline methods for different NLA tasks are provided in Appendix~\ref{appendix:task_objectives}.

%% file: arxiv_v2/sections/methodology_arxiv.tex
\section{Neural Network Engine of \framework}
\subsection{Model Structure}\label{sec:model_structure}
In this section, we introduce the neural network engine of \framework. As indicated in Figure~\ref{fig:framework_teaser}, the neural network engine learns to construct a set of transition operators $\mathbf{M}_1, \dots, \mathbf{M}_d$ for each target matrix $\mathbf{X}$. It consists of an embedding layer that extracts the relevant spectral features from coarse spectral information, and a backbone model that maps the features to transition operators. 




\paragraph{Backbone Model (Pre-training).} The backbone model mirrors the iterative update formula~\cref{eq:modular_lin_update} using an unfolded modular structure, where each module represents a state transition operator $\mathbf{M}_k$, parameterized by coefficients $\{\rho_k, \gamma_k, \eta_k, \alpha_k, \beta_k\}$ learned by a neural network layer given spectral features $\mathbf{e}$ as input. Subsequently, the transition matrices $\mathbf{M}_k$ formulate the recurrence $\mathbf{V}_{k+1} = \mathbf{M}_k\mathbf{V}_k$, that results in the algorithm output $P(\mathbf{X}) = \mathbf{V}_{d+1}$. Since the generated recurrence is applied directly to training data, the backbone model is end-to-end trainable. The detailed backbone design is shown in Algorithm~\ref{alg:train_layer} in Appendix~\ref{apprndix:model_structure_detail}. 


\noindent 
\paragraph{Embedding Layer (Post-training).} 
The embedding layer $g_{\phi}$ takes as input a coarse spectral probe, obtained from a cheap eigensolver warm-start. It learns a mapping from the spectral probe to an embedding vector, $(\widehat{\boldsymbol{\lambda}}, \widehat{\boldsymbol{r}}) \mapsto \mathbf{e}$, encoding the essential spectral features. Details of embedding layer is shown in Algorithm~\ref{alg:embed_layer_eigenproblem} in Appendix~\ref{apprndix:model_structure_detail}

During training, we first pre-train the backbone model with a synthetic dataset, using prescribed spectral features as input. For post-training, we fix the backbone and train the embedding layer using spectral probes of the dataset as input. We fix the model during inference time on realistic problems.

\subsection{Training and Evaluation Strategy}
\label{sec:training}
Here, we provide neural network training and evaluation strategies. Additional training details are provided in Appendices~\ref{appendix:training_settings} and~\ref{appendix:training_details}.

\paragraph{Training and Evaluation Data. }
Our training procedure does not follow a conventional in-distribution learning setup based on sampling spectra from a fixed probability distribution. 
Instead, we train on small synthetic operators specifically designed to exhibit spectral properties (e.g., slow versus fast versus heavy-tailed decay of dominant eigenvalues) representative of given application regimes. Both pre-training and post-training stages use synthetic spectrums. At test time, we apply the trained model on realistic matrices whose spectral profiles are unknown but share similar structural properties to the training data. Thus, our evaluation measures generalization across operator instances within a structurally defined spectral class, rather than i.i.d. distributional generalization.
By conditioning solely on dimension-agnostic spectral summaries rather than size-dependent features, the model learns a size-independent mapping from spectral input to polynomial coefficients. This enables extrapolation from training operators with $n \le \mathcal{O}(10^3)$ to real-world matrices with dimensions up to $\mathcal{O}(10^6)$, provided they share similar spectral structure. This setup reflects the intended use case in NLA tasks, where algorithms are designed to generalize across broad operator classes characterized by spectral properties rather than fixed input distributions.
Details of the training data generation procedure are provided in Appendix~\ref{sec:training_data}.

\paragraph{Efficient Training on Diagonal Matrices. }
\label{sec:train_diagonal_matrices} 
Training on full matrices can incur significant computational overhead. Observe that if two diagonalizable matrices $\mathbf{X}$ and $\mathbf{Y}$ share the same spectrum, then $P(\mathbf{X})$ and $P(\mathbf{Y})$ also share the same spectrum. Furthermore, our objective depends only on spectral quantities. Consequently, without loss of generality, we may restrict training to diagonal matrices of the form $\mathbf{X} = \operatorname{diag}(\boldsymbol{\lambda})$.
This representation significantly improves training efficiency by avoiding operations on full matrices, especially when realized on fully synthetic data.

\paragraph{Stable Training with Scale-invariant Update. }
Learning higher degree algorithms poses the risk of numerical overflows for the iterative update when training the neural network. We note that objectives in NLA tasks are often scale-invariant, such as maximizing eigenvalue gaps or minimizing condition numbers. Under the state-space view of the learned recurrence in Section~\ref{sec:state_space_view}, we can see that scaling the states  does not change the recurrence relation $\mathbf{V}_{k+1}=\mathbf{M}_k\mathbf{V}_k$, but it prevents the recurrence from growing exponentially with polynomial degree. Therefore, in training, we can normalize the state by its norm after each recurrence update. This effectively controls the norm of the states and prevents numerical overflows.

\paragraph{Optimization Objective.} For all NLA tasks, we use a unified template for optimization objective of the model. Based on the learning objective~\cref{eq:objective}, we propose a three-term unsupervised optimization loss $\mathcal{L}$ for training the neural network engine, that optimizes \objective for different spectral tasks:

\vspace{-18pt}
\begin{equation}\label{eq:general_loss_function}
    \begin{split}
        \mathcal{L} = \frac{1}{N}\sum_{i=1}^{N}\left[ c_1\mathcal{L}_{\text{obj}}(\mathbf{X}_i) + c_2\mathcal{L}_{\text{struct}}(\mathbf{X}_i) + c_3\mathcal{L}_{\text{reg}}(\mathbf{X}_i)\right] 
    \end{split} 
\end{equation}
\vspace{-10pt}

where $N$ is the size of dataset $\{\mathbf{X}_i\}_{i=1}^{N}$. In~\cref{eq:general_loss_function}: 
$\mathcal{L}_{\text{obj}}$ is the inverse of the learning objective \objective, i.e., $\mathcal{L}_{\text{obj}} = \frac{1}{\rho_{\text{log}}} = \frac{{\log r_{\text{baseline}}}}{\log r_{\text{model}}}$, which enforces desired spectral properties uniformly across the dataset; 
$\mathcal{L}_{\text{struct}}$ is a structural objective that imposes structural constraints on the learned algorithm (e.g., the condition number of the preconditioner); and 
$\mathcal{L}_{\text{reg}}$ is a regularization term that penalizes outliers and promotes learning more generalizable algorithms. 
More details on the loss function can be found in Appendix~\ref{appendix:optimization_objectives}. 

%% file: arxiv_v2/sections/experiments_arxiv.tex
\begin{figure}[!htb]
    \centering
    \begin{subfigure}{0.6\linewidth}
        \includegraphics[width=\textwidth]{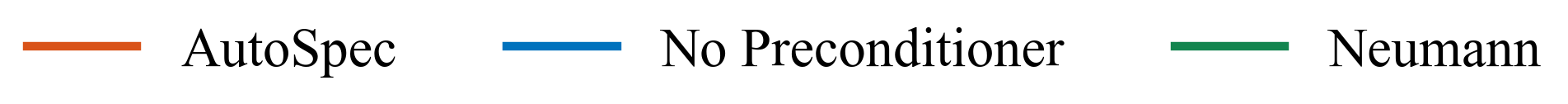}
    \end{subfigure}

    \vspace{-1mm}
    
    \begin{subfigure}[t]{0.28\linewidth}
        \includegraphics[width=\textwidth]{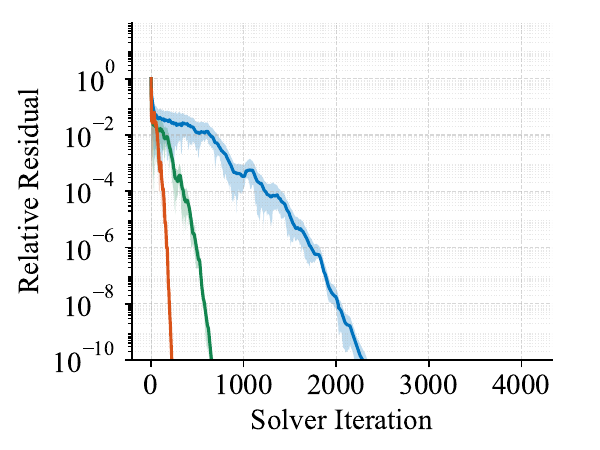}
        \caption{\texttt{G2\_circuit}}
    \end{subfigure}
    \hspace{-0.4cm}
    \begin{subfigure}[t]{0.251\linewidth}
        \includegraphics[width=\textwidth]{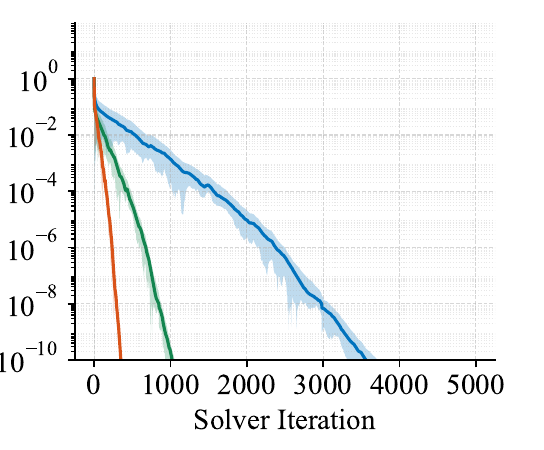}
        \caption{\texttt{parabolic\_fem}}
    \end{subfigure}
    \hspace{-0.4cm}
    \begin{subfigure}[t]{0.251\linewidth}
        \includegraphics[width=\textwidth]{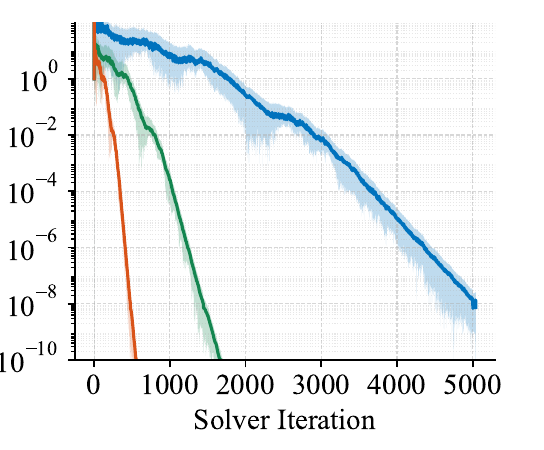}
        \caption{\texttt{Shipsec5}}
    \end{subfigure}
    \hspace{-0.4cm}
    \begin{subfigure}[t]{0.251\linewidth}
        \includegraphics[width=\textwidth]{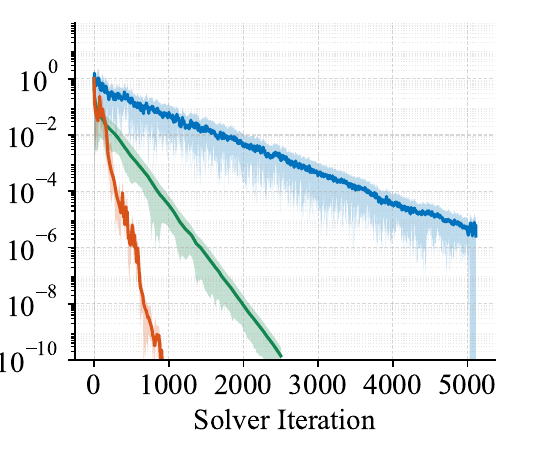}
        \caption{\texttt{Fault\_639}}
    \end{subfigure}

    \caption{Convergence of CG solver with polynomial preconditioning. Curves show the median error and the 20th/80th percentiles over five random seeds. }
    \label{fig:main_comparison_linsolve} 
    \vspace{-5mm}
\end{figure}

\section{Empirical Results}
\label{sec:experiments}
In this section, we describe the empirical performance of the \framework framework.
We trained neural network engines for linear systems, eigenvalue problems, and matrix inverse square roots, and we evaluated their performance.

    


\subsection{Evaluation Setup}\label{sec:evaluation_setup}

\paragraph{Linear systems and eigenproblems.}
We consider a computational regime in which the dominant cost is determined by the number of outer Krylov iterations, a standard metric for comparing polynomial preconditioners of the same degree~\cite{saad1985practical,barrett1994templates,saad2003iter,saad2011eig,bergamaschi2021parallel}. In this setting, a trained neural engine is deployed as a plug-in and invoked from \textsc{Matlab} to output the coefficient sequence that defines the polynomial updates. Spectral probes are obtained from a short warm start: for eigenproblems, we run 20 iterations of \texttt{eigs} (Krylov--Schur algorithm) using the same search subspace dimension as the main run; for linear systems, we run 50 Lanczos steps without storing the Lanczos basis.

We use a broad set of sparse test matrices drawn from the SuiteSparse Matrix Collection~\citep{DavisHu2011,Kolodziej2019}, which spans electronic-structure, thermal finite-element, structural/mechanics, and circuit/model-reduction problems, with sizes $n\in[468,\,1.6\times 10^6]$ and condition numbers ranging by orders of magnitude. Furthermore, all test matrices are spd. For linear systems, we use polynomial preconditioning as a second-level accelerator on top of inexpensive explicit preconditioners, such as Jacobi, AMG or incomplete Cholesky. In eigenvalue problems, the indefinite operators are implicitly squared.    Full experimental details are given in  Appendix~\ref{sec:realworld_appendix}.

\paragraph{Matrix functions. }
We consider approximation of $\mathbf{X}^{-1/2}$ with a polynomial for covariance whitening of a regularized DNA similarity Gram matrix derived from \textit{E.\ coli} K-12 MG1655 (U00096.3). Using a $9$-mer spectrum embedding with $262{,}144$ features and $250$ sequences, we construct a feature matrix $\mathbf{A}$ and compute inverse square root approximation of $\mathbf{X}=\mathbf{A}\mathbf{A}^\mathrm{T}+\lambda\mathbf{I}.$
The matrix $\mathbf{A}$ is highly sparse (with nonzeros density around $0.375\%$).

\subsection{Preconditioned Linear Systems}
\label{sec:linsolve_results}

We train a neural network engine to generate degree-11 polynomial preconditioners for accelerating Conjugate Gradient (CG) solvers, and evaluate its performance on SuiteSparse systems with random Gaussian right-hand sides. 
Spectral probes are obtained from 50 Lanczos steps\footnote{with cost comparable to $\sim$50 CG iterations.} and used as input to the neural network engine.

\begin{figure}[!htb]
    \centering

    \begin{subfigure}[t]{0.6\linewidth}
        \centering

        \includegraphics[width=0.5\linewidth]{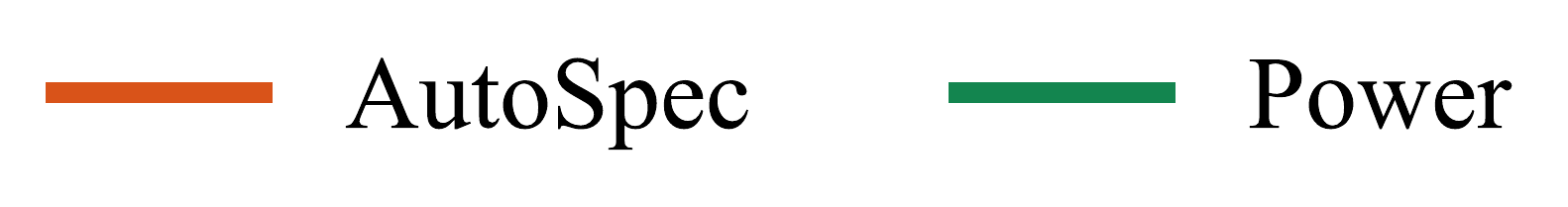}

        \vspace{3pt}

        \begin{minipage}[t]{0.45\linewidth}
            \centering
            \includegraphics[width=\textwidth]{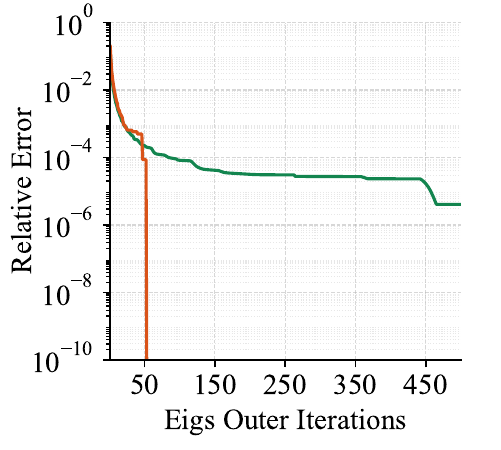}

            \vspace{-2pt}
            {\scriptsize (1a) $\lambda_{10}$ of \texttt{SiO2} ($k=10$)}
        \end{minipage}
        \hfill
        \begin{minipage}[t]{0.45\linewidth}
            \centering
            \includegraphics[width=\textwidth]{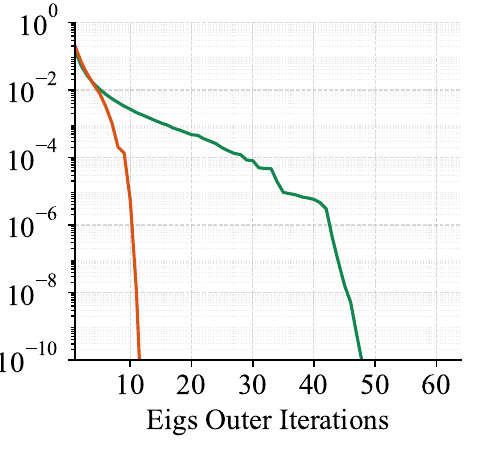}

            \vspace{-2pt}
            {\scriptsize (1b) $\lambda_{20}$ of \texttt{thermal2} ($k=20$)}
        \end{minipage}

        \label{fig:eigs_convergence}
    \end{subfigure}
    \hfill
    \begin{subfigure}[t]{0.34\linewidth}
        \centering

        \vspace{0pt}
        \includegraphics[width=0.9\textwidth]{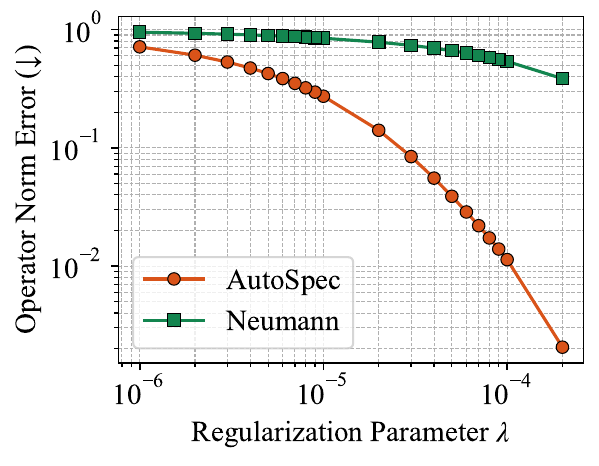}

        \vspace{5pt}
        {\scriptsize (2) Approximation of $\mathbf{X}^{-1/2}$}
        \label{fig:cov_whiten}
    \end{subfigure}

    \caption{
    \textbf{Left:} Convergence of eigenvalue approximations versus \texttt{eigs} outer iterations for (shifted) \texttt{SiO2} and \texttt{thermal2}, for varying target numbers of eignevalues $k$ and subspace dimension $l=4k$. The NN preconditioner is produced using a spectral probe from 20 \texttt{eigs} warm-start iterations. 
    \textbf{Right:} Whitening the covariance matrix of the \textit{E. coli} K-12 MG1655 reference genome sequence.
    }
    \label{fig:eigs_and_cov_whiten}
    \vspace{-6mm}
\end{figure}

Table~\ref{tab:realistic_matrix_results_a} reports the number of CG iterations required to reach a relative residual error of $10^{-10}$, along with estimated condition numbers before and after preconditioning. We observe that the learned preconditioners achieve order-of-magnitude acceleration and substantially outperform spectrum-agnostic Neumann preconditioning across all test cases.  The convergence curves for representative operators are given in Figure~\ref{fig:main_comparison_linsolve}. See Appendix~\ref{sec:realworld_appendix} for extended real-world experiments and Appendix~\ref{appendix:synthetic_results_linsolve} for results on synthetic systems.

\subsection{Eigenvalue Problems}
\label{sec:eigval_problems}

For eigenvalue problems, the downstream task is to compute the $k$ smallest-magnitude eigenpairs using \texttt{eigs} under a fixed search subspace budget $l$ (with user-defined $k$ and $l$~\footnote{Typically $l = c\,k$ for a small constant $c$, e.g., $c = 4$.}). After applying an identity shift to make the target modes dominant, we use a degree-21 polynomial transform generated by the neural network engine to enhance spectral separation near the target boundary, here chosen to be defined by the eigengap between the $k$-th and $1.5k$-th eigenvalues.

Figures~\ref{fig:eigs_and_cov_whiten} and~\ref{fig:eigs_convergence_complementary} (in the appendix) compare \texttt{eigs}\footnote{We request $k + 5$ eigenvalues in \texttt{eigs}, which is standard practice to improve convergence.} with \framework preconditioning versus \texttt{eigs} with a power preconditioner baseline $P(\mathbf{X}) = \mathbf{X}^d$. The \framework preconditioning consistently significantly outperforms the baseline across different $(k, l)$ configurations and matrices, which indicates generalization. Figure~\ref{fig:ablation_zero_residual_eigs} further shows that performance improves with probe quality, suggesting the neural engine adapts to the spectral profiles. See Appendix~\ref{sec:realworld_appendix} for extended real-world matrix experiments and Appendix~\ref{appendix:synthetic_results_eigenproblems} for complementary synthetic results.



\subsection{Approximating Matrix Functions}

We train a neural network to predict a degree-21 polynomial $P(\mathbf{X})$ that approximates the matrix inverse square root, $P(\mathbf{X}) \approx \mathbf{X}^{-1/2}$, and we apply it to the covariance whitening of the regularized DNA
sequence similarity Gram matrix. 
We measure whitening quality via the operator-norm residual $\lVert P(\mathbf{X})\,\mathbf{X}\,P(\mathbf{X})-\mathbf{I}\rVert_2$, and we compare against a truncated Neumann-series baseline. 
As shown in Figure~\ref{fig:eigs_and_cov_whiten}, the learned approximation algorithm achieves order-of-magnitude improvement over Neumann series approximation in the operator norm error.
See Appendix~\ref{appendix:synthetic_results_matfunc} for complementary results on synthetic matrices.

%% file: arxiv_v2/sections/discussions.tex
\begin{figure}[!htb]
    \centering

    \begin{subfigure}{0.45\linewidth}
        \centering
        \includegraphics[width=\textwidth]{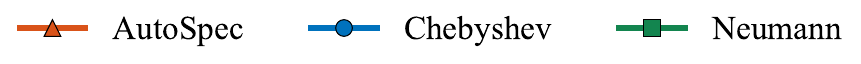}
    \end{subfigure}


    \begin{subfigure}[t]{0.49\linewidth}
        \centering

        \begin{minipage}[t]{0.53\linewidth}
            \centering
            \includegraphics[width=\textwidth]{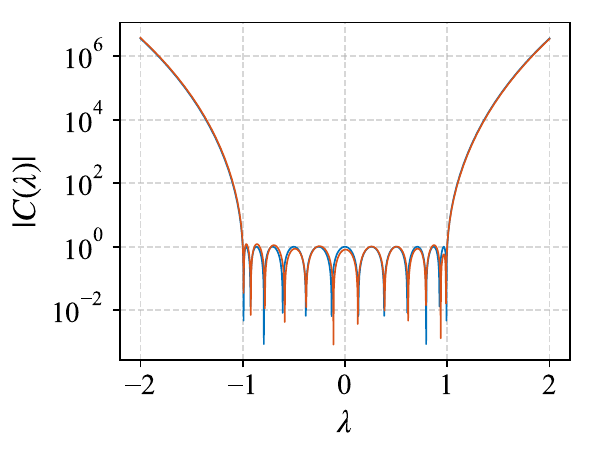}

            \vspace{-2pt}
            {\scriptsize \texttt{G3\_circuit}}
        \end{minipage}
        \hfill
        \begin{minipage}[t]{0.44\linewidth}
            \centering
            \includegraphics[width=\textwidth]{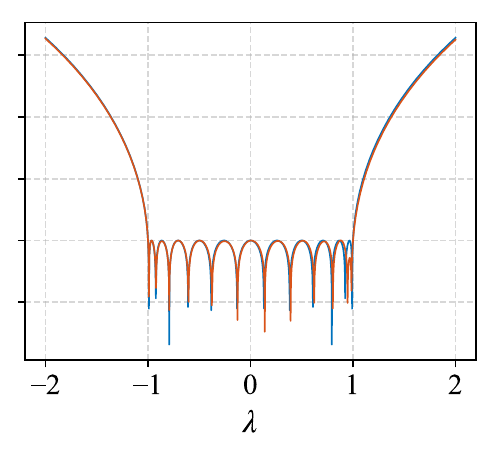}

            \vspace{-2pt}
            {\scriptsize Laplacian}
        \end{minipage}

        \caption{Post-fitted residual polynomials.}
        \label{fig:plot_polynomials_postfit}
    \end{subfigure}
    \hspace{-2mm}
    \begin{subfigure}[t]{0.49\linewidth}
        \centering

        \begin{minipage}[t]{0.53\linewidth}
            \centering
            \includegraphics[width=\textwidth]{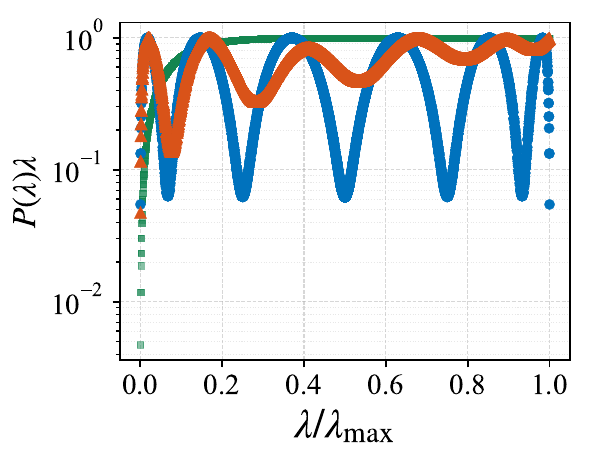}

            \vspace{-2pt}
            {\scriptsize Degree 11}
        \end{minipage}
        \hfill
        \begin{minipage}[t]{0.444\linewidth}
            \centering
            \includegraphics[width=\textwidth]{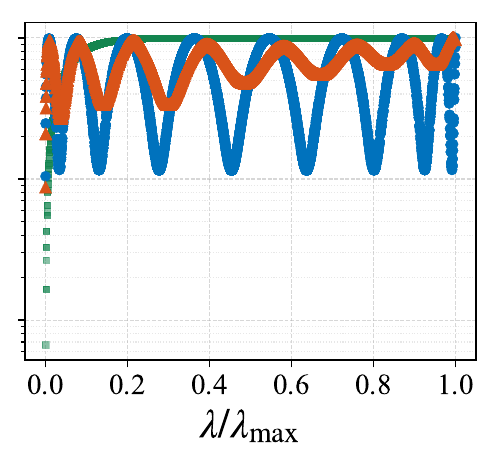}

            \vspace{-2pt}
            {\scriptsize Degree 16}
        \end{minipage}

        \caption{Eigenvalues $P(\lambda)\lambda$ of the preconditioned Laplacian  }
        \label{fig:compare_polynomial}
    \end{subfigure}

    \caption{(a) Residual polynomials $r(\lambda) = 1-\lambda P(\lambda)$ induced by \framework trained without structural objective $\mathcal{L}_{\mathrm{struct}}$ can be post-fitted by an affine shift/scale
    $C(\lambda) = a\,r(b\,\lambda+c)+d$ to resemble Chebyshev polynomials. (b) Incorporating $\mathcal{L}_{\mathrm{struct}}$ allows \framework to depart from standard minimax constructions and generate $P(\cdot)$ avoiding clusters of small eigenvalues in $P(\mathbf{X})\mathbf{X}$.}
    \label{fig:plot_polynomials}
    \vspace{-5mm}
\end{figure}

\section{Analysis and Ablation Studies} \label{sec:ablation}

\paragraph{Connection to Classical Algorithms.}

To position the algorithms discovered by \framework relative to classical methods, we test whether some of the generated polynomials exhibit structure analogous to  Chebyshev polynomials.

Chebyshev polynomials are a widely used baseline for polynomial spectral acceleration in NLA, owing to their minimax and equioscillation properties. We evaluate \emph{minimax} behavior as follows: for a polynomial $P(\cdot)$, we search over affine interval parameters and minimize an \emph{optimality gap}  that compares the observed peak value of $P(\cdot)$ on the candidate interval with the theoretical minimax bound. Optimality gap $\to 0$ indicates that the polynomial closely matches Chebyshev-like behavior. Details are given in Appendix~\ref{appendix:eval_properties}. Table~\ref{tab:eval_cheb_property} shows that on realistic eigenproblems, the learned polynomial achieves a minimax error close to the optimal value, whereas a randomly generated polynomial (``Random'') has an error larger by orders of magnitude. This indicates that the learned polynomial possesses \emph{minimax} properties strongly analogous to those of Chebyshev polynomials. Furthermore, when \framework is trained for linear-system acceleration optimizing without $\mathcal{L}_\mathrm{struct}$ in the loss, the residual polynomial $1-xP(x)$ is expected to exhibit the same minimax/equioscillation behavior after an appropriate shift and scaling. Figure~\ref{fig:plot_polynomials_postfit} confirms this behavior empirically.

\paragraph{Beyond Classical Constructions.}


\framework supports objectives beyond classical function-approximation criteria, as illustrated here for linear systems. In addition to the standard minimax objective $\|\mathbf{I}-P(\mathbf{X})\mathbf{X}\|$, we introduce a structural objective $\mathcal{L}_{\mathrm{struct}}$ motivated by a known limitation of Chebyshev polynomial preconditioners: although they can substantially improve $\mathrm{cond}(P(\mathbf{X})\mathbf{X})$, they may also create clusters of small eigenvalues, which can adversely affect the numerical behavior of the resulting solver and lead to suboptimal convergence. We mitigate this effect by controlling $\mathrm{cond}(P(\mathbf{X}))$ together with $\mathrm{cond}(P(\mathbf{X})\mathbf{X})$, with the goal of keeping $\mathrm{cond}(P(\mathbf{X}))$ close to the natural lower bound $\mathrm{cond}(P(\mathbf{X})) \;\ge\; 
\frac{\mathrm{cond}(\mathbf{X})}{\mathrm{cond}(P(\mathbf{X})\mathbf{X})}.$
This allows \framework to trade a modest reduction in the conditioning improvement of the preconditioned operator for better conditioning of the preconditioner itself, yielding accelerators that are better aligned with the downstream linear-solver task. We note that classical NLA approaches also exist for addressing eigenvalue clustering in Chebyshev polynomial preconditioners~\cite{bergamaschi2021parallel}. Polynomials of degrees 11 and 16 generated by \framework, after training with $\mathcal{L}_{\mathrm{struct}}$, for the FD discretization of the Laplacian on the unit square are shown in Figure~\ref{fig:compare_polynomial}.

\paragraph{Potential to Advancing State-of-the-Art NLA.} 

To demonstrate the potential of the \framework framework for using ML to automate state-of-the-art NLA, we conduct additional experiments comparing discovered preconditioning method for linear systems with classical Chebyshev polynomial preconditioning. The results show that the discovered algorithms operate more reliably under limited spectral information, adapt to the structure of the input operator, and often outperform Chebyshev preconditioners, sometimes by significant margins.  For details, see Appendix~\ref{sec:cheb_comparison} and~\ref{sec:second_level_comparison}.

Ablations on the role of residual features and the robustness of learned algorithms are in Appendix~\ref{sec:appendix_ablation}.

\paragraph{Limitations.} 

The current evaluation focuses on spd problems, where strong classical methods, including Chebyshev preconditioners, already exist. Extending \framework to indefinite and unsymmetric problems is a natural direction for future work. We evaluate \framework using the number of outer iterations, a standard metric for comparing polynomial preconditioners. Reductions in this metric directly translate into runtime speedups over an unpreconditioned solver when memory and communication costs of outer Krylov iterations dominate. Finally, \framework requires a rich synthetic training set whose spectra contain the key features of the matrices encountered at test time. For example, in eigenvalue problems, if the model is trained only on spectra with larger eigengaps than those encountered at test time, performance can degrade, as illustrated in Appendix~\ref{sec:limitations}.

We also note that \framework can support richer update classes beyond the linear recurrences studied in the main text, including
higher-order polynomial recurrences (yielding more expressive bases) and nonlinear transition maps that preserve the
iterative structure while increasing expressivity; see Appendix~\ref{app:nonlinear_higher_order} for details. 
Together, these results point to a broader pathway for using ML methodology to learn robust, deployable NLA and numerical optimization algorithms that generalize across operator instances.

%% file: arxiv_v2/sections/conclusions.tex
\section{Conclusion}
We introduced \framework, a neural network framework for discovering spectrum-adaptive numerical iterative algorithms by training a neural
engine to output the coefficients of an executable recurrence. Across multiple NLA tasks, the resulting learned
recurrences consistently improve convergence and/or accuracy over spectrum-agnostic baselines, demonstrating the effectiveness of the algorithm discovery.

%% file: arxiv_v2/sections/appendix.tex
\clearpage
\section*{Appendix}

\section{Details of the \framework Learning Framework}

\subsection{Extensions} \label{app:nonlinear_higher_order}
We outline extensions of the linear state-transition model used in the \framework framework in main text.

\paragraph{Higher-order updates.} 
Higher-order updates can be realized by allowing nonlinear dependence on the state $\mathbf{V}_k$.
For example, a quadratic transition can be defined as
\begin{equation}\label{eq:quadratic_update}
M_k(\mathbf{V}_k,\mathbf{X})
:= \mathbf{I}_k(\mathbf{X})\,\mathbf{V}_k\,\mathbf{J}_k(\mathbf{X})\,\mathbf{V}_k^{\mathsf{T}}\,\mathbf{K}_k(\mathbf{X}),
\end{equation}
where matrices $\mathbf{I}_k(\mathbf{X}),\mathbf{J}_k(\mathbf{X}),\mathbf{K}_k(\mathbf{X})$ admit block decompositions with
subblocks given by low-degree polynomials in $\mathbf{X}$.
Such higher-order updates are less suited to sparse, matrix-free regimes than linear transitions, since they generally require explicit matrix--matrix products. However, on parallel hardware, especially in optimization loops that already use explicit matrix-function approximations (e.g., inverse-root preconditioning in Shampoo and orthogonalization/polar updates in Muon), they can be highly attractive \cite{higham2008fm,gupta2018shampoo,jordan2024muon,amsel2025polarexpress,grishina2025cans,ahn2025dion}. Discovering and training such higher-order recurrences for optimization is a promising direction for future work.

\paragraph{Rational Models.} 
In some settings, such as approximating matrix functions with sharp variation near the spectrum (e.g., $\mathbf{X}^{-1/2}$ or $\log(\mathbf{X})$), or computing interior eigenvalues, it can be advantageous to model $P(\cdot)$ as a \emph{rational} function rather than a polynomial function. Methods based on rational spectral transformations play a central role in NLA and are particularly prominent in solvers based on Krylov approximation \cite{higham2008fm,druskin1998extended,guettel2013rationalkrylov,nakatsukasa2018aaa}.
Within the \framework framework, rational constructions can be incorporated naturally by allowing the transition blocks $M^{(i,j)}_k(\cdot)$  to be rational functions. For example, in the three-term update formula~\cref{eq:three_term_recurrence}, the linear shifts may be replaced by resolvents,
$$\alpha_k \mathbf{I} + \beta_k \mathbf{X} \rightarrow (\alpha_k \mathbf{I} + \beta_k \mathbf{X})^{-1},~\text{for~~$1 \leq k \leq d$.}   $$
This substitution trades inexpensive matvecs with $\mathbf{X}$ for more costly shifted linear solves during inference, but it can yield substantially improved approximations and preconditioners.
\paragraph{Gradient-driven transitions.}
The same state-transition view can also express first-order optimization recurrences by treating the operator in
\cref{eq:general_update} as a nonlinear map accessed through gradient evaluations, i.e.,
$\mathbf{X}(\mathbf{z})=\nabla f(\mathbf{z})$.\footnote{Here we overload $\mathbf{X}$ to denote an operator/oracle rather
than a matrix.}
For instance, with an augmented state $\mathbf{V}_k=(\boldsymbol{\theta}_k,\mathbf{m}_k)$, the transition recovers gradient descent with
momentum,
\[
\mathbf{m}_{k+1}=\beta_k \mathbf{m}_k+\mathbf{X}(\boldsymbol{\theta}_{k}),\qquad
\boldsymbol{\theta}_{k+1}=\boldsymbol{\theta}_k-\alpha_k \mathbf{m}_{k+1},
\]
and yields Nesterov acceleration by evaluating the gradient at an extrapolated point
(e.g., $\mathbf{X}(\boldsymbol{\theta}_k+\gamma_k\mathbf{m}_k)$) \cite{polyak1964,nesterov1983,nesterov2004}.
This suggests learning \emph{gradient-driven} neural engines that map cheap probe features (e.g., Hessian spectral
estimates, gradient norms/inner products, or curvature sketches) to coefficient sequences $(\alpha_k,\beta_k,\gamma_k,\ldots)$,
thereby discovering accelerated or task-adaptive optimizers within universal recurrence formalism.

\subsection{Learning Objectives for Different Tasks}
\label{appendix:task_objectives}

In~\cref{eq:objective}, we introduced the generalized objective $\rho_{\log}$, which compares a learned method against a
task-specific baseline through a scalar performance metric $r(\mathbf{X})$. Concretely, each application requires the following:
(i) a metric $r$ that quantifies convergence rate/accuracy; and 
(ii) a baseline method (without learned acceleration) used to form
$r_{\text{baseline}}$. 
For the tasks in this paper, we used the following.

\begin{itemize}
\item \textbf{Eigenvalue problems.}
 We measure separation after applying the polynomial spectral transform $P(\cdot)$~via
 \begin{align}\label{eq:obj_eigenproblem}
 r(\mathbf{X})
 = \frac{\min_{i=1,\ldots,k}\,\big|P(\lambda_i(\mathbf{X}))\big|}
 {\max_{I\subset\{1,\ldots,m\},\,|I|=l}\; \min_{j\in I}\,\big|P(\lambda_j(\mathbf{X}))\big|}\,,
 \end{align}
where $\{\lambda_i(\mathbf{X})\}$ denotes the spectrum of $\mathbf{X}$, $k$ is the number
 of requested eigenpairs, and $l$ is the dimension of Ritz approximation subspaces. As a baseline we use standard power/subspace
 iteration polynomial $P(\lambda_j(\mathbf{X})) = \lambda_j(\mathbf{X})^d$.

 \item \textbf{Linear systems.}
We use the operator-norm residual of the normalized preconditioned operator,
\begin{align}\label{eq:obj_linsolve}
 r(\mathbf{X})
 = \left\| \mathbf{I} - \frac{P(\mathbf{X})\mathbf{X}}{\|P(\mathbf{X})\mathbf{X}\|_2} \right\|_2 ,
 \end{align}
 with Richardson iteration polynomial as the baseline.

 \item \textbf{Inverse square root.}
For approximating $\mathbf{X}^{-1/2}$, we use the worst-case relative error in operator norm,
\begin{align}\label{eq:obj_matrix_function}
 r(\mathbf{X})
 = \left\| \mathbf{I} - P(\mathbf{X})\,\mathbf{X}^{1/2} \right\|_2 ,
 \end{align}
with a Neumann (truncated Taylor expansion) polynomial as the baseline.
\end{itemize}

\section{Details of Neural Network Engine}
\label{apprndix:model_structure_detail}
We provide the details of the neural network engine, including the backbone model and embedding layer, and specifications for different NLA tasks.

\subsection{Backbone Model}
Corresponding to Section~\ref{sec:model_structure}, the detailed structure of each layer of the backbone model is shown in Algorithm~\ref{alg:train_layer}. 

\begin{algorithm}
\caption{NN Layer $k$ of Backbone Model $f_{\theta_k}$}
\label{alg:train_layer}
\begin{algorithmic}[1]

\Param Weight matrices $\mathbf{W}_k\in\mathbb{R}^{d\times5}, \mathbf{w}_k\in\mathbb{R}^{d}$, Bias $\mathbf{b}_k\in\mathbb{R}^5, b_k\in\mathbb{R}$, $\epsilon=10^{-8}$
\Input Input to neural network layer $\mathbf{e}\in\mathbb{R}^d$, matrix $\mathbf{X}$.
\Output State transition operators $\mathbf{M}_k(\mathbf{X})$

\vspace{0.1cm}
\State $[\rho_k, \gamma_k, \eta_k, \alpha_k, \beta_k]^{\mathrm{T}} \gets \mathbf{W}_k^{\mathrm{T}}\mathbf{e} + \mathbf{b}_k$ \Comment{Coefficients}
\State $\delta_k \gets \mathbf{w}_k^{\mathrm{T}}\mathbf{e} + b_k$ \Comment{Learned Scaling Factor}
\State $\gamma_k, \eta_k, \alpha_k, \beta_k \gets \frac{\gamma_k}{\delta_k + \epsilon}, \frac{\eta_k}{\delta_k + \epsilon}, \frac{\alpha_k}{\delta_k + \epsilon}, \frac{\beta_k}{\delta_k + \epsilon}$
\vspace{0.15cm}
\If{$1 \le k < d$}
    define $\mathbf{M}_k$ as 
    $$ \big (\mathbf{x},
         \mathbf{y},
        \mathbf{z} \big )
    \mapsto \big (
       \mathbf{x} + \rho_k\mathbf{z} ,\,
       \mathbf{z} , \,
        \gamma_k \mathbf{x} + \eta_k \mathbf{y} +\alpha_k\mathbf{z} + \beta_k \mathbf{X} \mathbf{z} \big ) $$
\ElsIf{$k = d$}
    define $\mathbf{M}_k$ as
     $$\big (
        \mathbf{x},
         \mathbf{y},
        \mathbf{z} \big)
    \mapsto
       \gamma_k \mathbf{x} + \eta_k \mathbf{y} +\alpha_k\mathbf{z} + \beta_k \mathbf{X} \mathbf{z}$$
\EndIf

\vspace{0.1cm} 
\State \Return $\mathbf{M}_{k}$
\end{algorithmic}
\end{algorithm}

\subsection{Embedding Layer}
\label{appendix:embedding_structure}

\noindent\paragraph{Embedding Layer for Eigenvalue Problems. }To compute the largest or smallest $k$ eigenvalues of $\mathbf{X}$ using iterative eigensolvers such as Krylov-Schur with subspace dimension $l$, an effective preconditioning algorithm should enlarge the spectral gap between the $k$-th and $l$-th eigenvalues. Varying $k$ or $l$ changes the target portion of the spectrum, requiring the model to generalize across different target spectral regions. We design an embedding layer that allows the model to adapt to different $k$ and $l$ and construct effective preconditioners, requiring only truncation and padding to the input spectral probes.  

Algorithm~\ref{alg:embed_layer_eigenproblem} shows the structure of the embedding layer for eigenvalue problems. It uses two multi-layer subnetworks to exploit spectral information from the input spectral probe, and it returns the spectral embedding for the $k$-th and $l$-th eigenvalue, respectively. The first subnetwork, parameterized by $\mathbf{W}_1, \mathbf{W}_3$, first obtains $\mathbf{y}_1$ that encodes the spectral information of top-$k$ eigenvalues, and then selects the smallest magnitude $y_1$ as the embedding for the $k$-th eigenvalue. The second subnetwork, parameterized by $\mathbf{W}_2, \mathbf{W}_4$, outputs an embedding $\mathbf{y}_2$ encoding spectral information in the neighborhood of the $l$-th eigenvalue. Then the two embeddings are concatenated to form the final output embedding. The layer has a fixed input size of $2l_0$, corresponding to $l$ approximated eigenvalues and corresponding residuals.  

In practice, the target rank $k'$ and subspace dimension $l'$ may differ from the fixed input window size $(k_0,l_0)$ expected by the embedding layer. 
To produce a fixed-length input of $l_0$ spectral probes (eigenvalue estimates and residuals), we construct a normalized window by truncation/padding while preserving the two boundary indices. 
Specifically, we first form the leading block of $k_0$ probes from the top-$k'$ approximate eigenvalues by truncating or padding as needed, and we enforce that the $k_0$-th probe corresponds exactly to the $k'$-th approximate eigenvalue. 
We then form the remaining $l_0-k_0$ probes from the range $\{\widehat\lambda_{k'},\ldots,\widehat\lambda_{l'}\}$, again truncating or padding to length $l_0-k_0$, and we enforce that the final ($l_0$-th) probe corresponds exactly to the $l'$-th approximate eigenvalue. 
This construction ensures that, for any $(k',l')$, the network input always contains a consistent spectral window anchored at the $k'$-th and $l'$-th eigenvalues.

\begin{algorithm}
\caption{Embedding Layer $g_{\phi}$ for Eigenvalue Problems}
\label{alg:embed_layer_eigenproblem}
\begin{algorithmic}[1] 
\Config Number of eigenvalues to compute $k_0$ with subspace dimension $l_0$, Model input dimension $d_{\text{in}} = 2l_0$, Hidden dimension $d_{\text{hid}} = 4k_0+8$

\Param $\mathbf{W}_1, \mathbf{W}_2\in\mathbb{R}^{d_{\text{in}}\times d_{\text{hid}}}, \mathbf{W}_3\in\mathbb{R}^{d_{\text{hid}}\times k_0}, \mathbf{W}_4\in\mathbb{R}^{d_{\text{hid}}\times 4}$, GeLU activation function $\sigma$

\Input Approximated eigenvalues $\widehat{\boldsymbol{\lambda}} = \left[\widehat{\lambda}_1, \dots, \widehat{\lambda}_{l_0}\right]^{\mathrm{T}}$, Residuals $\widehat{\boldsymbol{r}} = \left[\widehat{r}_1, \dots, \widehat{r}_{l_0}\right]^{\mathrm{T}}$

\vspace{0.1cm}
\Output Embedding $\mathbf{e}=\left[e_1, \dots, e_5\right]^{\mathrm{T}}\in\mathbb{R}^5$
\vspace{0.1cm}

\State $\mathbf{x} \gets \left[\widehat{\boldsymbol{\lambda}}^{\mathrm{T}}, \widehat{\boldsymbol{r}}^{\mathrm{T}}\right]^{\mathrm{T}}$ \Comment{Construct input $\mathbf{x}\in\mathbb{R}^{2l_0}$}
\vspace{0.1cm}
\State $\mathbf{y}_1 \gets \mathbf{W}_3^{\mathrm{T}}\sigma(\mathbf{W}_1^{\mathrm{T}}\mathbf{x})$ \Comment{First subnetwork, $\mathbf{y}_1\in\mathbb{R}^{k_0}$} 
\vspace{0.1cm}
\State $y_1 = \operatorname*{min}_{1 \le i \le n} |(\mathbf{y}_1)_i|$
\vspace{0.1cm}
\State $\mathbf{y}_2 \gets \mathbf{W}_4^{\mathrm{T}}\sigma(\mathbf{W}_3^{\mathrm{T}}\mathbf{x})$ \Comment{Second subnetwork, $\mathbf{y}_2\in\mathbb{R}^4$}
\vspace{0.1cm}
\State $\mathbf{e} \gets \left[y_1, \mathbf{y}_2^{\mathrm{T}}\right]^{\mathrm{T}}$
\vspace{0.2cm} 
\State \Return $\mathbf{e}$

\end{algorithmic}
\end{algorithm}

\noindent\paragraph{Embedding Layer for Preconditioned Linear Systems.}
The learning objective of preconditioning algorithms for linear systems it to reduce the condition number of operators. 
Therefore, the task-relevant portion of the spectrum is the two ends of the spectrum, and the spectral probe consists of the largest and smallest $k$ eigenvalues (with corresponding residuals). 
In our experiments, we choose $k=20$. 
Algorithm~\ref{alg:embed_layer_linsolve} describe the procedure of the embedding layer for preconditioned linear systems. 
We use two subnetworks to process the spectral probe for the top and bottom eigenspectrum separately, and we concatenate their outputs to form the embedding. 
Each subnetwork uses the spectral probe of corresponding spectral regions and exploit the structure of the eigenspectrum.

\begin{algorithm}
\caption{Embedding Layer $g_{\phi}$ for Preconditioned Linear Systsems}\label{alg:embed_layer_linsolve}
\begin{algorithmic}[1] 
\Config Number of largest/smallest input eigenvalue estimations $k$, Model input dimension $d_{\text{in}} = 2k$, Hidden dimension $d_{\text{hid}} = 2k$, Matrix dimension $m$

\Param $\mathbf{W}_1, \mathbf{W}_2\in\mathbb{R}^{d_{\text{in}}\times d_{\text{hid}}}, \mathbf{W}_3, \mathbf{W}_4\in\mathbb{R}^{d_{\text{hid}}\times 5}$, GeLU activation function $\sigma$

\Input Approximated largest eigenvalues $\widehat{\boldsymbol{\lambda}}_{\text{max}} = \left[\widehat{\lambda}_1, \dots, \widehat{\lambda}_{k}\right]^{\mathrm{T}}$, Residuals $\widehat{\boldsymbol{r}}_{\text{max}} = \left[\widehat{r}_1, \dots, \widehat{r}_{k}\right]^{\mathrm{T}}$, Approximated smallest eigenvalues $\widehat{\boldsymbol{\lambda}}_{\text{min}} = \left[\widehat{\lambda}_{m-k+1}, \dots, \widehat{\lambda}_m\right]^{\mathrm{T}}$, Residuals $\widehat{\boldsymbol{r}}_{\text{min}} = \left[\widehat{r}_{m-k+1}, \dots, \widehat{r}_{m}\right]^{\mathrm{T}}$

\vspace{0.1cm}
\Output Embedding $\mathbf{e}=\left[e_1, \dots, e_5\right]^{\mathrm{T}}\in\mathbb{R}^5$
\vspace{0.1cm}

\State $\mathbf{x}_1, \mathbf{x}_2 \gets \left[\widehat{\boldsymbol{\lambda}}_{\text{max}}^{\mathrm{T}}, \widehat{\boldsymbol{r}}_{\text{max}}^{\mathrm{T}}\right]^{\mathrm{T}}, \left[\widehat{\boldsymbol{\lambda}}_{\text{min}}^{\mathrm{T}}, \widehat{\boldsymbol{r}}_{\text{min}}^{\mathrm{T}}\right]^{\mathrm{T}}$ \Comment{Construct input $\mathbf{x}_1, \mathbf{x}_2\in\mathbb{R}^{2k}$}
\vspace{0.1cm}
\State $\mathbf{y}_1 \gets \mathbf{W}_3^{\mathrm{T}}\sigma(\mathbf{W}_1^{\mathrm{T}}\mathbf{x})$ \Comment{First subnetwork, $\mathbf{y}_1\in\mathbb{R}^5$} 
\vspace{0.1cm}
\State $\mathbf{y}_2 \gets \mathbf{W}_3^{\mathrm{T}}\sigma(\mathbf{W}_2^{\mathrm{T}}\mathbf{x})$ \Comment{Second subnetwork, $\mathbf{y}_2\in\mathbb{R}^5$}
\vspace{0.1cm}
\State $\mathbf{y}_1 \gets \mathrm{sort}_{\downarrow}\!\left(\left|\mathbf{y}_1\right|\right)$ \Comment{Sort by magnitude (descending)}
\vspace{0.1cm}
\State $\mathbf{y}_2 \gets \mathrm{sort}_{\downarrow}\!\left(\left|\mathbf{y}_2\right|\right)$ \Comment{Sort by magnitude (descending)}

\State $\mathbf{e} \gets \left[\mathbf{y}_1^{\mathrm{T}}, \mathbf{y}_2^{\mathrm{T}}\right]^{\mathrm{T}}$
\vspace{0.2cm} 
\State \Return $\mathbf{e}$

\end{algorithmic}
\end{algorithm}

\noindent\paragraph{Embedding Layer for Approximating Matrix Functions. }For approximating matrix function such as inverse square root, we adopt the same embedding layer as for linear systems, as the top and bottom region of the eigenspectrum are also task-relevant. In our experiments, we choose $k=20$.

\section{Detailed Training Settings}\label{appendix:training_settings}

\subsection{Computer Resources}\label{sec:compute_resources}
All stages of the training of the neural network engine were performed on one NVIDIA A100 GPU with 40GB of memory.

\subsection{Training Data Curation}\label{sec:training_data}
As shown in Section~\ref{sec:train_diagonal_matrices}, we can train the model with only diagonal matrices, represented as vectors of eigenspectra. We construct a synthetic eigenspectrum generator that generates positive eigenvalue spectra by sampling a small set of continuous shape parameters using quasi-random Sobol sequences, ensuring broad and uniform coverage of admissible spectral configurations. Each spectrum is formed as a normalized blend of flat, exponential, and power-law decay profiles, with additional concavity modulation to control the condition number and induce slow spectral decay. To better reflect non-idealized operators, we introduce mild local irregularities through multiplicative noise and randomized tail perturbations, while preserving the overall spectral structure.  

For specific NLA tasks, the eigenspectrum generator also explicitly controls the spectral features most relevant to the task. For eigenvalue problems, since in practice polynomial preconditioners are most effective for matrices with clustered eigenvalues (slow decay), we apply rejection criteria that enforce unit normalization and small leading eigenvalue separation (e.g., $\lambda_2/\lambda_1$ close to one), yielding a curated set of eigenspectra. For preconditioned linear systems and matrix function approximation, since the condition numbers of operators are most critical for preconditioning and approximation, we explicitly control the condition number of the generated eigenspectra, and create a training set that covers a wide range of condition numbers.

\noindent\paragraph{Training Data Normalization.}
To maintain training stability while ensuring robustness of the learned model, we preprocess the training matrices with normalization. Specifically, we ensure that the largest eigenvalue of the matrix is bounded and around 1. During data generation, we first enforce all synthetic matrices to have unit spectral norm, and we then perform a perturbation scaling for each matrix by either 1) randomly scaling each matrix with a random scalar in $[1-\epsilon, 1+\epsilon]$ (we choose $\epsilon=0.2)$, or 2) scaling with a coarse approximation of the spectral norm. This ensures that the training matrices have bounded norms but not strictly unit-norm, which could otherwise lead to overfitting to this specific spectral property, as in practice we cannot ensure unit norm for realistic matrices and can only normalize the matrix by approxiate spectral norm.

\subsection{Details of Optimization Objectives}
\label{appendix:optimization_objectives}
In the \framework framework, we use the unified optimization loss function defined in~\cref{eq:general_loss_function} of Section~\ref{sec:training} to train neural networks to discover spectral algorithms. Here we describe the detailed loss function and its design philosophy for the primary applications in our paper.

\noindent\paragraph{Eigenvalue Problems. }
To discover a preconditioning algorithm $P(\cdot)$ for eigenvalue problems, we expect the learned algorithm to have the following properties: 1) it outperforms standard power or subspace iteration methods; and 2) it can be extended to higher degrees, so that for any first $k$ iterations of the algorithm, it can be a degree-$k$ preconditioner outperforming $k$ power/subspace iterations. For 1), we choose the baseline method as standard power/subspace iteration; and for 2), we adopt a layer-wise loss function that optimizes the objective for the first $k$ neural network layers ($k=1, \dots, d$). Adopting the formula in ~\cref{eq:obj_eigenproblem}, we define the objective for our model:
\begin{align}
    r_{\text{nn}}(\mathbf{X}) &= \frac{\operatorname*{min}_{i=1, \dots, k}|P(\lambda_i(\mathbf{X}))|}{\operatorname*{max}_{I\subset\{1, \dots, m\}, |I|=l}\operatorname{min}_{j\in I}|P(\lambda_j(\mathbf{X}))|} ,
\end{align}
and for standard subspace iteration:
\begin{align}
    r_{\text{subspace}}(\mathbf{X}) &= \frac{\operatorname*{min}_{i=1, \dots, k}|\lambda_i(\mathbf{X})|}{\operatorname*{max}_{I\subset\{1, \dots, m\}, |I|=l}\operatorname{min}_{j\in I}|\lambda_j(\mathbf{X})|} .
\end{align}

\begin{figure}[!htb]
    \centering
    \begin{subfigure}[t]{0.45\linewidth}
        \includegraphics[width=\textwidth]{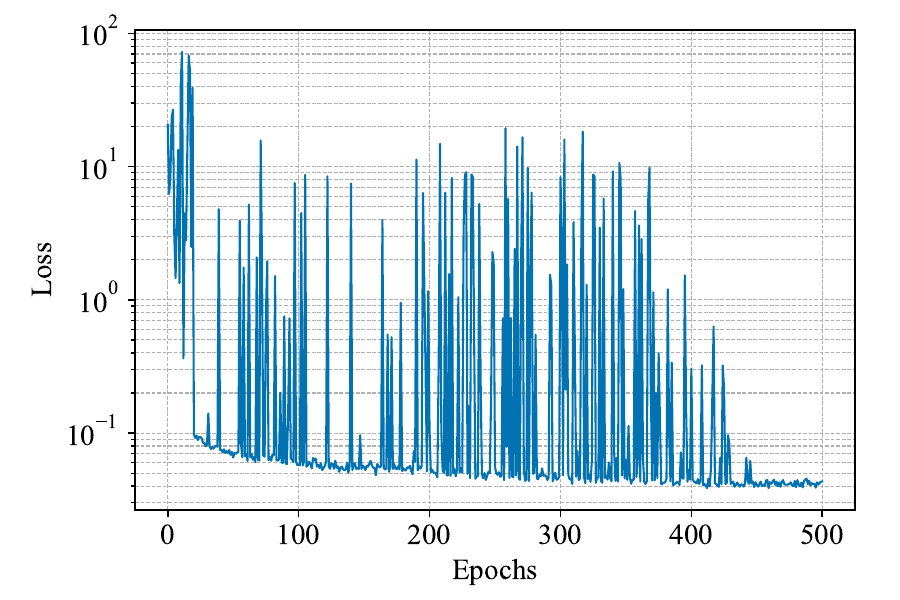}
        \caption{Inverse of $\rho_{\text{log}}$}
    \end{subfigure}
    \begin{subfigure}[t]{0.45\linewidth}
        \includegraphics[width=\textwidth]{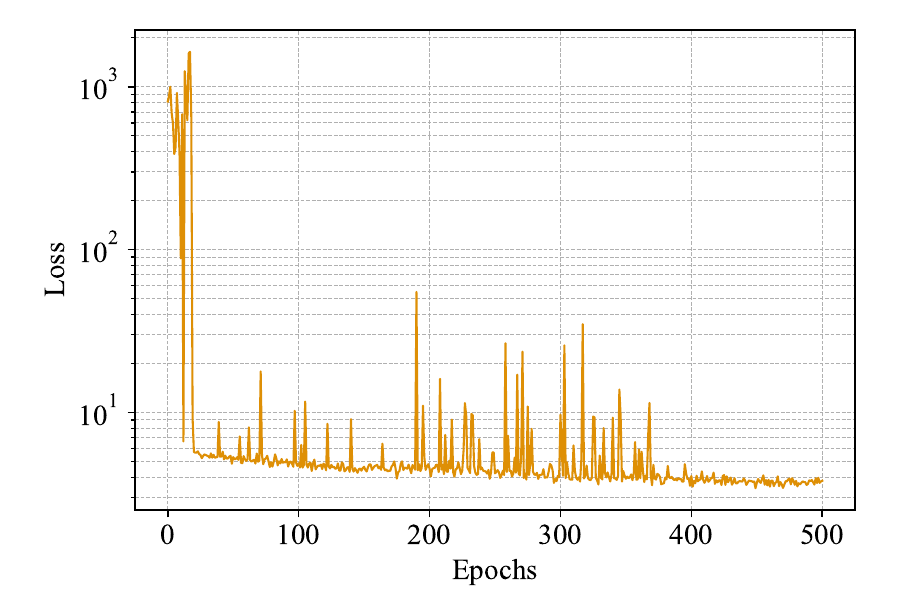}
        \caption{$\mathcal{L}_{\text{reg}}$}
    \end{subfigure}

    \caption{Last-layer training loss of training neural network to discover preconditioning algorithms for eigenvalue problems.}
    \label{fig:eigs_loss} 
    \vspace{-5mm}
\end{figure}

Therefore, the loss function $\mathcal{L}$ computes a weighted sum of the loss term $\mathcal{L}_k$ for each layer, with weight $w_k$ for layer $k$. During training, the weight term $w_k$ is dynamically adjusted to focus on optimizing the loss of different layers. For each layer, the loss follows the form in ~\cref{eq:general_loss_function} (we omit the $\mathcal{L}_{\text{struct}}$ as it is not necessary): the inverse of \objective, and $\mathcal{L}_{\text{reg}}$, which penalizes samples on which the algorithms fail:
\begin{gather}
    \label{eq:loss_func_eigenproblem}
    \mathcal{L} = \sum_{k=1}^{D} w_k \mathcal{L}_k\\
    \mathcal{L}_k = \frac{1}{N}\sum_{i=1}^{N}\left[\underbrace{\frac{k\cdot\log r_{\text{subspace}}}{\operatorname{clamp}\left(\log r_{\text{nn}}, \epsilon\right)}}_{\mathcal{L}_{\text{obj}}} + \underbrace{10\exp{\left[-10\cdot r_{\text{nn}} - \epsilon\right]}}_{\mathcal{L}_{\text{reg}}}\right] .
\end{gather}
Note that $\rho_{\text{log}}$ is a ratio of two logarithms, and the denominator is not guaranteed to be positive throughout training. Therefore we clamp the denominator with a small positive scalar. Figure~\ref{fig:eigs_loss} shows the trajectory of two loss terms of the last layer. At the early stage of training, $\mathcal{L}_{\text{reg}}$ dominates the loss, during which the model learns to produce algorithms that perform moderately well on each training sample. Once $\mathcal{L}_{\text{reg}}$ diminishes, $\rho_{\text{log}}$ becomes the primary optimization objective, encouraging the model to find parametrization outperforming the baseline method as much as possible.

\noindent\paragraph{Dynamic Weight Adjustment to Loss Function.}
At the beginning of the training, $w$ is larger for earlier layers, while towards the end of the training, $w$ is larger for later layers. Suppose the total number of training steps is $T$, the scheduling function for $w_k$ at step $t$ is:
\begin{align}
    s &= \lfloor\frac{T}{t}\rfloor \\
    w_k^t &= \begin{cases}
        1 & \text{if } k = s \\
        \frac{1}{(k - s)^2} & \text{otherwise }\\
    \end{cases} .
\end{align}
During training, we normalize the weight by the sum across all layers: $w_k = \frac{w_k}{\sum_{i=1}^{D}w_i}$. We call $s$ the ``Anchor Layer,'' which has the largest weight in the loss function. As training proceeds, the anchor layer moves from the first layer to the last layer. The motivation of this design is to ensure that the polynomial of \textit{any} degree generated by the model can accelerate the convergence when applied to the matrix, instead of only the largest-degree polynomial. This can better disentangle the roles of each layer.

\begin{figure}[!htb]
    \centering
    \begin{subfigure}[t]{0.3\linewidth}
        \includegraphics[width=\textwidth]{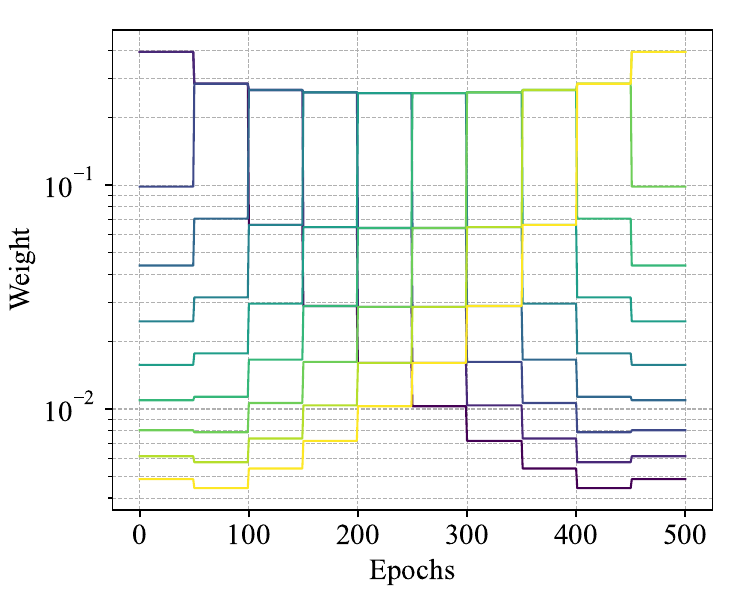}
        \caption{Loss weight of each layer}
    \end{subfigure}
    \begin{subfigure}[t]{0.3\linewidth}
        \includegraphics[width=\textwidth]{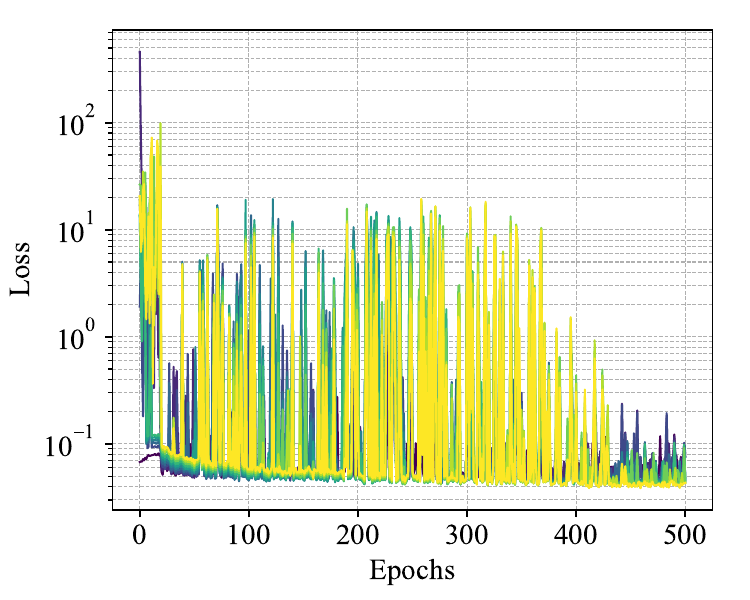}
        \caption{Inverse of $\rho_{\text{log}}$}
    \end{subfigure}
    \begin{subfigure}[t]{0.3\linewidth}
        \includegraphics[width=\textwidth]{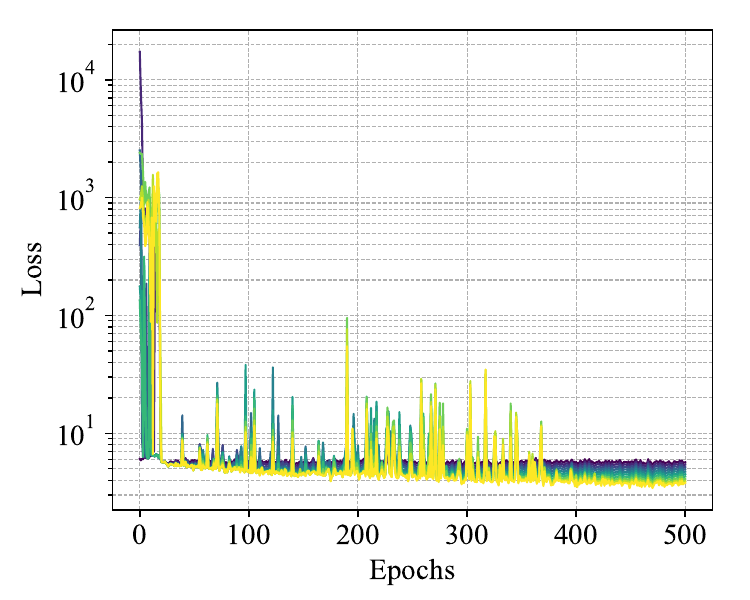}
        \caption{$\mathcal{L}_{\text{reg}}$}
    \end{subfigure}
    \begin{subfigure}[t]{0.05\linewidth}
        \includegraphics[width=\textwidth]{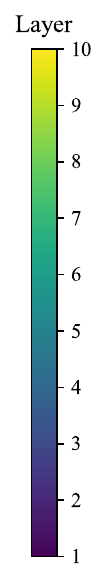}
    \end{subfigure}

    \caption{Dynamic adjustment of training loss of all layers. In the early stage of training, earlier layers have a larger weight in the total loss, and are optimized more; towards the end of training, later layers have a larger weight in the total loss, and are optimized more.}
    \label{fig:eigs_loss_all_layers} 
    \vspace{-5mm}
\end{figure}

\noindent\paragraph{Preconditioned Linear Systems.}
Similarly, we design the optimization objective for generating preconditioning algorithm $P(\cdot)$ for solving linear systems, based on the objective in ~\cref{eq:obj_linsolve}. When training with diagonal matrices $\mathbf{X} = \operatorname{diag}(\boldsymbol{\lambda})$, the residual objective can be reduced to
\begin{align}\label{eq:obj_linsolve_nn}
    r_{\text{nn}}(\mathbf{X}) &= \operatorname*{max}_j \left| 1 - \frac{P(\lambda_j(\mathbf{X}))\lambda_j(\mathbf{X})}{\operatorname{max}_i|P(\lambda_i(\mathbf{X}))\lambda_i(\mathbf{X})|} \right| ,
\end{align}
which is easy to obtain during training. Note that due to the normalization, we have $r_{\text{nn}}\in[0, 2]$. As the baseline, we use the following residual:
\begin{align}\label{eq:obj_linsolve_richardson}
    r_{\text{Richardson}}(\mathbf{X}) &= 1 - \frac{\operatorname{min}_i\lambda_i(\mathbf{X})}{\operatorname{max}_i\lambda_i(\mathbf{X})} .
\end{align}
This represents the convergence rates of a basic Richardson method for solving linear systems. Following the form in ~\cref{eq:general_loss_function} define the optimization objective as

\begin{gather}
    \label{eq:loss_func_linsolve}
    \mathcal{L}_k = \frac{1}{N}\sum_{i=1}^{N}\left[\underbrace{\operatorname{clamp}\left(\frac{k\cdot\log r_{\text{Richardson}}(\mathbf{X}_i)}{\log r_{\text{nn}}(\mathbf{X}_i)}, \epsilon\right)}_{\mathcal{L}_{\text{obj}}} + \underbrace{r_{\text{nn}}(\mathbf{X}_i)}_{\mathcal{L}_{\text{reg}}} + \underbrace{c\cdot\frac{\operatorname{max}_i|P(\lambda_i(\mathbf{X}_i)|}{\operatorname{min}_i|P(\lambda_i(\mathbf{X}_i)|}}_{\mathcal{L}_{\text{struct}}}\right]
\end{gather}
on training dataset $\{\mathbf{X}_i\}_{i=1}^N$, where we set $c = 5\times d^3$, $d$ is the degree of the constructed polynomial. The term $\mathcal{L}_{\text{struct}}$ is a structural constraint that minimizes the condition number of the preconditioner $P(\mathbf{X})$. Our empirical study on realistic matrices suggests that constraining the condition number of the preconditioner can be useful for improving convergence stability.  

In the early stage of training, $r_{\text{nn}}$ is unstable and would revolve around $1$, causing the $\mathcal{L}_{obj}$ to fluctuate around 5, which is difficult to optimize. Therefore we perform sample-wise clipping of $\mathcal{L}_{\text{obj}}$ from below with a small positive constant ($10^{-8}$), while optimizing $\mathcal{L}_{\text{reg}}$ and $\mathcal{L}_{\text{struct}}$. As the training proceeds, $\mathcal{L}_{\text{reg}}$ becomes smaller than $1$ across the dataset, where $\mathcal{L}_{\text{obj}}$ is activated, becoming the primary optimization objective.

\noindent\paragraph{Approximating Matrix Functions.}
In this work, we consider learning a polynomial $P(\cdot)$ that approximate matrix inverse square root, and we select the Neumann (Taylor) series approximation $T(\cdot)$ as our baseline reference method. Following ~\cref{eq:obj_matrix_function}, we define the objective for our model and baseline for training on synthetic diagonal matrices $\mathbf{X} = \operatorname{diag}(\boldsymbol{\lambda})$:
\begin{gather}\label{eq:obj_matfunc_diag}
    r_{\text{nn}}(\mathbf{X}) = \operatorname*{max}_i|1 - P(\widetilde{\lambda_i}(\mathbf{X}))\widetilde{\lambda_i}(\mathbf{X})^{\frac{1}{2}}|\\
    r_{\text{Neumann}}(\mathbf{X}) = \operatorname*{max}_i|1 - T(\lambda_i(\mathbf{X}))\lambda_i(\mathbf{X})^{\frac{1}{2}}| ,
\end{gather}
where $\widetilde{\lambda}(\mathbf{X})$ are eigenvalues augmented with random sampled values in $[\lambda_{\text{min}}(\mathbf{X}), \lambda_{\text{max}}(\mathbf{X})]$. 
This is to regularize the approximation accuracy of $P(\cdot)$ not only on the discrete eigenvalues, but also on the continuous region in which the eigenspectrum is located. 
This augmentation promotes the learning of a more robust approximation algorithm. Based on the form in ~\cref{eq:general_loss_function} (we omit the $\mathcal{L}_{\text{struct}}$ as it is not necessary for matrix function approximation), we design the loss function as
\begin{align}
    \label{eq:loss_func_matrix_func}
    \mathcal{L}_k = \frac{1}{N}\sum_{i=1}^{N}\left[\underbrace{\operatorname{clamp}\left(\frac{\log r_{\text{Neumann}}(\mathbf{X}_i)}{\log r_{\text{nn}}(\mathbf{X}_i)}, \epsilon\right)}_{\mathcal{L}_{\text{obj}}} + \underbrace{r_{\text{nn}}(\mathbf{X}_i)}_{\mathcal{L}_{\text{reg}}}\right] .
\end{align}

\subsection{Extending Polynomial Degree}
For eigenvalue problems, problem difficulty strongly correlates with eigenvalue gaps, which requires polynomials of different degree to achieve reliable performance. Our \framework framework offers an easy way to extend the backbone model to higher degrees, where one can append more neural network to already pre-trained backbone models, and continue training the model. This works because the optimization objective in ~\cref{eq:loss_func_eigenproblem} promotes the model to learn a solution such that, for any degree $k$, polynomials generated by the first $k$ layers can sufficiently accelerate the convergence on eigenvalue problems. Figure~\ref{fig:eigval_gap_vs_degree} shows that any degree $k$ polynomial preconditioner from the first $k$ layers consistently has a larger eigenvalue gap than standard iterative methods.

\section{Training Details}\label{appendix:training_details}

\subsection{Backbone Model (Pre-training)}\label{appendix:pre_training}
For pre-training, we follow the optimization objectives described in Appendix~\ref{appendix:optimization_objectives} for each NLA task. We perform training using synthetic diagonal matrices (represented as vectors of eigenvalues).  

\noindent\paragraph{Eigenvalue Problems. }For eigenvalue problems, we create a synthetic dataset of size 50000 with matrix dimensions ranging from $[50, 1000]$, which contains matrices that have slow decay in dominant eigenvalues. We use a batch size of 500 and train the model for 1000 epochs. We use the AdamW optimizer with cosine learning rate decay after $10\%$ of warmup steps. We search for the best learning rate among $\{10^{-3}, 5\times10^{-4}, 10^{-4}\}$ and select the best learning rate, and use a weight decay of $5\times 10^{-4}$.

\noindent\paragraph{Preconditioned Linear Systems. }For preconditioner linear systems, we create a synthetic dataset of size 50000 with matrix dimensions ranging from $[50, 1000]$, which contains matrices whose condition number ranges $[10^2, 10^{5}]$. We use a batch size of 500 and train the model for 1000 epochs. We use the AdamW optimizer with cosine learning rate decay after $10\%$ of warmup steps. We search for the best learning rate among $\{10^{-3}, 5\times 10^{-4}, 10^{-4}\}$ and select the best learning rate, and use a weight decay of $5\times 10^{-4}$.

\noindent\paragraph{Approximating Matrix Functions. }For matrix function approximations, we adopt the same training strategy as for preconditioned linear systems.

\subsection{Embedding Model (Post-training)}
\label{appendix:post_training_details}
Following the introduction of the structure of the embedding layer in Section~\ref{appendix:embedding_structure}, we introduce the training process of the embedding layer. The post-training stage aims to enable the neural network engine to generate effective algorithms when only given spectral probes (coarse spectral estimates). We therefore train an embedding layer that encodes information about structures of task-relevant spectral regions and generates embeddings. In practice, we use the same optimization objective as training the backbone model, and we freeze the backbone model while training the embedding layer. 

\noindent\paragraph{Eigenvalue Problems}
For eigenvalue problems, we design the embedding layer as shown in Algorithm~\ref{alg:embed_layer_eigenproblem}, which adapts to different target numbers of eigenvalues $k$ and subspace dimensions $l$ via adaptive truncation/padding of the input spectral probe. To enable the model that generalizes to different $k$ and $l$, we construct a training dataset with a mixture of $k$, $l$, and spectral probes obtained with a varying number of subspace iterations.  

Specifically, we create a synthetic dataset with matrix dimension ranging from $[200, 2000]$, spectral probes using subspace dimension $l$ in $[20, 100]$, and number of subspace iterations in $[1, 500]$. We train the embedding layer while freezing the backbone model for 500 epochs, with a dataset of size 50000 and a batch size of 200. We use AdamW optimizer with cosine learning rate decay after $10\%$ of warmup steps. We search for the best learning rate among $\{10^{-4}, 10^{-5}, 5\times 10^{-6},  10^{-6}\}$ and select the best learning rate, and use a weight decay among the best of $\{10^{-2}, 10^{-3}, 5 \times 10^{-4}\}$.

\noindent\paragraph{Preconditioned Linear Systems. }For preconditioned linear systems, we use the embedding layer design in Algorithm~\ref{alg:embed_layer_linsolve}, we design a training dataset that consists of matrices with dimensions in $[200, 2000]$, and we constrain the condition number of matrices spanning $[10, 10^5]$. Spectral probes are obtained with different numbers of subspace iterations in $[1, 500]$. The spectral probe contains the largest and smallest $k$ eigenvalue estimates and corresponding residual norms, obtained using subspace itreations. In practice, we choose $k=20$, and for estimates of the smallest eigenvalues, we obtain them by performing subspace iteration on shifted matrices to obtain the largest eigenpairs of the shifted matrix. We then obtain the estimates for the smallest eigenvalue by reverse-shifting the estimations and compute the residual using approximated eigenvectors on the unshifted matrix.  

We train the embedding layer while freezing the backbone model for 500 epochs, with a dataset of size 50000 and a batch size of 200. We use AdamW optimizer with cosine learning rate decay after $10\%$ of warmup steps. We search for the best learning rate among $\{10^{-3}, 5\times 10^{-4}, 10^{-4}\}$ and select the best learning rate, and we use a weight decay among the best of $\{10^{-2}, 10^{-3}, 5 \times 10^{-4}\}$.

\noindent\paragraph{Approximating Matrix Functions. }For approximating matrix functions, we adopt the same embedding layer structure as preconditioned linear systems, shown in Algorithm~\ref{alg:embed_layer_linsolve}. We construct the training dataset with the same method as for linear systems, while increasing the portion of matrices with small condition numbers, as the polynomial method is most suitable for moderate condition number regime.

We train the embedding layer while freezing the backbone model for 500 epochs, with a dataset of size 50000 and a batch size of 200. We use AdamW optimizer with cosine learning rate decay after $10\%$ of warmup steps. We search for the best learning rate among $\{10^{-3}, 5 \times  10^{-4}, 10^{-4}\}$ and select the best learning rate, and we use a weight decay among the best of $\{10^{-2}, 10^{-3}, 5 \times 10^{-4}\}$.

\section{Extended Discussion of Numerical Experiments on Real-World Eigenproblems and Linear Systems}
\label{sec:realworld_appendix}

We provide a detailed description of the experimental setup used in the evaluation in~\cref{sec:experiments}, together with additional results and analysis. 
This includes absolute and relative performance of our discovered preconditioning methods, sensitivity to spectral probe quality, and comparisons with state-of-the-art Chebyshev preconditioning.

\subsection{Experimental setting}

We evaluate preconditioners by iteration count reduction, a natural metric for comparing polynomial preconditioners of the same degree. This metric is especially relevant when memory and communication dominate, Krylov basis storage is impractical, and global inner products make outer Krylov iterations expensive. In such regimes, polynomial preconditioning based on short recurrences is a standard acceleration strategy; see, e.g.,~\cite{saad2003iter,saad2011eig,bergamaschi2021parallel}.

The \framework engines are trained offline and, at inference time, are called from \textsc{Matlab} to produce polynomial recurrence coefficients from spectral probes. For eigenvalue problems, these probes are obtained by running 20 iterations of \texttt{eigs}, which implements the Krylov--Schur algorithm. From the resulting Ritz spectrum, the engine receives 10 eigenvalue estimates associated with the target region, corresponding to the first $k$ requested eigenvalues, and 94 eigenvalue estimates associated with the out-of-target region, corresponding to the subsequent Ritz values. If fewer than 10 target values are available, we sample from the available values with replacement; if more than 10 are available, we uniformly subsample 10 values. The same procedure is used to obtain exactly 94 out-of-target values.

For linear systems, spectral probes are produced by 50 or 200 Lanczos steps without storing the Lanczos basis. From the resulting approximate spectrum, we select the 20 largest and 20 smallest Ritz values and use them as the spectral input to the neural engine.

The forward pass of the \framework engine has negligible cost compared to other computations. The overall preconditioner construction cost is dominated by the spectral probe stage: a cost equivalent to 20 iterations of the corresponding unpreconditioned eigensolver or 50 iterations of the corresponding unpreconditioned linear solver, and is also negligible for most tested systems, while modest for the remaining cases.

We focus on spd operators and spd preconditioners. Indefinite eigenvalue problems are converted to spd form by implicitly squaring the operator.

\subsection{Polynomial Preconditioners as Second-level Accelerators}\label{sec:second_level}
In linear systems experiments, we first apply a classical inexpensive preconditioner and then construct a polynomial preconditioner for second-level acceleration. In most cases, the first-level preconditioner is Jacobi (i.e. $\mathrm{diag}(\mathbf{A})^{-1}$); whenever other preconditioners, such as AMG or iChol(0), are used instead, this is stated explicitly. We describe this approach below.

Consider iterative solution of a linear system
\begin{equation} \label{eq:lin_sys}
    \mathbf{A}\mathbf{x} = \mathbf{b}
\end{equation}
with an spd first-level preconditioner $\mathbf{B}$. We can further accelerate the solver by applying a polynomial preconditioner to $\mathbf{A}\mathbf{B}$. Specifically, we can construct a polynomial $P(\mathbf{A}\mathbf{B})$ and use the composite preconditioner $\widehat{\mathbf{B}} = \mathbf{B} P(\mathbf{A}\mathbf{B})$
for the solution of~\cref{eq:lin_sys}. The resulting preconditioned operator is
$\mathbf{A}\widehat{\mathbf{B}}
    =
    \mathbf{A}\mathbf{B} P(\mathbf{A}\mathbf{B}).$
Its spectrum coincides with that of
\[
    \widehat{\mathbf{A}} P(\widehat{\mathbf{A}}),
    \qquad
    \widehat{\mathbf{A}}
    =
    \mathbf{B}^{1/2}\mathbf{A}\mathbf{B}^{1/2}.
\]
Thus, polynomial acceleration can be realized on an spd operator $\widehat{\mathbf{A}}$.

In practice, however, an explicit factorization of $\mathbf{B}$ may be unavailable, as is often the case for AMG. This makes it impractical to apply an eigensolver warm start directly to $\widehat{\mathbf{A}}$ in order to compute a spectral probe. We therefore obtain spectral probes through a generalized Lanczos process applied to
$\mathbf{A}\mathbf{v}_i = \lambda_i \mathbf{B}^{-1}\mathbf{v}_i,$
which is equivalent to the standard eigenproblem
$
\widehat{\mathbf{A}}\mathbf{u}_i = \lambda_i \mathbf{u}_i.$
This yields Ritz values $\lambda_i$  together with residual estimates
\[
    \left\|
    \mathbf{A}\mathbf{v}_i - \lambda_i \mathbf{B}^{-1}\mathbf{v}_i
    \right\|_{\mathbf{B}}
    =
    \left\|
    \widehat{\mathbf{A}}\mathbf{u}_i - \lambda_i \mathbf{u}_i
    \right\|_2,
    \qquad
    \mathbf{u}_i = \mathbf{B}^{-1/2}\mathbf{v}_i,
\]
which are used as inputs to the neural engine.

\subsection{Test Matrices}
Test matrices are drawn from the SuiteSparse Matrix Collection and cover electronic-structure, thermal finite element, structural/mechanics, and circuit/model-reduction applications. Matrix sizes range from $n = 468$ to $n = 1.6\times 10^6$, with estimated condition numbers between $50$ and $4.5\times 10^6$. We also use a finite difference discretization of the Laplacian on the unit square, of size $6084$, with known eigenvalues, to compare the behavior of the preconditioned operator induced by \framework with that obtained from Chebyshev acceleration; see Figure~\ref{fig:compare_polynomial}.

\begin{table*}[t]
\centering
\small
\caption{Large matrices from the SuiteSparse Matrix Collection used in our experiments.}
\label{tab:test_matrix_details}
\setlength{\tabcolsep}{4pt}
\renewcommand{\arraystretch}{1.10}

\begin{subtable}[t]{0.48\textwidth}
\centering
\caption{Linear Systems}
\label{tab:test_matrix_details_a}
\begin{tabular}{@{} l r c r @{}}
\toprule
\textbf{Matrix $\mathbf{X}$} & \textbf{dim} &
\textbf{Precond $\mathbf{B}$} &
$\boldsymbol{\mathrm{cond}(\mathbf{X B})}$ \\
\midrule
\texttt{Laplacian}          & 6084     & Jacobi & 2.5e3 \\
\texttt{Dubcova2}           & 65025    & Jacobi & 1.1e4 \\
\texttt{thermal1}           & 82654    & Jacobi & 5e5 \\
\texttt{FEM\_3D\_thermal2}  & 147900   & Jacobi & 5e1 \\
\texttt{G2\_circuit}        & 150102   & Jacobi & 6.8e5 \\
\texttt{shipsec5}           & 179860   & Jacobi & 1.4e6 \\
\texttt{parabolic\_fem}     & 525825   & Jacobi & 2.3e5 \\
\texttt{Fault\_639}         & 638802   & Jacobi & 6.3e5 \\
\texttt{Emilia\_923}        & 923136   & Jacobi & 1.7e6 \\
\texttt{thermal2}           & 1228045  & Jacobi & 4.5e6 \\
\texttt{G3\_circuit}        & 1585478  & Jacobi & 2.3e5 \\
\bottomrule
\end{tabular}
\end{subtable}
\hfill
\begin{subtable}[t]{0.48\textwidth}
\centering
\caption{Eigenproblems}
\label{tab:test_matrix_details_b}
\begin{tabular}{@{} l r c r @{}}
\toprule
\textbf{Matrix $\mathbf{X}$} & \textbf{dim} &
\textbf{squared} &
$\boldsymbol{\mathrm{cond}(\mathbf{X})}$ \\
\midrule
\texttt{thermal1}          & 82654    & no  & 3.2e5 \\
\texttt{thermal2}          & 1228045  & no  & 4.0e6 \\
\texttt{SiO2}              & 155331      & yes  & 5.6e6 \\
\texttt{CO}         & 221119    & yes  & 6.3e7 \\
\bottomrule
\end{tabular}
\end{subtable}

\vspace{4pt}
\end{table*}

Table~\ref{tab:test_matrix_details} summarizes the large real-world matrices used in our experiments. For linear systems, it reports the first-level preconditioners applied before polynomial acceleration. For eigenvalue problems, it indicates whether the operator is implicitly squared to obtain an spd formulation.

\subsection{Quality of Learned Second-level Accelerators for Linear Systems.}
Table~\ref{tab:realistic_matrix_results} evaluates the preconditioners produced by \framework engines on the systems from Table~\ref{tab:test_matrix_details}.

\begin{table*}[!t]
\centering
\small
\caption{Quality of preconditioners produced by \framework.}
\label{tab:realistic_matrix_results}
\setlength{\tabcolsep}{6pt} 
\renewcommand{\arraystretch}{1.10}

\begin{subtable}{\textwidth}
\centering
\caption{Linear systems: CG iteration counts for linear systems at residual tolerance $10^{-10}$ using \framework acceleration with 50-step and 200-step spectral probes, Neumann acceleration, and no acceleration. Results are reported over five random seeds.}
\label{tab:realistic_matrix_results_a}
\begin{tabular}{@{} l r r r r @{}}
\toprule
\textbf{Matrix XB} & 
\makecell{\textbf{numIters}\\(probe=200)} & 
\makecell{\textbf{numIters}\\(probe=50)} & 
\makecell{\textbf{numIters}\\(Neumann)} &  
\makecell{\textbf{numIters}\\(no acceleration)} \\
\midrule
\texttt{laplacian}         & 29 $\pm$ 0   & 27 $\pm$ 0   & 79 $\pm$ 0    & 275 $\pm$ 1 \\
\texttt{Dubcova2}          & 36 $\pm$ 0   & 35 $\pm$ 0   & 102 $\pm$ 1   & 357 $\pm$ 3 \\
\texttt{thermal1}          & 154 $\pm$ 1  & 162 $\pm$ 2  & 480 $\pm$ 6   & 1642 $\pm$ 18 \\
\texttt{FEM\_3D\_thermal2} & 7 $\pm$ 0    & 6 $\pm$ 0    & 16 $\pm$ 0    & 56 $\pm$ 0 \\
\texttt{G2\_circuit}       & 205 $\pm$ 2  & 219 $\pm$ 1  & 641 $\pm$ 3   & 2242 $\pm$ 12 \\
\texttt{shipsec5}          & 421 $\pm$ 2  & 442 $\pm$ 1  & 1307 $\pm$ 5  & 4528 $\pm$ 25 \\
\texttt{parabolic\_fem}    & 317 $\pm$ 6  & 337 $\pm$ 7  & 984 $\pm$ 22  & 3444 $\pm$ 70 \\
\texttt{Fault\_639}        & 800 $\pm$ 13 & 890 $\pm$ 12 & 2480 $\pm$ 13 & $>$7900 \\
\texttt{Emilia\_923}       & 1097 $\pm$ 5 & 1148 $\pm$ 6 & 3411 $\pm$ 27 & 10497 $\pm$ 5 \\
\texttt{thermal2}          & 592 $\pm$ 3  & 625 $\pm$ 2  & 1847 $\pm$ 9  & 6365 $\pm$ 27 \\
\texttt{G3\_circuit}       & 397 $\pm$ 4  & 436 $\pm$ 8  & 1216 $\pm$ 10 & 4306 $\pm$ 48 \\
\bottomrule
\end{tabular}
\end{subtable}

\vspace{18pt} 

\begin{subtable}{\textwidth}
\centering
\caption{Eigenproblems: \framework preconditioners are compared to basic power method-based (i.e., $P(\mathbf{X})=\mathbf{X}^d$) preconditioners.  The number of \texttt{eigs} iterations required to reach relative eigenvalue error $10^{-10}$. For all matrices, we compute $k{=}10$ and $k{=}20$ eigenvalues using \texttt{eigs} with $k{+}5$ requested eigenvalues and subspace dimension $l{=}4k$.} 
\label{tab:realistic_matrix_results_b}
\setlength{\tabcolsep}{4pt}
\begin{tabular}{@{} l r r r r @{}}
\toprule
\textbf{Matrix X} & 
\makecell{\textbf{numIters}\\(\framework, $\lambda_{10}$)} & 
\makecell{\textbf{numIters}\\(\framework, $\lambda_{20}$)} & 
\makecell{\textbf{numIters}\\(power, $\lambda_{10}$)} & 
\makecell{\textbf{numIters}\\(power, $\lambda_{20}$)} \\
\midrule
\texttt{thermal1} & 4   & 3 & 19 & 12 \\
\texttt{thermal2} & 21 & 11 & 112 & 45 \\
\texttt{SiO2}     & 23      & 23 & $>$500 & 108 \\
\texttt{CO}       & 55      & 27 & $>$500 & 126 \\
\bottomrule
\end{tabular}
\end{subtable}

\end{table*}

We assess preconditioner quality by the number of CG iterations required to reach residual tolerance $10^{-10}$, using spectral probes obtained from either 50 or 200 Lanczos steps. For reference, we also report iteration counts for Neumann acceleration and unaccelerated CG at the same tolerance. The results show that the learned preconditioners substantially reduce iteration counts across all tested problems and remain effective over a wide range of condition numbers, even when constructed from coarse spectral probes.

\begin{figure}[!htb]
    \centering
    \begin{subfigure}{1.0\linewidth}
        \includegraphics[width=\textwidth]{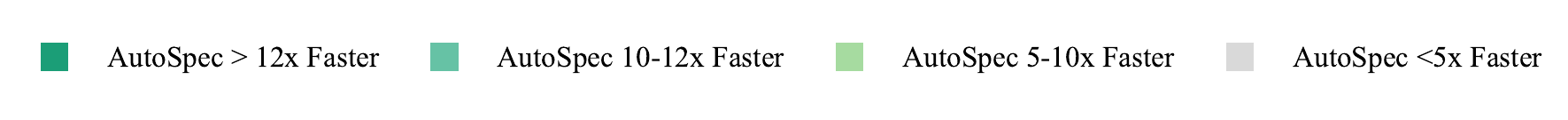}
    \end{subfigure}
    
    \begin{subfigure}[t]{0.25\linewidth}
        \includegraphics[width=\textwidth]{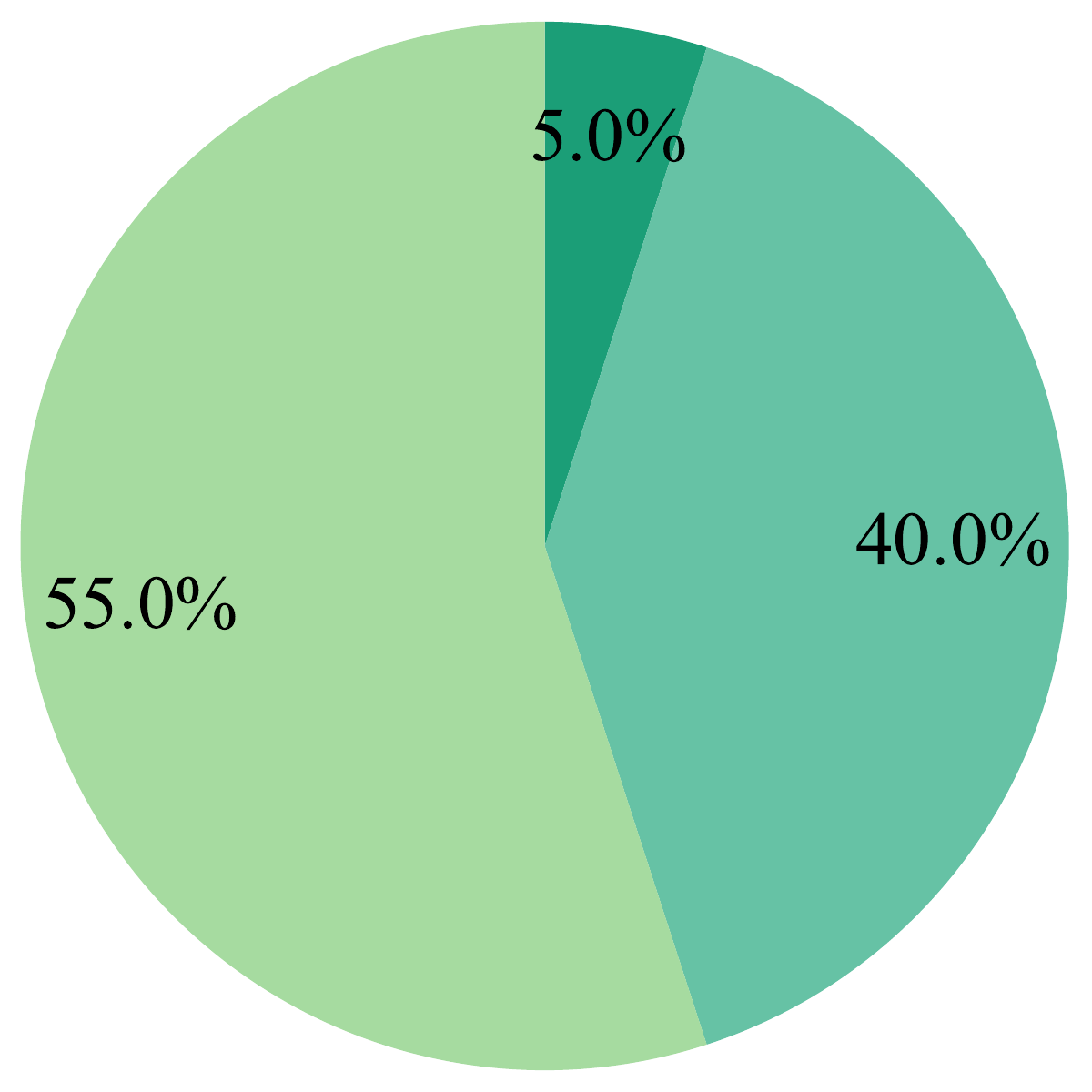}
        \caption{50 Probe Iterations}
    \end{subfigure}
    \hspace{1cm}
    \begin{subfigure}[t]{0.25\linewidth}
        \includegraphics[width=\textwidth]{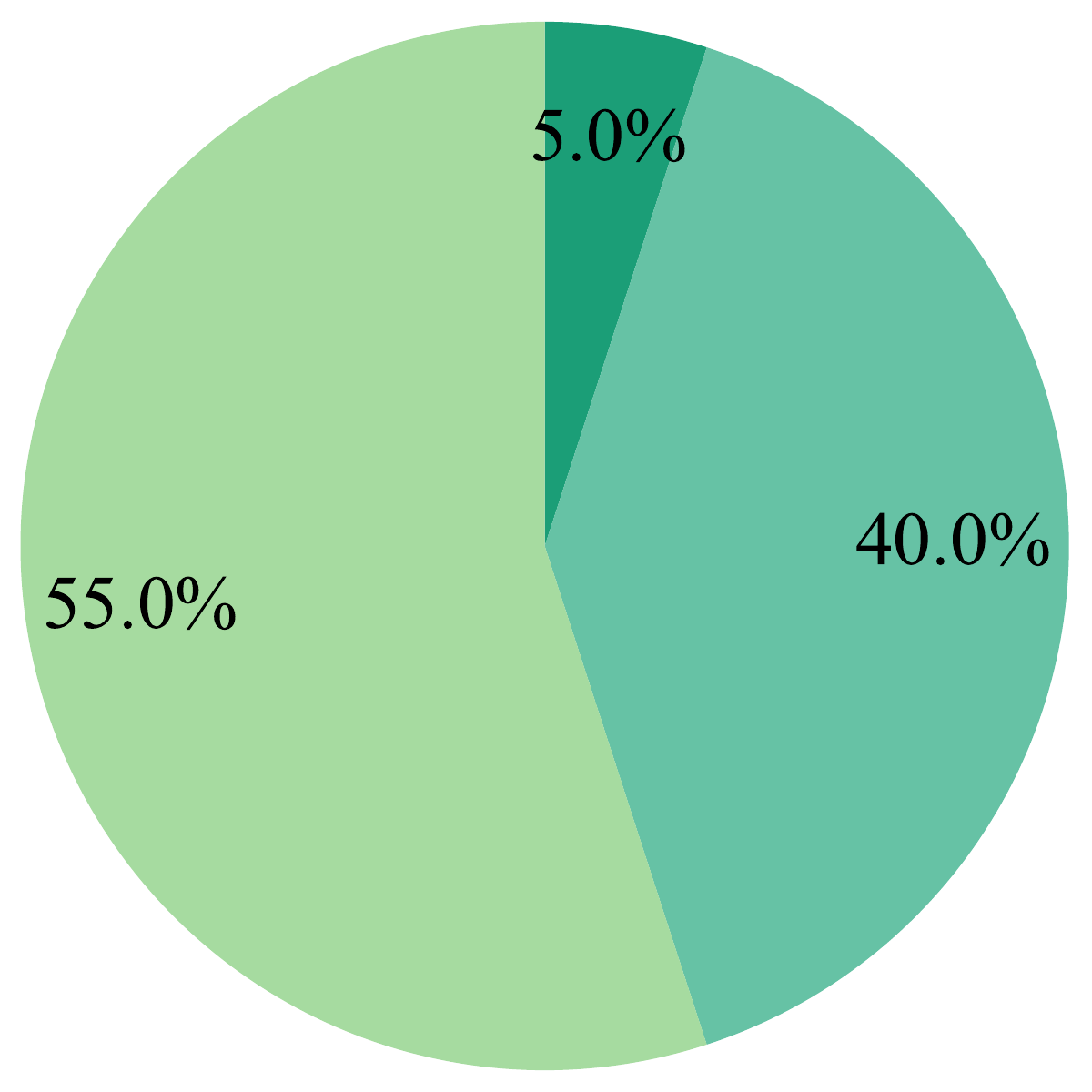}
        \caption{200 Probe Iterations}
    \end{subfigure}

    \caption{Comparing average iteration counts over 5 seeds of unaccelerated CG, and CG accelerated by \framework, on all spd SuiteSparse matrices with sizes between $10^4$ and $10^5$ and at most $10^6$ nonzeros.  }
    \label{fig:autospec_vs_base} 
\end{figure}

To broaden the empirical analysis, we evaluate all spd matrices in the SuiteSparse Matrix Collection with sizes between $10^4$ and $10^5$ and at most $2\times 10^6$ nonzeros; see Figure~\ref{fig:autospec_vs_base} for a statistical summary of the resulting iteration counts. In this broader evaluation, polynomial preconditioners consistently reduce CG iteration counts, often by up to an order of magnitude.

\subsection{Quality of Learned Preconditioners for Eigenvalue Problems.}
For the eigenproblems defined by the matrices in Table~\ref{tab:test_matrix_details_b}, Table~\ref{tab:realistic_matrix_results_b} reports the number of \texttt{eigs} iterations required to attain a relative eigenvalue error of $10^{-10}$ on the spectrally transformed operator $\mathbf{I} - \mathbf{X}/1.1c$, where $c \approx \lambda_{\mathrm{max}}(\mathbf{X})$. We report this count for the $k$-th largest eigenvalue of the shifted operator, with $k=10$ and $k=20$; these eigenvalues correspond to the $k$-th smallest eigenvalues of the original operator. The learned preconditioners consistently reduce the iteration counts relative to the power-preconditioner baseline $P(\mathbf{X})=\mathbf{X}^d$, demonstrating their effectiveness in accelerating Krylov–Schur iterations under tight memory and communication budgets.

\begin{figure}[!htb]
    \centering

    \begin{subfigure}{\linewidth}
        \centering
        \includegraphics[width=0.35\textwidth]{figures/eigen/legend.pdf}
    \end{subfigure}

    \vspace{4pt}

    \begin{subfigure}[t]{\linewidth}
        \centering

        \begin{minipage}[t]{0.32\linewidth}
            \centering
            \includegraphics[width=\textwidth]{figures/eigen/eigconv_i10_matrixSiO2_k10_nn_vs_noprecond_y_True.pdf}

            \vspace{-2pt}
            {\scriptsize $\lambda_{10}$ ($k$ = 10)}
        \end{minipage}
        \hfill
        \begin{minipage}[t]{0.32\linewidth}
            \centering
            \includegraphics[width=\textwidth]{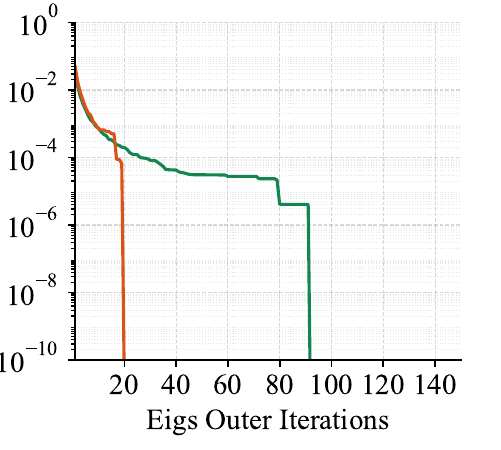}
            \vspace{-2pt}
            {\scriptsize $\lambda_{10}$ ($k$ = 20)}
        \end{minipage}
        \hfill
        \begin{minipage}[t]{0.32\linewidth}
            \centering
            \includegraphics[width=\textwidth]{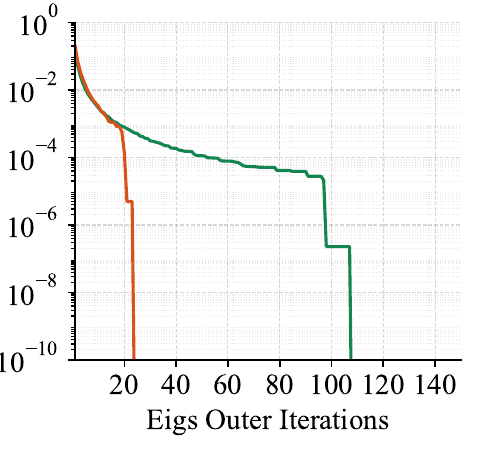}

            \vspace{-2pt}
            {\scriptsize $\lambda_{20}$ ($k$ = 20)}
        \end{minipage}

        \caption{\texttt{SiO2}}
        \label{fig:eigs_results_SiO2}
    \end{subfigure}

    \vspace{6pt}

    \begin{subfigure}[t]{\linewidth}
        \centering

        \begin{minipage}[t]{0.32\linewidth}
            \centering
            \includegraphics[width=\textwidth]{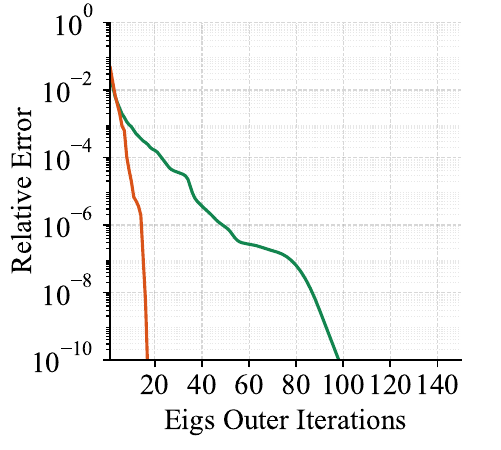}

            \vspace{-2pt}
            {\scriptsize $\lambda_{5}$ ($k$ = 10)}
        \end{minipage}
        \hfill
        \begin{minipage}[t]{0.32\linewidth}
            \centering
            \includegraphics[width=\textwidth]{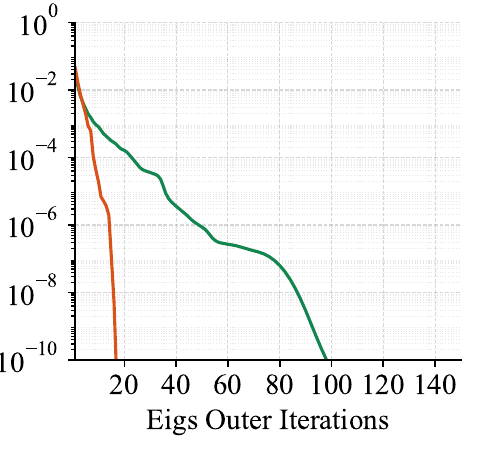}

            \vspace{-2pt}
            {\scriptsize $\lambda_{10}$ ($k$ = 10)}
        \end{minipage}
        \hfill
        \begin{minipage}[t]{0.32\linewidth}
            \centering
            \includegraphics[width=\textwidth]{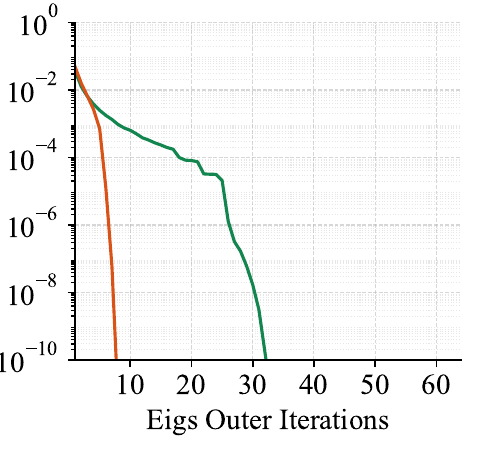}

            \vspace{-2pt}
            {\scriptsize $\lambda_{10}$ ($k$ = 20)}
        \end{minipage}

        \caption{\texttt{thermal2}}
        \label{fig:eigs_results_thermal2}
    \end{subfigure}

    \caption{Convergence of eigenvalue estimates versus Krylov--Schur outer iterations for \texttt{SiO2} and \texttt{thermal2} matrices. Parameter $k$ denotes the total number of smallest-magnitude eigenpairs computed with subspace dimension~$l=4k$.}
    \label{fig:eigs_convergence_complementary}
\end{figure}

In addition to the results in Section~\ref{sec:eigval_problems}, we present complementary experiments on convergence of preconditioned Krylov--Schur with varying numbers of requested eigenvalues $k$ and corresponding subspace dimensions $l$. Figure~\ref{fig:eigs_convergence_complementary} shows results for the \texttt{SiO2} and \texttt{thermal2} matrices with targets $k = 10$ and $k = 20$.

In both cases, the preconditioners obtained with \framework engine require substantially fewer Krylov--Schur iterations to reach target errors, indicating that the learned preconditioning method adapts and remains effective for different choices of $k$ and $l$.

\subsection{Robustness of Discovered Algorithms to Spectral Input Perturbations. }
 To evaluate the robustness of the learned \framework engine for linear systems, we vary the quality of the spectral probe by changing the number of Lanczos iterations. As shown in Figure~\ref{fig:convergence_counts}, the number of CG iterations required to reach a residual error of $10^{-10}$ remains stable, indicating that the discovered algorithm is robust.

\subsection{Comparison with Chebyshev polynomial preconditioners.}\label{sec:cheb_comparison}
We next compare \framework preconditioners with classical Chebyshev polynomial preconditioners; see, e.g.,~\cite{saad1985practical,barrett1994templates,saad2003iter,saad2011eig}. The Chebyshev preconditioners are built from an estimate of the spectral interval $[a,b]$ to be controlled. We use the affine map
\[
\theta(\lambda)
=
\frac{2\lambda - (a+b)}{b-a},
\qquad
\lambda \in [a,b],
\]
which maps $[a,b]$ to $[-1,1]$, and let $T_d$ denote the degree-$d$ Chebyshev polynomial on $[-1,1]$, defined by the recurrence

\[
T_0(t)=1,\qquad
T_1(t)=t,\qquad
T_{j+1}(t)=2tT_j(t)-T_{j-1}(t).
\]

For linear systems, $[a,b]$ is chosen to contain the spectrum of the spd operator. We define the degree-$d$ Chebyshev residual polynomial
$
R_d(\lambda)
={T_d(\theta(\lambda))}/{T_d(\theta(0))}
$
and set
\[
P_d(\lambda)
=
\frac{1-R_d(\lambda)}{\lambda}.
\]
Then $\lambda P_d(\lambda)=1-R_d(\lambda)$, where $R_d$ is the scaled Chebyshev polynomial solving 
\[
\min_{\substack{r \in \mathcal{P}_d\\ r(0)=1}}
\max_{\lambda \in [a,b]} |r(\lambda)|.
\]
Equivalently, $P_d$ solves
\[
\min_{q \in \mathcal{P}_{d-1}}
\max_{\lambda \in [a,b]}
\bigl|1-\lambda q(\lambda)\bigr|,
\]
and therefore provides the degree-$(d-1)$ minimax Chebyshev approximation to $1/\lambda$ on $[a,b]$. 


We implement the actions $\mathbf{z}\mapsto P_d(\mathbf{X})\mathbf{z}$ using the low storage, and inner-product free recurrence from~\citet[Algorithm~2]{bergamaschi2021parallel}. Table~\ref{tab:realistic_matrix_results_with_cheb} compares the number of CG iterations needed to reach a residual tolerance of $10^{-10}$ when the Chebyshev method uses the same extremal eigenvalue estimates as the neural engine, obtained from either 50 or 200 Lanczos steps. In most cases, \framework preconditioners match or outperform Chebyshev preconditioning, often by a large margin, and are more robust to changes in probe quality.

\begin{table*}[!t]
\centering
\small
\caption{Comparison of \framework and Chebyshev polynomial accelerators on the SuiteSparse matrices listed in Table~\ref{tab:test_matrix_details_a}. Chebyshev accelerators use the same estimated extremal eigenvalues provided to the neural network engine. We report CG iteration counts required to reach residual tolerance $10^{-10}$, using preconditioners constructed from 50- or 200-step Lanczos spectral probes.}
\label{tab:realistic_matrix_results_with_cheb}
\setlength{\tabcolsep}{6pt}
\renewcommand{\arraystretch}{1.10}

\begin{subtable}{\textwidth}
\centering
\setlength{\tabcolsep}{4pt}

\resizebox{\textwidth}{!}{%
\begin{tabular}{@{} l r r r r @{}}
\toprule
\textbf{Matrix $\mathbf{X}$} &
\makecell{\textbf{numIters}\\(\framework, probe=200)} &
\makecell{\textbf{numIters}\\(\framework, probe=50)} &
\makecell{\textbf{numIters}\\(Cheb, probe=200)} &
\makecell{\textbf{numIters}\\(Cheb, probe=50)}  \\
\midrule
\texttt{laplacian}         & 29 $\pm$ 0   & 27 $\pm$ 0   & 48 $\pm$ 0    & 39 $\pm$ 5 \\
\texttt{Dubcova2}          & 36 $\pm$ 0   & 35 $\pm$ 0   & 67 $\pm$ 1    & 71 $\pm$ 24 \\
\texttt{thermal1}          & 154 $\pm$ 1  & 162 $\pm$ 2  & 236 $\pm$ 40  & 145 $\pm$ 1 \\
\texttt{FEM\_3D\_thermal2} & 7 $\pm$ 0    & 6 $\pm$ 0    & 5 $\pm$ 0     & 5 $\pm$ 0 \\
\texttt{G2\_circuit}       & 205 $\pm$ 2  & 219 $\pm$ 1  & 288 $\pm$ 54  & 331 $\pm$ 194 \\
\texttt{shipsec5}          & 421 $\pm$ 2  & 442 $\pm$ 1  & 402 $\pm$ 2   & 397 $\pm$ 9 \\
\texttt{parabolic\_fem}    & 317 $\pm$ 6  & 337 $\pm$ 7  & 408 $\pm$ 146 & 830 $\pm$ 184 \\
\texttt{Fault\_639}        & 800 $\pm$ 13 & 890 $\pm$ 12 & 761 $\pm$ 12  & 938 $\pm$ 29 \\
\texttt{Emilia\_923}       & 1097 $\pm$ 5 & 1148 $\pm$ 6 & 1030 $\pm$ 5  & 1034 $\pm$ 4 \\
\texttt{thermal2}          & 592 $\pm$ 3  & 625 $\pm$ 2  & 557 $\pm$ 2   & 557 $\pm$ 3 \\
\texttt{G3\_circuit}       & 397 $\pm$ 4  & 436 $\pm$ 8  & 585 $\pm$ 176 & 387 $\pm$ 8 \\
\bottomrule
\end{tabular}%
}
\end{subtable}

\end{table*}

\begin{figure}[!htb]
    \centering
    \begin{subfigure}{1.0\linewidth}
        \includegraphics[width=\textwidth]{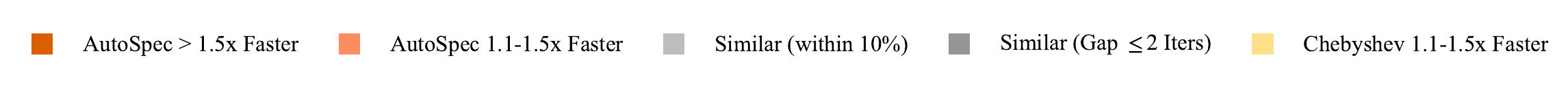}
    \end{subfigure}
    
    \begin{subfigure}[t]{0.25\linewidth}
        \includegraphics[width=\textwidth]{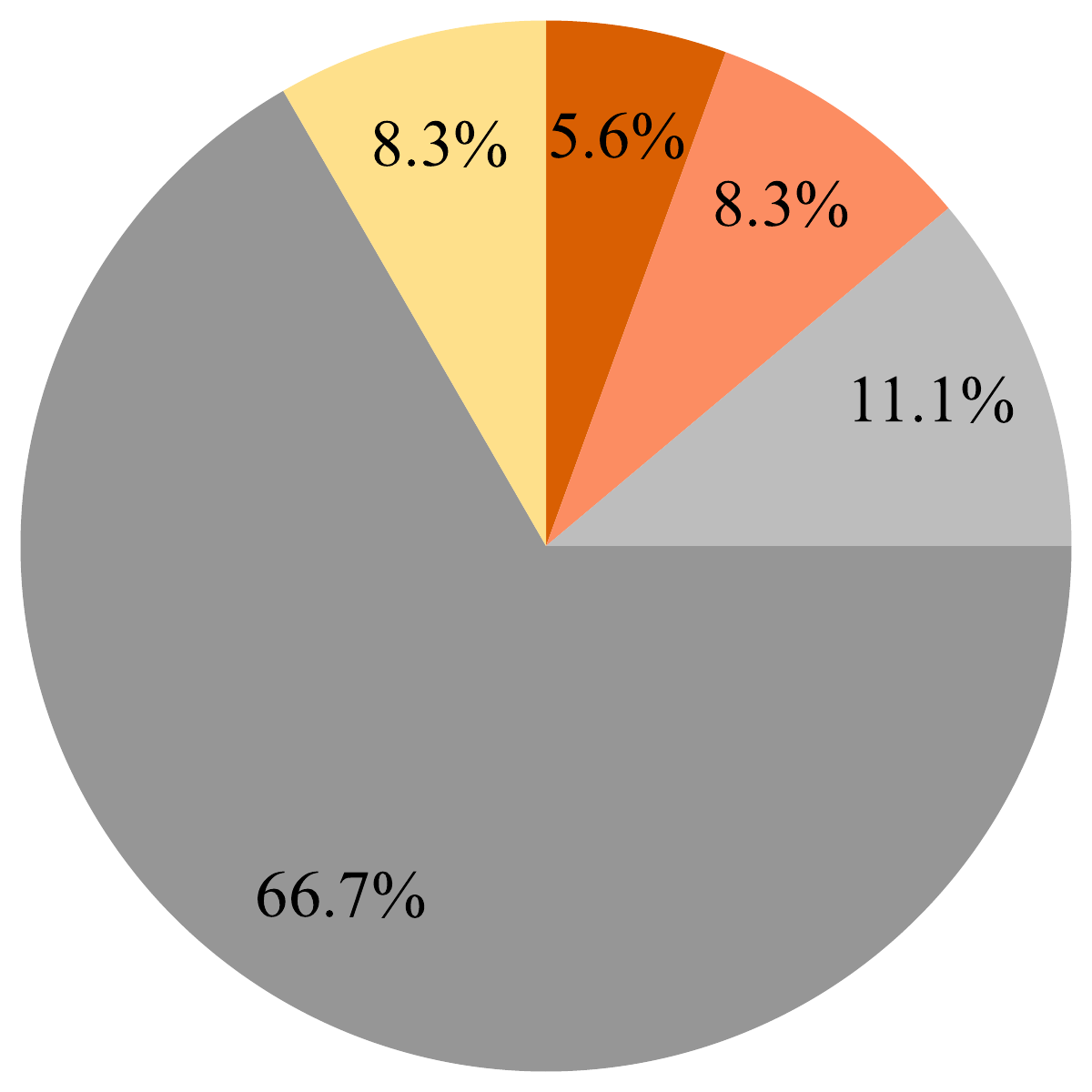}
        \caption{50 Probe Iterations}
    \end{subfigure}
    \hspace{1cm}
    \begin{subfigure}[t]{0.25\linewidth}
        \includegraphics[width=\textwidth]{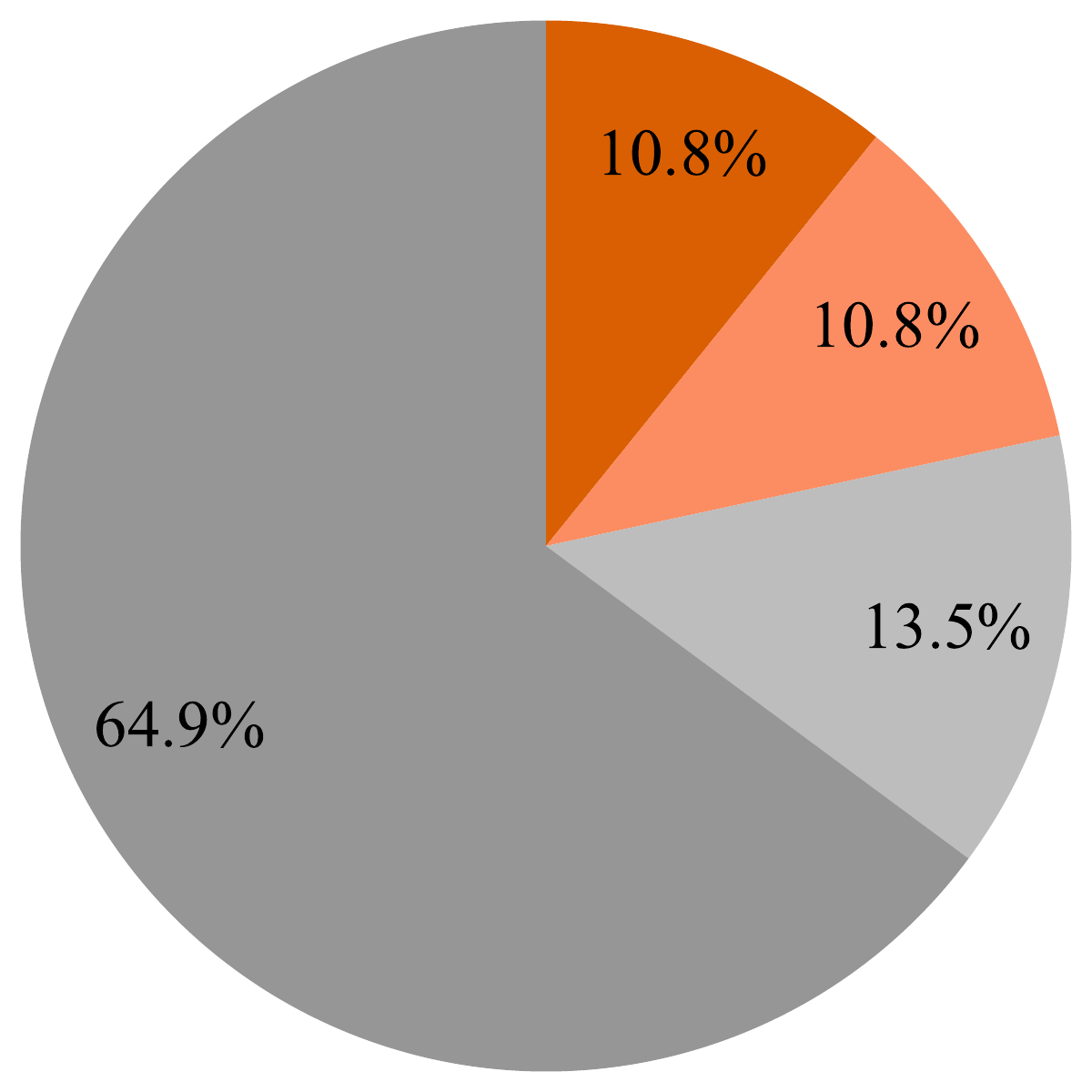}
        \caption{200 Probe Iterations}
    \end{subfigure}

    \caption{Comparing CG iteration counts over 5 seeds of AutoSpec and Chebyshev polynomial preconditioning, on all spd SuiteSparse matrices with sizes between $10^4$ and $10^5$ and at most $2 \times 10^6$ nonzeros.}
    \label{fig:autospec_vs_chebyshev} 
\end{figure}

To further strengthen our claims, we evaluate all spd matrices in the SuiteSparse Matrix Collection with sizes between $10^4$ and $10^5$ and at most $2\times 10^6$ nonzeros. Figure~\ref{fig:autospec_vs_chebyshev} summarizes the resulting iteration counts.

\subsection{Comparison with Chebyshev Acceleration under Different First-Level Preconditioners}\label{sec:second_level_comparison}

\begin{figure}[!htb]
    \centering

    \begin{subfigure}{\linewidth}
        \centering
        \includegraphics[width=0.9\textwidth]{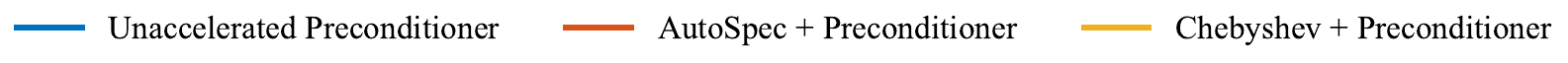}
    \end{subfigure}

    \vspace{4pt}

    \begin{subfigure}[t]{\linewidth}
        \centering

        \begin{minipage}[t]{0.32\linewidth}
            \centering
            \includegraphics[width=\textwidth]{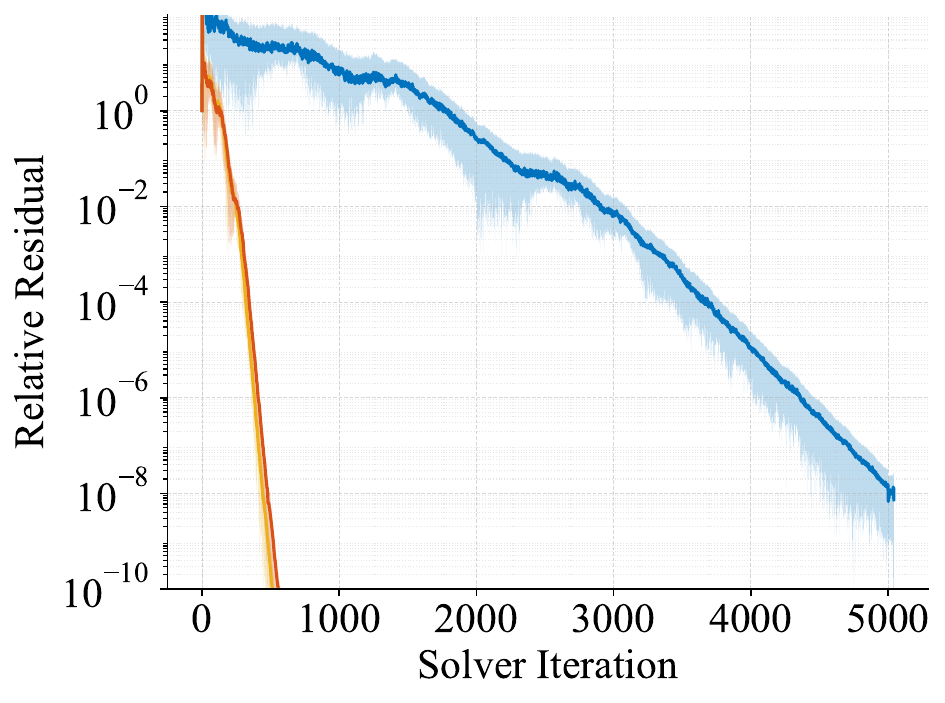}

            \vspace{-2pt}
            {\scriptsize Jacobi}
        \end{minipage}
        \hfill
        \begin{minipage}[t]{0.32\linewidth}
            \centering
            \includegraphics[width=\textwidth]{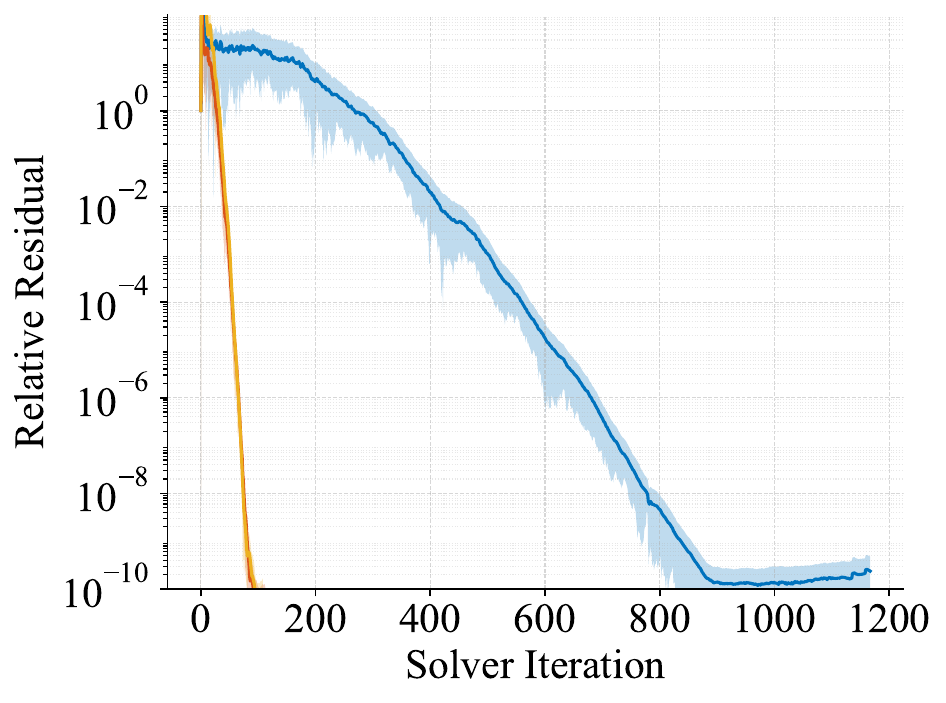}

            \vspace{-2pt}
            {\scriptsize AMG}
        \end{minipage}
        \hfill
        \begin{minipage}[t]{0.32\linewidth}
            \centering
            \includegraphics[width=\textwidth]{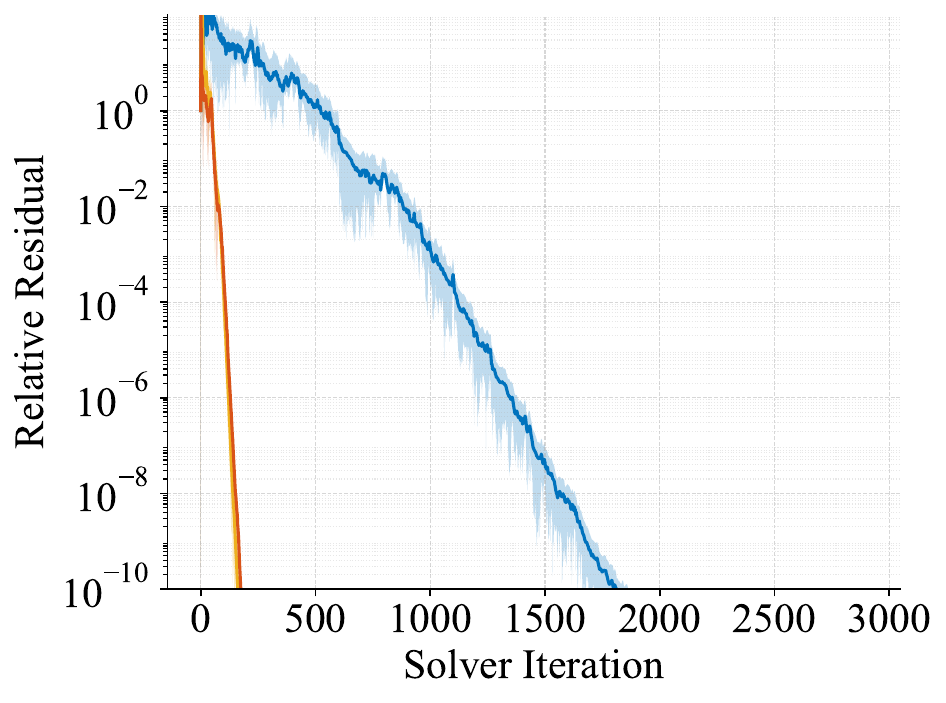}

            \vspace{-2pt}
            {\scriptsize iChol(0)}
        \end{minipage}

        \caption{\texttt{Shipsec5} with 50 Probe Iterations}
        \label{fig:second_level_shipsec_50}
    \end{subfigure}

    \vspace{6pt}

    \begin{subfigure}[t]{\linewidth}
        \centering

        \begin{minipage}[t]{0.32\linewidth}
            \centering
            \includegraphics[width=\textwidth]{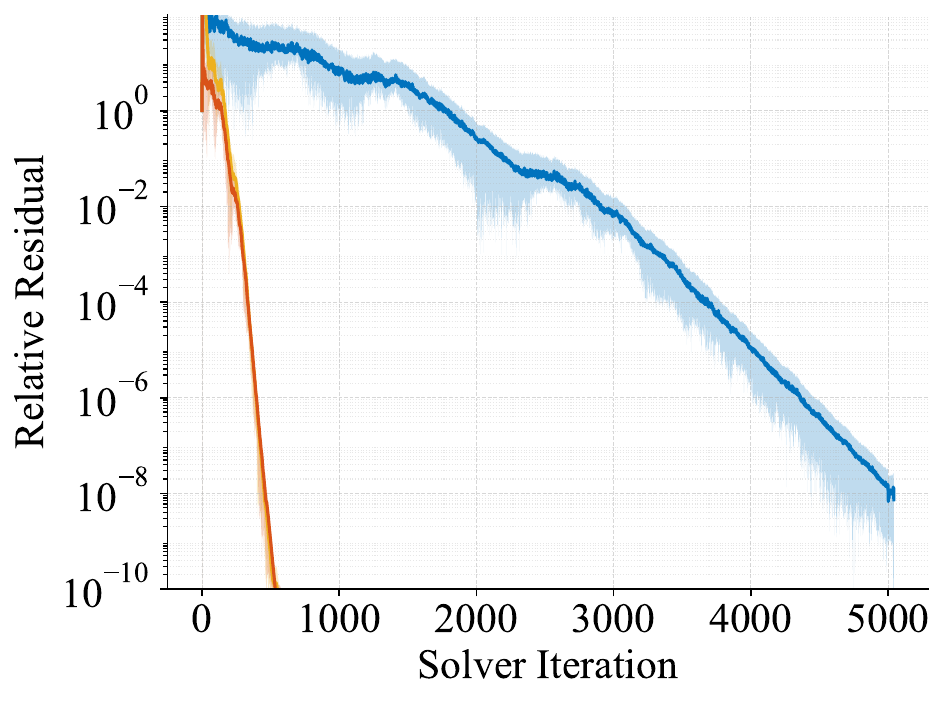}

            \vspace{-2pt}
            {\scriptsize Jacobi}
        \end{minipage}
        \hfill
        \begin{minipage}[t]{0.32\linewidth}
            \centering
            \includegraphics[width=\textwidth]{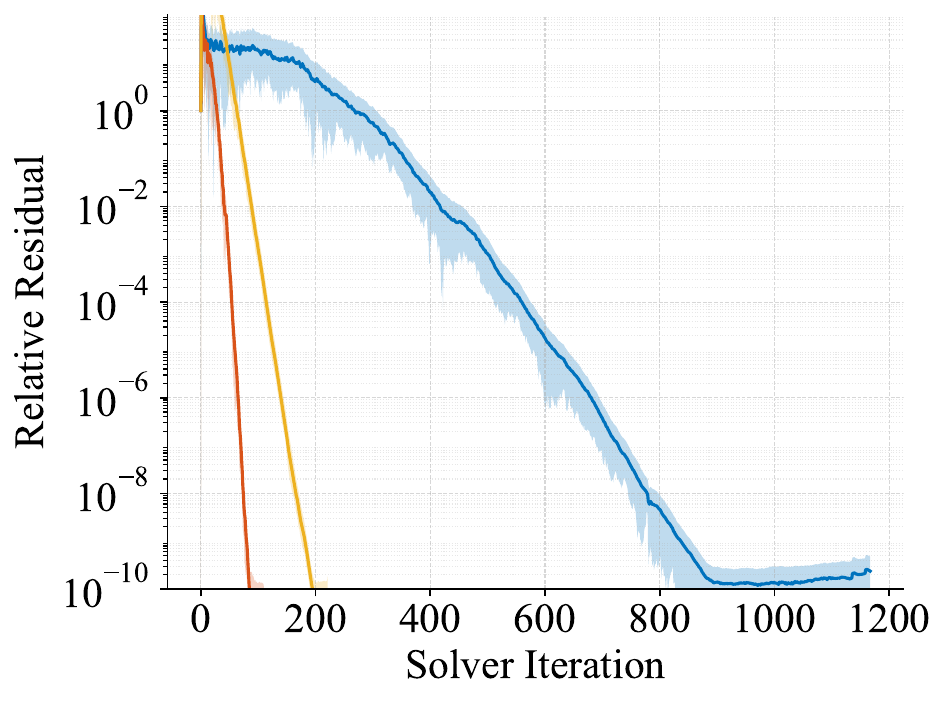}

            \vspace{-2pt}
            {\scriptsize AMG}
        \end{minipage}
        \hfill
        \begin{minipage}[t]{0.32\linewidth}
            \centering
            \includegraphics[width=\textwidth]{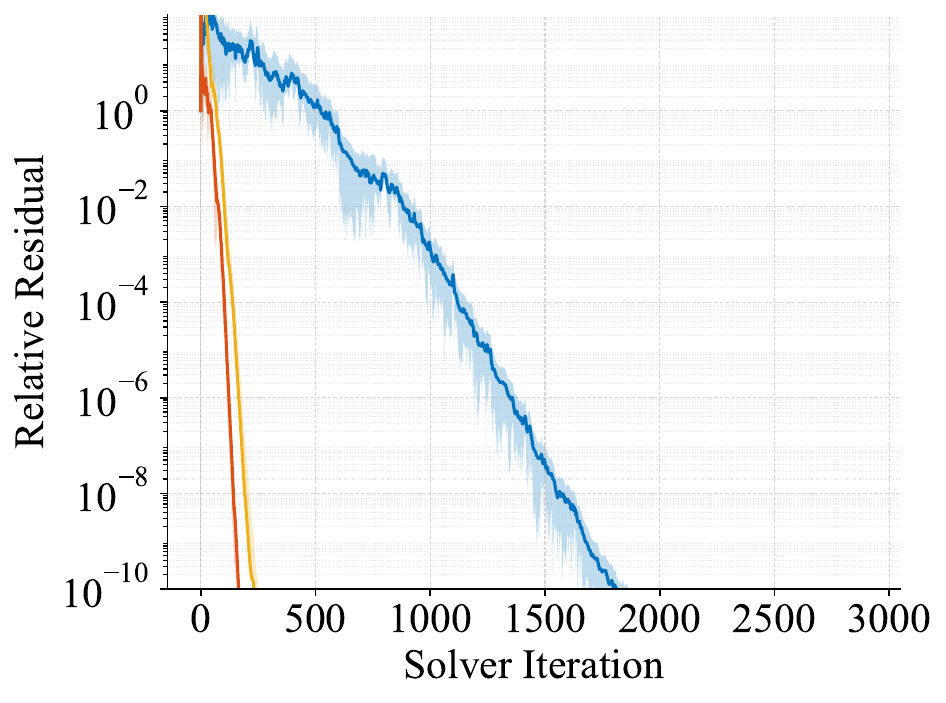}

            \vspace{-2pt}
            {\scriptsize iChol(0)}
        \end{minipage}

        \caption{\texttt{Shipsec5} with 200 Probe Iterations}
        \label{fig:second_level_shipsec_200}
    \end{subfigure}

    \caption{AutoSpec and Chebyshev as second-level accelerators for CG, preconditioned using Jacobi, AMG, and iChol(0) on the \texttt{Shipsec5} system.}
    \label{fig:second_level_shipsec}
\end{figure}

\begin{figure}[!htb]
    \centering

    \begin{subfigure}{\linewidth}
        \centering
        \includegraphics[width=0.9\textwidth]{figures/linear_systems/second_level/legend.pdf}
    \end{subfigure}

    \vspace{4pt}

    \begin{subfigure}[t]{\linewidth}
        \centering

        \begin{minipage}[t]{0.32\linewidth}
            \centering
            \includegraphics[width=\textwidth]{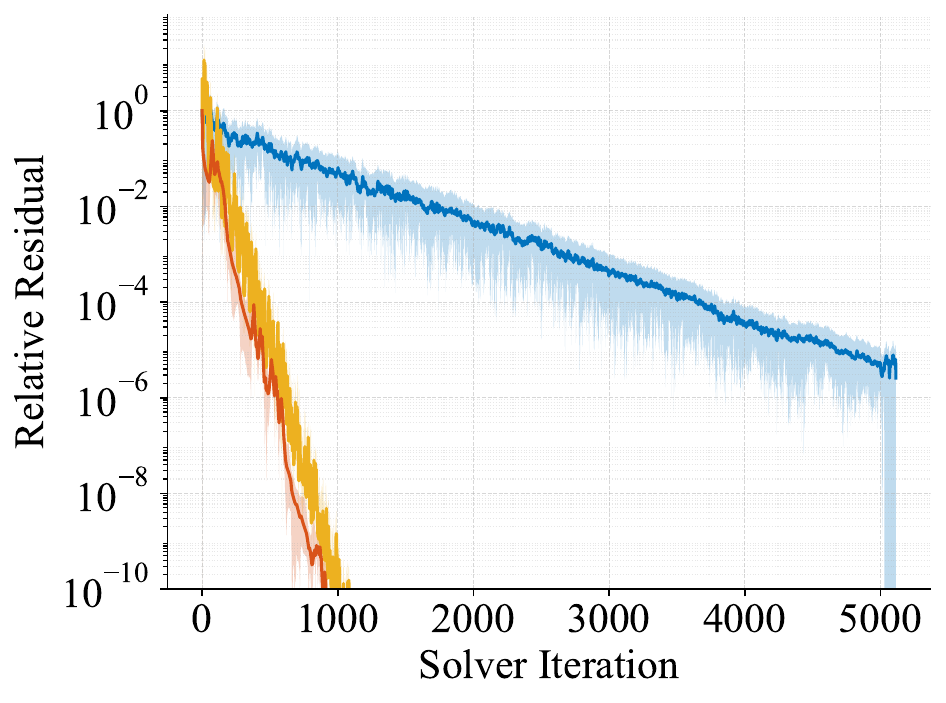}

            \vspace{-2pt}
            {\scriptsize Jacobi}
        \end{minipage}
        \hfill
        \begin{minipage}[t]{0.32\linewidth}
            \centering
            \includegraphics[width=\textwidth]{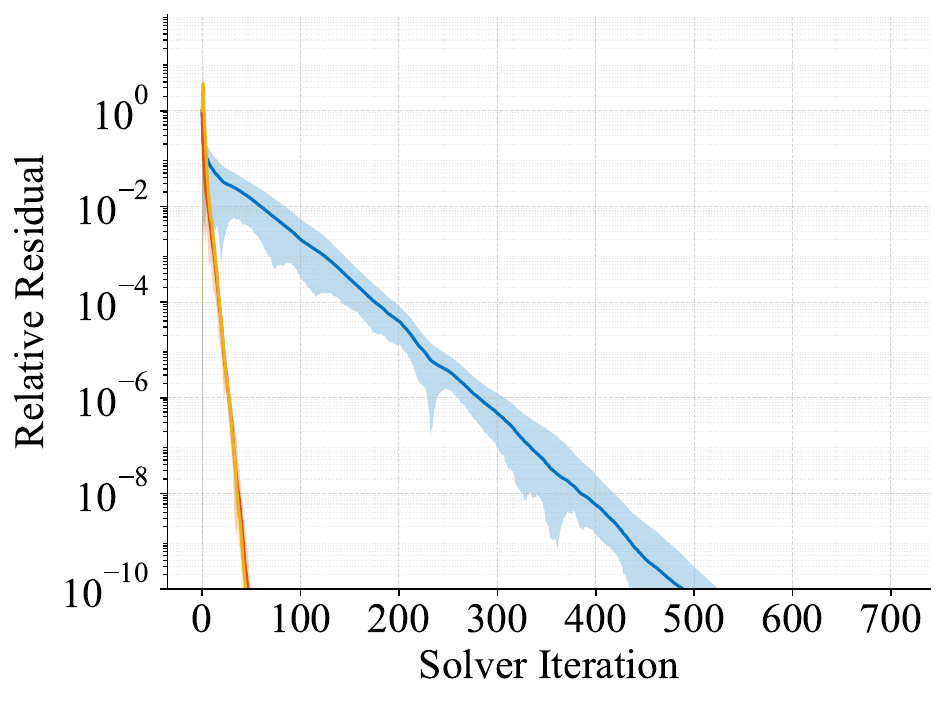}

            \vspace{-2pt}
            {\scriptsize AMG}
        \end{minipage}
        \hfill
        \begin{minipage}[t]{0.32\linewidth}
            \centering
            \includegraphics[width=\textwidth]{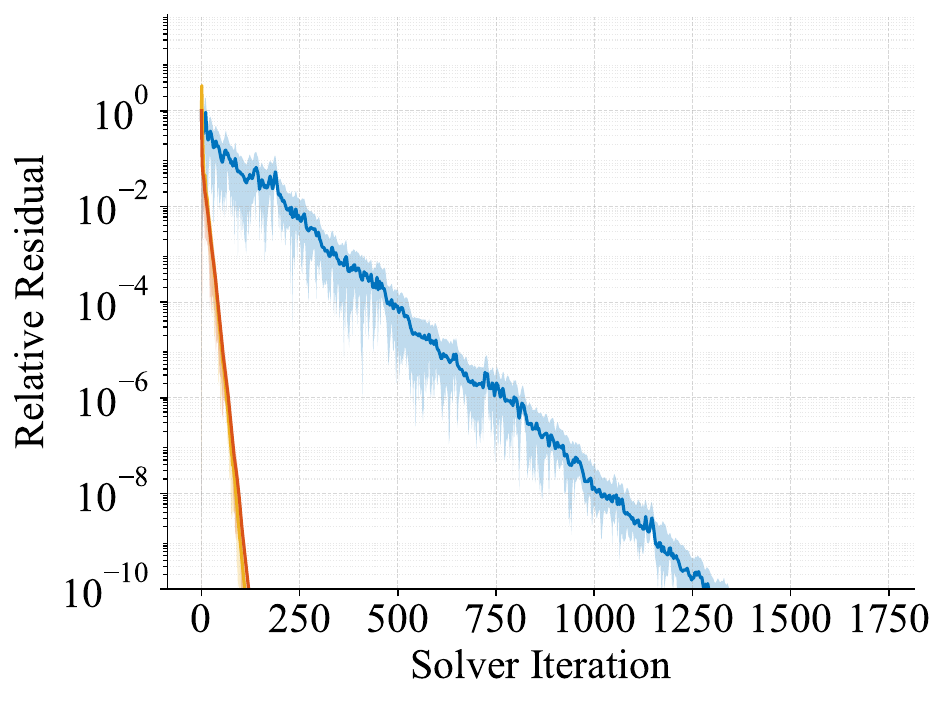}

            \vspace{-2pt}
            {\scriptsize iChol(0)}
        \end{minipage}

        \caption{\texttt{Fault\_639} with 50 Probe Iterations}
        \label{fig:second_level_fault_639_50}
    \end{subfigure}

    \vspace{6pt}

    \begin{subfigure}[t]{\linewidth}
        \centering

        \begin{minipage}[t]{0.32\linewidth}
            \centering
            \includegraphics[width=\textwidth]{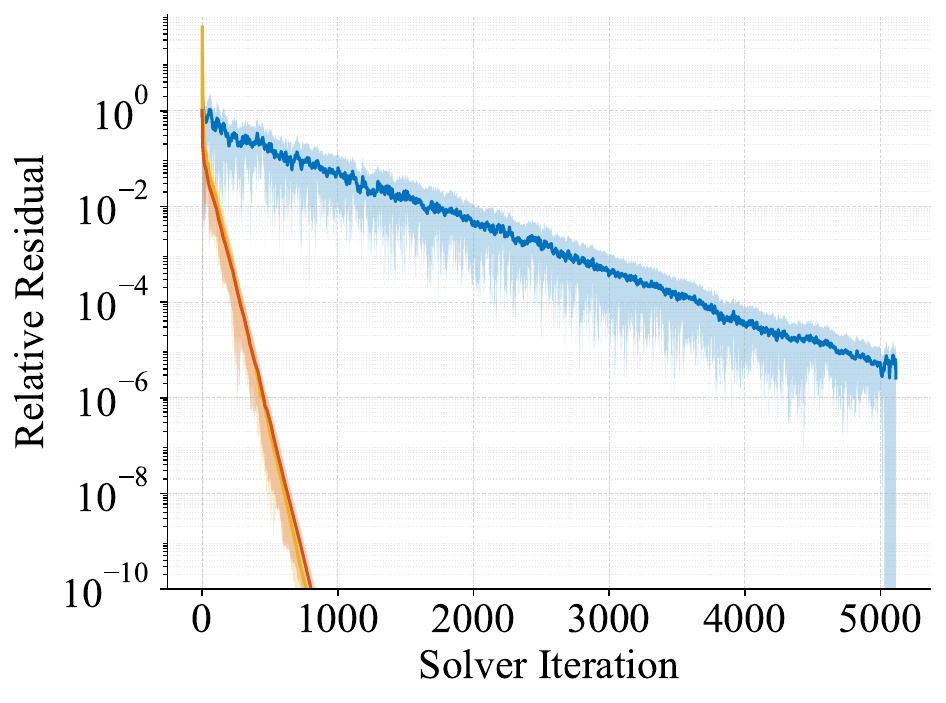}

            \vspace{-2pt}
            {\scriptsize Jacobi}
        \end{minipage}
        \hfill
        \begin{minipage}[t]{0.32\linewidth}
            \centering
            \includegraphics[width=\textwidth]{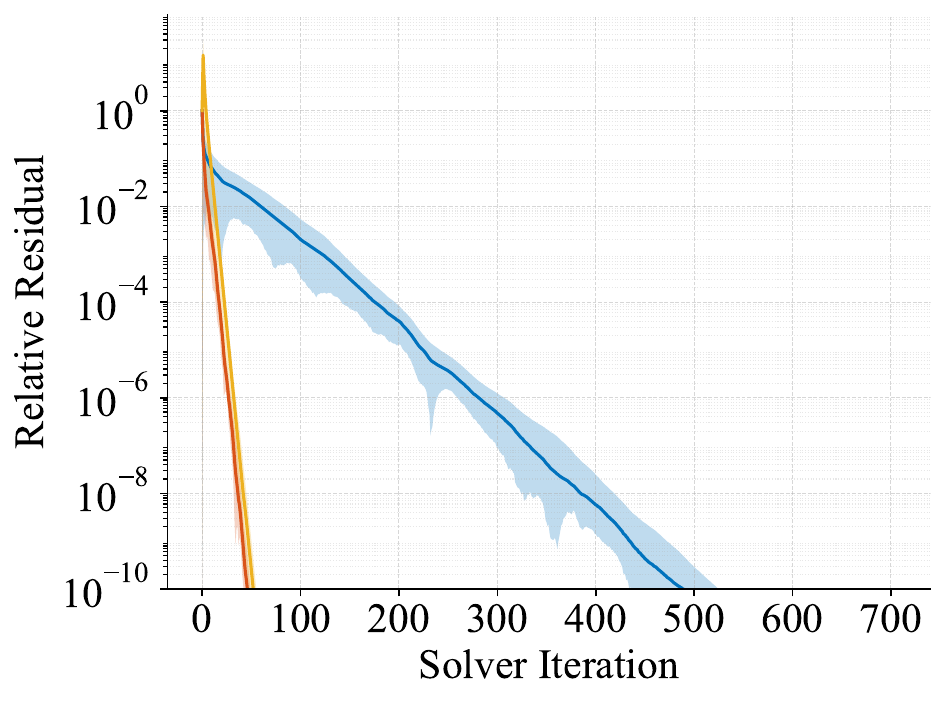}

            \vspace{-2pt}
            {\scriptsize AMG}
        \end{minipage}
        \hfill
        \begin{minipage}[t]{0.32\linewidth}
            \centering
            \includegraphics[width=\textwidth]{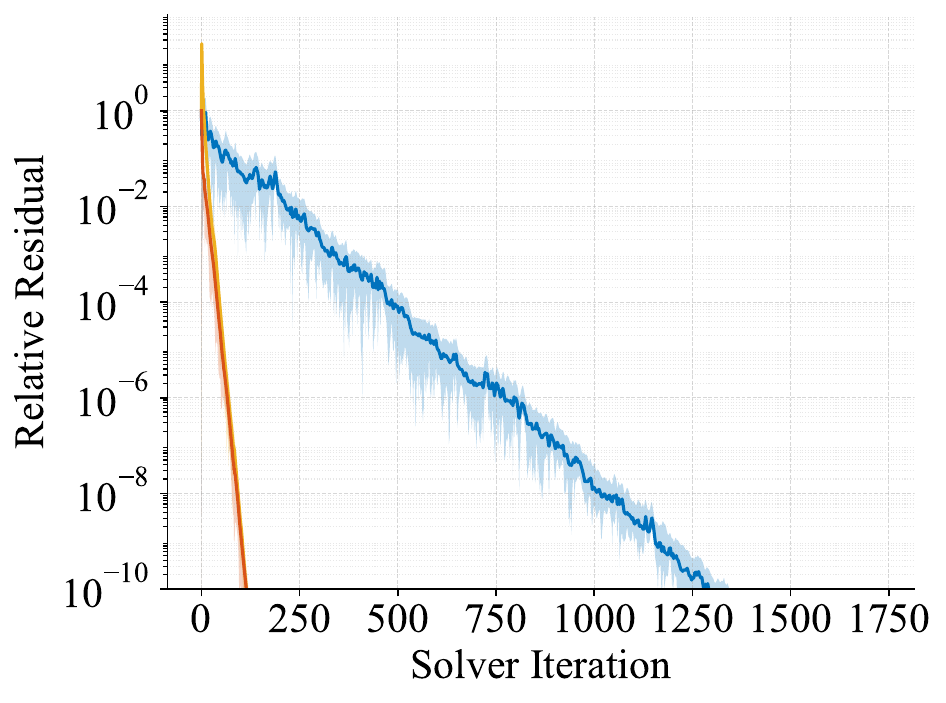}

            \vspace{-2pt}
            {\scriptsize iChol(0)}
        \end{minipage}

        \caption{\texttt{Fault\_639} with 200 Probe Iterations}
        \label{fig:second_level_fault_639_200}
    \end{subfigure}

    \caption{AutoSpec and Chebyshev as second-level accelerators for CG, preconditioned using Jacobi, AMG, and iChol(0) on the \texttt{Fault\_639} system.}
    \label{fig:second_level_fault_639}
\end{figure}

We evaluated \framework on \texttt{Shipsec5} and \texttt{Fault\_639} using Jacobi, AMG, and iChol(0) as first-level preconditioners, and compare its performance with Chebyshev acceleration under the same settings (see Figures~\ref{fig:second_level_shipsec} and~\ref{fig:second_level_fault_639} ).

Overall, our results show that \framework polynomial preconditioners provide strong and robust acceleration across diverse real-world matrices, supporting our algorithm-discovery approach. They further suggest that \framework can help automate advances in numerical linear algebra: the discovered preconditioners operate reliably using only limited spectral information, adapt to the structure of the input operator, and sometimes can outperform widely used classical methods.

\subsection{Limitations}\label{sec:limitations}
\begin{figure}[!htb]
    \centering
    \begin{subfigure}{0.7\linewidth}
        \includegraphics[width=\textwidth]{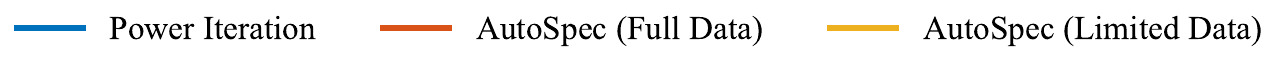}
    \end{subfigure}
    
    \begin{subfigure}[t]{0.4\linewidth}
        \includegraphics[width=\textwidth]{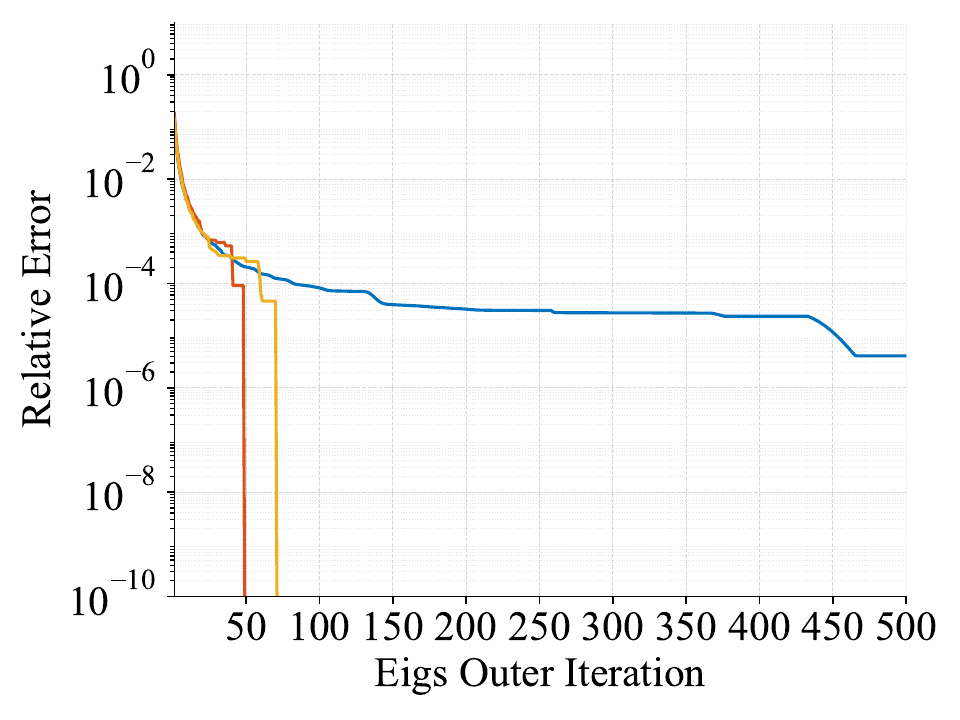}
        \caption{Convergence of the eigenvalue approximations versus the number of outer iterations of \texttt{eigs}.}
    \end{subfigure}
    \hspace{0.5cm}
    \begin{subfigure}[t]{0.4\linewidth}
        \includegraphics[width=\textwidth]{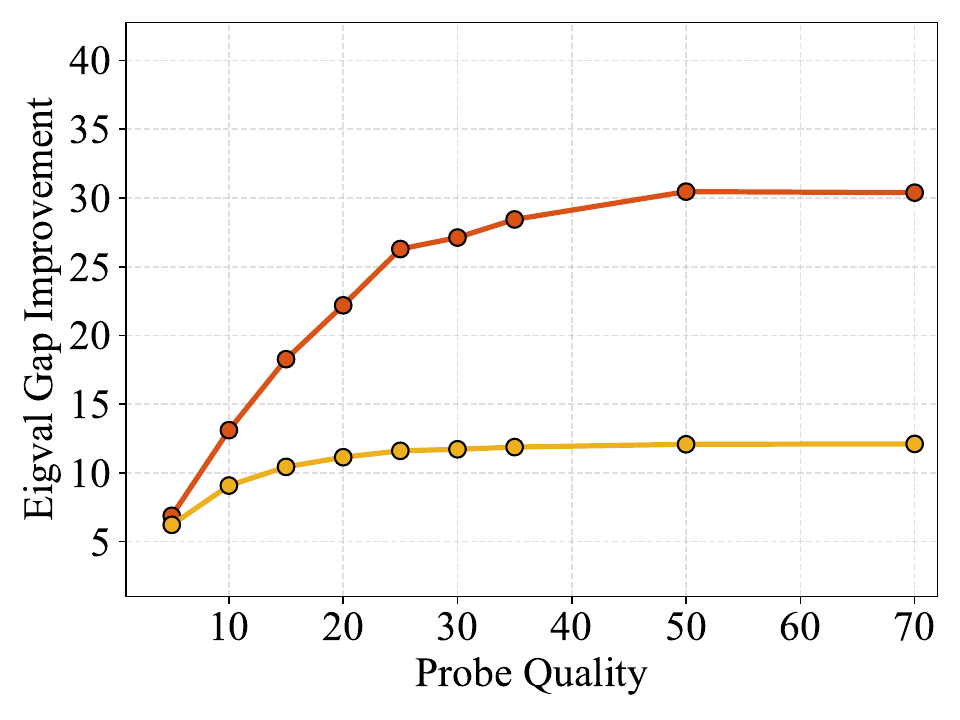}
        \caption{Improvement in the eigenvalue gap relative to the baseline filter $P(X)=X^d$.}
    \end{subfigure}

    \caption{We compare the spectral filter produced for the SiO2 matrix using the neural-engine checkpoint from the paper with that produced by a neural engine trained only on spectra exhibiting relatively large eigenvalue gaps.}
    \label{fig:failure_case}
\end{figure}

To illustrate the discussion on limitations in Section~\ref{sec:ablation}, we provide results where insufficient training leads to performance degradation. As shown in Figure~\ref{fig:failure_case}, for eigenvalue problems, if the model is trained only on limited training data containing spectra with larger eigengaps than those encountered at test time (yellow line), performance can degrade compared to training with more diverse data (red line).

\section{Ablation Studies}\label{sec:appendix_ablation}

\subsection{Role of Spectral Residual Features in Discovering Algorithms}
Given eigenvalue estimations augmented by corresponding residual features, the embedding layer ``reasons'' and exploits critical spectral information, resulting in the discovery of effective algorithms. To further analyze how the embedding layer uses the spectral residuals, we perform an ablation study by setting the residual to 0 and observing how it affects the construction of algorithms and performance on synthetic matrices.  

\begin{figure}[!htb]
    \centering
    \begin{subfigure}{0.4\linewidth}
        \includegraphics[width=\textwidth]{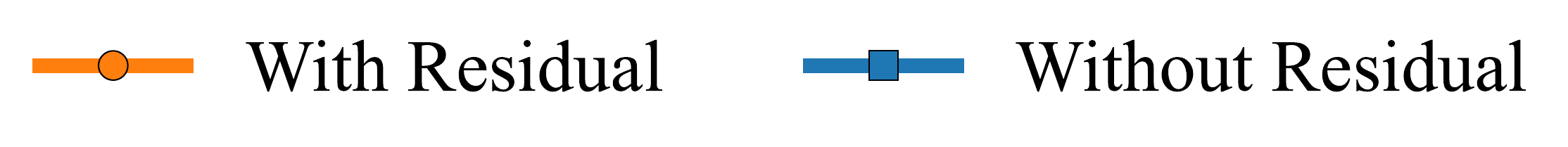}
    \end{subfigure}
    
    \begin{subfigure}[t]{0.35\linewidth}
        \includegraphics[width=\textwidth]{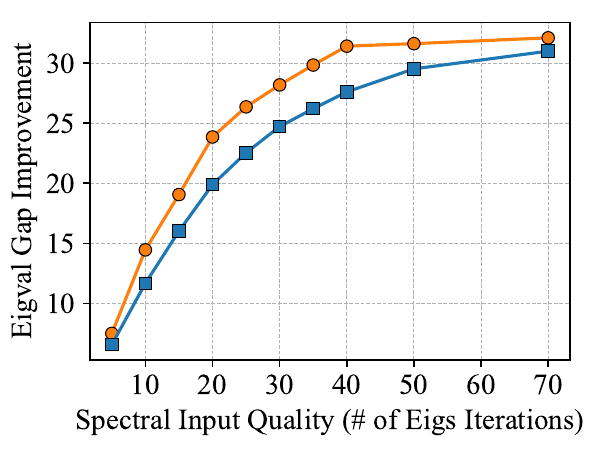}
        \caption{\texttt{SiO2}}
    \end{subfigure}
    \begin{subfigure}[t]{0.35\linewidth}
        \includegraphics[width=\textwidth]{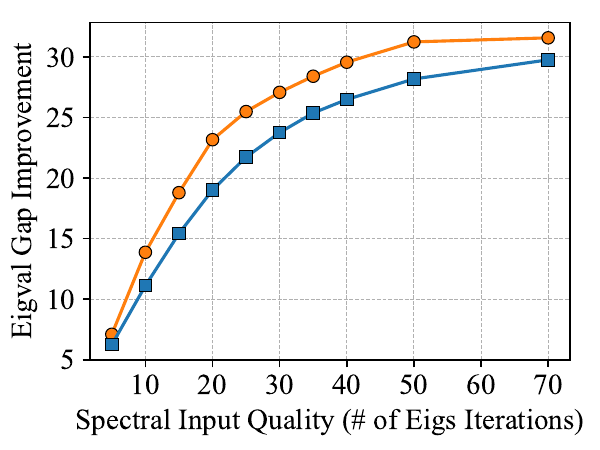}
        \caption{\texttt{thermal2}}
    \end{subfigure}

    \caption{Comparison of learned eigen preconditioners on \texttt{SiO2} and \texttt{thermal2}, generated with versus without providing estimated eigenvalue residuals as input to the neural engine. Performance is reported as the log improvement in the target-boundary eigenvalue gap relative to standard subspace iteration baseline $P(\mathbf{X}) = \mathbf{X}^d$.}
    \label{fig:ablation_zero_residual_eigs}
\end{figure}
As shown in Figure~\ref{fig:ablation_zero_residual_eigs}, incorporating residuals as input improves the effectiveness of the learned preconditioners. When the model is provided with residual features, it consistently achieves larger eigenvalue gap improvements over subspace iteration $P(\mathbf{X}) = \mathbf{X}^d$ across all iteration budgets. 
This indicates that residual features provide essential spectral-quality signals for the model to generate effective preconditioning algorithms. 
In Figure~\ref{fig:ablation_zero_residual_linsolve} (in the appendix), we present additional results on synthetic matrices, confirming that residual features play an important role in the algorithm generation of \framework.

\subsection{Robustness of Discovered Algorithms to Spectral Input Perturbations}
To assess the robustness of the discovered linear solver accelerator, we perturb the spectral probe by varying the Lanczos iteration counts. 
Figure~\ref{fig:convergence_counts} (in the appendix) shows that once probe quality is sufficiently high (above 75 Lanczos iterations), the CG iteration count to reach $10^{-10}$ error stabilizes, indicating robustness to spectral-input perturbations.

\begin{figure}[!htb]
    \centering
    \begin{subfigure}[t]{0.293\linewidth}
        \includegraphics[width=\textwidth]{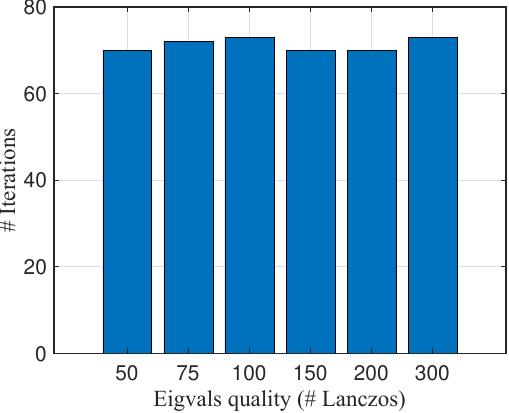}
        \caption{\texttt{G2\_circuit}}
    \end{subfigure}
    \begin{subfigure}[t]{0.3\linewidth}
        \includegraphics[width=\textwidth]{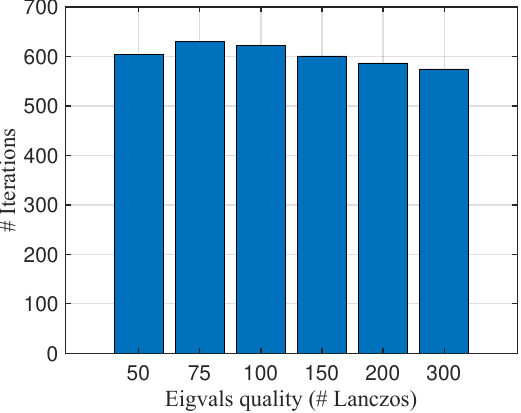}
        \caption{\texttt{thermal2}}
    \end{subfigure}

    \caption{CG iterations to reach $10^{-10}$ error vs Lanczos iterations used for NN input spectral probe.}
    \label{fig:convergence_counts} 
\end{figure}

\section{Complementary Results on Synthetic Matrices}

\subsection{Eigenvalue Problems}\label{appendix:synthetic_results_eigenproblems}

\begin{figure}[!htb]
    \centering
    \begin{subfigure}[t]{0.4\linewidth}
        \includegraphics[width=\textwidth]{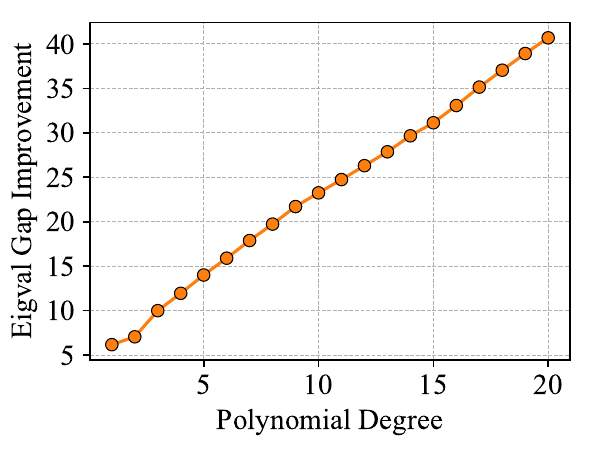}
        \caption{Top-$k$ degree polynomial by backbone model.}
        \label{fig:eigval_gap_vs_degree}
    \end{subfigure}
    \begin{subfigure}[t]{0.45\linewidth}
        \includegraphics[width=\textwidth]{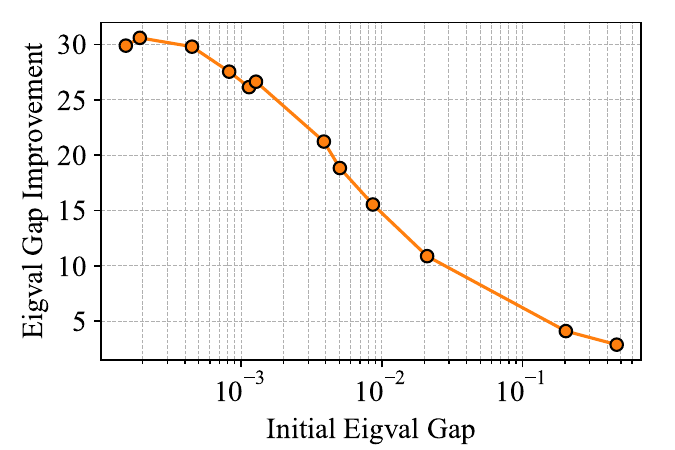}
        \caption{Eigenvalue gap improvement of learned polynomial.}
        \label{fig:eigval_gap_embedding}
    \end{subfigure}

    \caption{Eigenvalue gap improvement of learned polynomial preconditioning algorithms, compared to the standard subspace iteration preconditioner. \emph{Left}: degree-$k$ polynomial obtained by the first $k$ layer of the backbone model after pre-training. \emph{Right}: learned algorithm (after post-training) using spectral probe input obtained with 100 subspace iterations.}
    \label{fig:eigs_backbone_synthetic} 
\end{figure}

\noindent\paragraph{Backbone Model on Synthetic Matrices. }Figure~\ref{fig:eigval_gap_vs_degree} shows the eigenvalue gap improvement of polynomial preconditioning algorithms generated by the first $k$ layers of the backbone model after pre-training ($1\le k\le20$), compared to the standard subspace iteration preconditioner. We can see that the polynomial generated by any first $k$ layers of the backbone model brings significant improvement of the eigenvalue gap after preconditioning. This shows the pre-training of discovering preconditioning algorithms for eigenvalue problems achieves ``length-generalization,'' that any of the first $k$ layers can produce an effective algorithm. This enables us to extend the model to learn higher-degree preconditioning algorithms.

\noindent\paragraph{Embedding Model on Synthetic Matrices.}Figure~\ref{fig:eigval_gap_embedding} shows the eigenvalue gap improvement of the model after pre-training, given the spectral probe input obtained with 100 subspace iterations. We can see that the model can construct effective algorithms for operators with a wide range of initial eigenvalue gaps. Furthermore, the learned algorithms bring larger improvements as the initial eigenvalue gap becomes smaller. This shows the effectiveness of our model to be applied to realistic settings, where the initial eigenvalues have small gaps.

\subsection{Preconditioned Linear Systems}\label{appendix:synthetic_results_linsolve}

\noindent\paragraph{Evaluation on synthetic diagonal matrices.}
Following the results in Section~\ref{sec:linsolve_results}, we present evaluation results on synthetic diagonal matrices $\mathbf{X} = \operatorname{diag}(\boldsymbol{\lambda})$ with different initial condition numbers, and we evaluate the normalized residual of learned preconditioning algorithm $r_{\text{nn}}$ defined in~\cref{eq:obj_linsolve_nn} compared to the residual of standard Richardson iteration defined in~\cref{eq:obj_linsolve_richardson} in logarithm scale. This effectively measures the improvement of the condition number of the preconditioned operator.  

\begin{figure}[!htb]
    \centering
    \begin{subfigure}[t]{0.245\linewidth}
        \includegraphics[width=\textwidth]{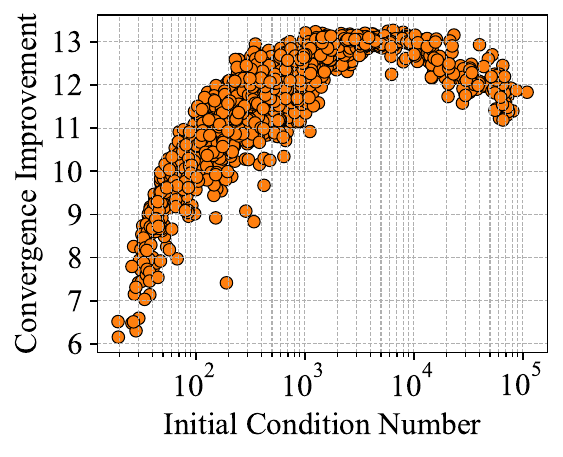}
        \caption{1 Subspace Iterations}
    \end{subfigure}
    \begin{subfigure}[t]{0.23\linewidth}
        \includegraphics[width=\textwidth]{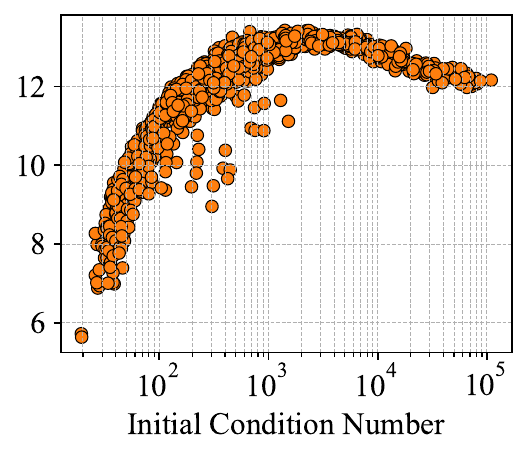}
        \caption{5 Subspace Iterations}
    \end{subfigure}
    \begin{subfigure}[t]{0.23\linewidth}
        \includegraphics[width=\textwidth]{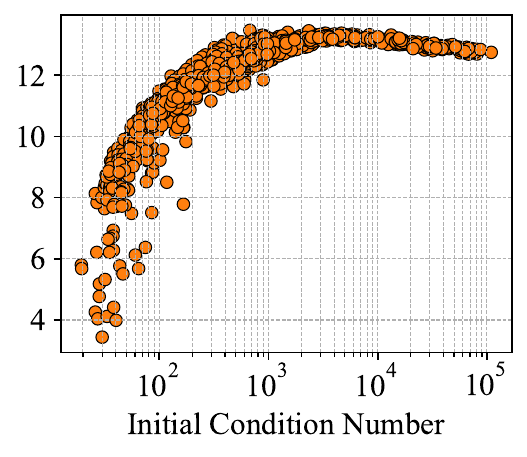}
        \caption{10 Subspace Iterations}
    \end{subfigure}
    \begin{subfigure}[t]{0.23\linewidth}
        \includegraphics[width=\textwidth]{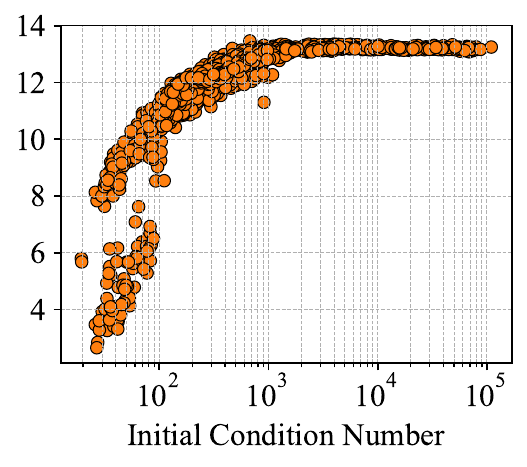}
        \caption{50 Subspace Iterations}
    \end{subfigure}

    \caption{Condition number improvement of the preconditioning algorithm generated by the neural network for synthetic linear systems with different initial condition numbers, given different spectral probe quality. The input spectral probe is obtained using subspace iteration; more iterations result in a higher-quality spectral probe.}
    \label{fig:synthetic_linsolve_niter} 
\end{figure}

As shown in Figure~\ref{fig:synthetic_linsolve_niter}, we plot the condition number improvement of each synthetic matrix, corresponding to the initial condition number of the operator. We can see that the learned algorithm consistently achieves improvement over Richardson iteration, where the algorithm has larger improvement when the initial condition number is larger. Furthermore, we observe that as the spectral probe input to the neural network becomes more accurate (by using more subspace iterations), the performance improvement becomes higher and more robust. This result shows the effectiveness and robustness of the neural network engine in generating effective preconditioning algorithms for operators of a wide range of condition numbers.

\noindent\paragraph{Roles of Spectral Residual Features in Discovering Algorithms.}
Following the results in Section~\ref{sec:ablation}, we provide complementary results on synthetic matrices, demonstrating the role of residual features in the learned algorithm.
\begin{figure}[!htb]
    \centering
    \begin{subfigure}{0.4\linewidth}
        \includegraphics[width=\textwidth]{figures/ablation/zero_residual/legend.pdf}
    \end{subfigure}
    
    \begin{subfigure}[t]{0.3\linewidth}
        \includegraphics[width=\textwidth]{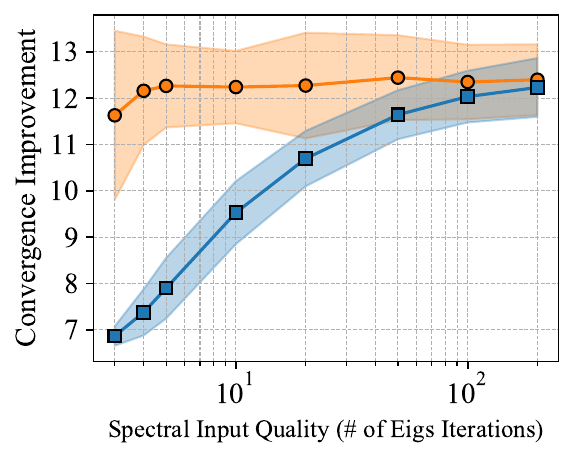}
        \caption{dim = 500, $\boldsymbol{\kappa}\in[1e2, 5e3]$}
    \end{subfigure}
    \begin{subfigure}[t]{0.3\linewidth}
        \includegraphics[width=\textwidth]{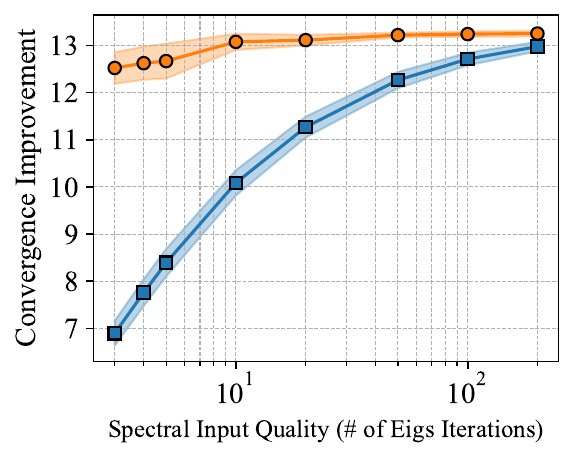}
        \caption{dim = 2000, $\boldsymbol{\kappa}\in[2e3, 1e5]$}
    \end{subfigure}

    \caption{Comparison of generated algorithms using eigenvalue estimation residual features (Ritz value residual norms) versus without residuals, under different spectral probe input quality (number of subspace iterations). Performance is measured by the ronvergence rate improvement relative to Richardson iteration in logarithms, on synthetic matrices with varying condition numbers $\kappa$.}
    \label{fig:ablation_zero_residual_linsolve}
\end{figure}
As shown in Figure~\ref{fig:ablation_zero_residual_linsolve}, we can see that the learned algorithm without residual features achieves significantly worse convergence improvement compared to using residual features, and the performance gap becomes larger as the spectral probe input quality becomes worse. Conversely, as the spectral input quality becomes better, the residual features becomes smaller, making the generated algorithm highly effective, regardless of the use of residuals.

\subsection{Approximating Matrix Functions}
\label{appendix:synthetic_results_matfunc}

\begin{figure}[!htb]
    \centering
    \begin{subfigure}[t]{0.245\linewidth}
        \includegraphics[width=\textwidth]{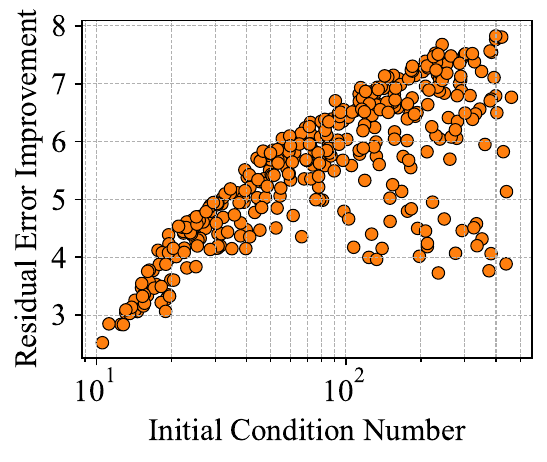}
        \caption{1 Subspace Iterations}
    \end{subfigure}
    \begin{subfigure}[t]{0.23\linewidth}
        \includegraphics[width=\textwidth]{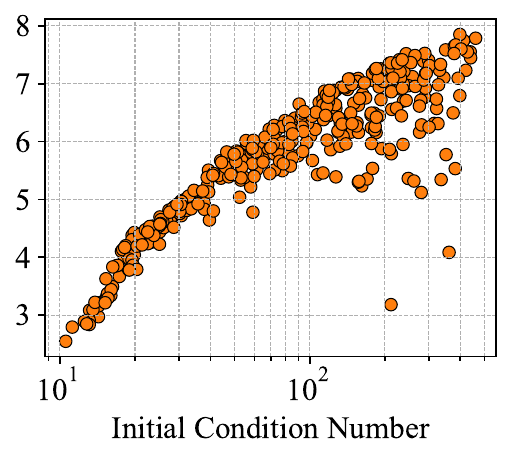}
        \caption{5 Subspace Iterations}
    \end{subfigure}
    \begin{subfigure}[t]{0.23\linewidth}
        \includegraphics[width=\textwidth]{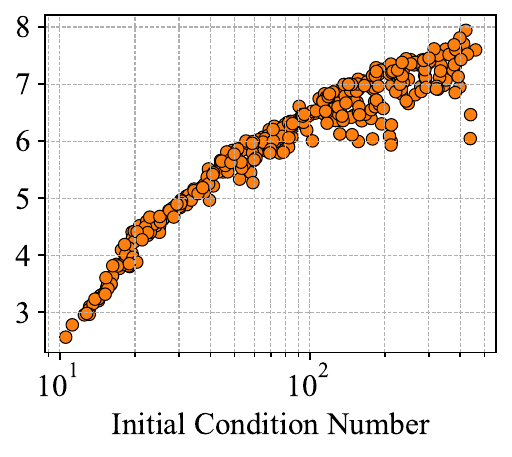}
        \caption{10 Subspace Iterations}
    \end{subfigure}
    \begin{subfigure}[t]{0.23\linewidth}
        \includegraphics[width=\textwidth]{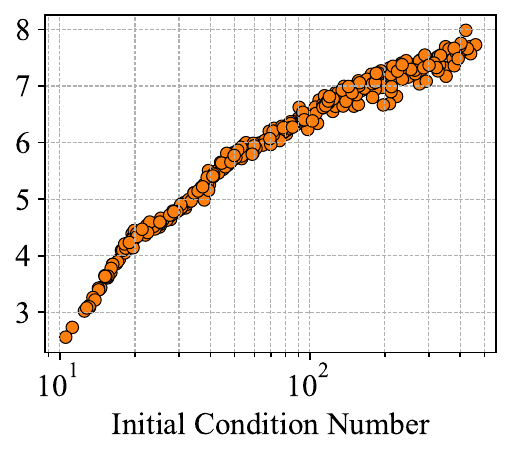}
        \caption{50 Subspace Iterations}
    \end{subfigure}

    \caption{Residual improvement of the algorithm generated by the neural network that approximates the inverse square root for synthetic matrices with different initial condition numbers, given different spectral probe quality. The input spectral probe is obtained using subspace iteration; more iterations result in a higher-quality spectral probe.}
    \label{fig:synthetic_matrix_func_niter} 
\end{figure}

To comprehensively analyze the performance of algorithms constructed by the \framework framework, we use synthetic diagonal matrices $\mathbf{X} = \text{diag}(\boldsymbol{\lambda})$ with different dimensions and condition number, and we evaluate the worst-case approximation error of the learned algorithm compared to Neumann series approximation of inverse square root, defined in~\cref{eq:obj_matfunc_diag}. We measure the ratio of errors in logarithms.  

As shown in Figure~\ref{fig:synthetic_matrix_func_niter}, we can see that the learned algorithm consistently improves the residual compared to the Neumann series approximation. Furthermore, we observe that the residual improvement increases as the initial condition number of the operator increases, and that using spectral probe with better quality results in the generation of more robust and effective approximation algorithms.

\section{Evaluating Properties of Learned Algorithms}\label{appendix:eval_properties}

Here, we examine whether the learned polynomial preconditioning algorithms have properties analogous to Chebyshev polynomials. In approximation theory, the Chebyshev polynomial $T_n(t)$ of degree $d$ is known to be the unique solution to the \textit{minimax} problem. Specifically, among all monic polynomials (polynomials with leading coefficient 1), the scaled Chebyshev polynomial $\tilde{T}_d(t) = 2^{-(n-1)} T_d(t)$ minimizes the infinity norm on the interval $[-1, 1]$.  

This optimality implies the following two testable conditions for any candidate polynomial $P(\lambda)$ on an arbitrary interval $\mathcal{D}$:
\begin{enumerate}
\item 
\textbf{Equioscillation:} The polynomial must oscillate between two bounds $\pm L$, achieving its maximum magnitude at $d+1$ distinct points (the Chebyshev alternation theorem).
\item 
\textbf{Minimal Norm Bound:} For a polynomial with leading coefficient $K$, the maximum absolute value on the optimal interval is lower-bounded by $|K| \cdot 2^{-(d-1)}$.
\end{enumerate}
To evaluate whether our learned algorithms satisfy these properties, without prior knowledge of the spectral bounds, we formulate a window-discovery optimization problem. We seek an affine transformation $\theta(t) = c_1 t + c_2$ that maps the canonical domain $t \in [-1, 1]$ to the optimal interval of the learned polynomial. We define the verification objective as minimizing the maximum norm on the interval $[-1, 1]$:
\begin{equation}
\mathcal{L}(c_1, c_2) = \frac{\max_{t \in [-1, 1]} |P(c_1 t + c_2)|}{A(c_1)} ,
\end{equation}
where $A(c_1)$ represents the leading coefficient of the transformed polynomial. Given the recurrence relations used to construct $P(\lambda)$, the leading coefficient scales as $A(c_1) = a_d c_1^d$, where $a_d$ is the leading coefficient of $P(\lambda)$ derived from the product of the recurrence scalars $\beta_k$. If the learned algorithm produces a true Chebyshev polynomial, there exists a unique window parameter $(c_1^*, c_2^*)$ such that $\mathcal{L} \to 0$. Conversely, $\mathcal{L} \gg 0$ indicates that the polynomial is suboptimal (i.e., its maximum value is larger than the theoretical minimum for its degree and leading coefficient). We solve this non-convex optimization problem using a grid search initialization followed by a Nelder-Mead simplex search.

\begin{center}
\vspace{-0.5em}
\captionof{table}{Minimax optimality gap, where lower is more Chebyshev-like. ``Random'' uses randomly generated polynomial coefficients.}
\label{tab:eval_cheb_property}
\vspace{0.25em}
{\scriptsize
\setlength{\tabcolsep}{5pt}
\renewcommand{\arraystretch}{0.9}
\begin{tabular}{lcccccc}
\toprule
\textbf{Matrix}
& \texttt{thermal1}
& \texttt{thermal2}
& \texttt{SiO2}
& \texttt{CO}
& Random
& Optimal \\
\midrule
\textbf{Minimax Optimality Gap (↓)}
& $9.390{\times}10^{-7}$
& $9.331{\times}10^{-7}$
& $8.656{\times}10^{-7}$
& $8.670{\times}10^{-7}$
& $0.954$
& $2.850{\times}10^{-7}$ \\
\bottomrule
\end{tabular}
}

\end{center}

%% file: example_paper.bib
@article{bergamaschi2021parallel,
  title={Parallel Newton--Chebyshev polynomial preconditioners for the conjugate gradient method},
  author={Bergamaschi, Luca and Martinez Calomardo, Angeles},
  journal={Computational and Mathematical Methods},
  volume={3},
  number={6},
  pages={e1153},
  year={2021},
  publisher={Wiley Online Library}
}

@article{dunham1982choice,
  title={Choice of basis for Chebyshev approximation},
  author={Dunham, Charles B},
  journal={ACM Transactions on Mathematical Software (TOMS)},
  volume={8},
  number={1},
  pages={21--25},
  year={1982},
  publisher={ACM New York, NY, USA}
}

@article{trifonov2026learning,
  title={Learning from linear algebra: A graph neural network approach to preconditioner design for conjugate gradient solvers},
  author={Trifonov, Vladislav and Rudikov, Alexander and Iliev, Oleg and Laevsky, Yuri M and Oseledets, Ivan and Muravleva, Ekaterina},
  journal={Computational Methods in Applied Mathematics},
  number={0},
  year={2026},
  publisher={De Gruyter}
}

@article{clenshaw1955summation,
  author  = {Clenshaw, C. W.},
  title   = {A Note on the Summation of Chebyshev Series},
  journal = {Mathematical Tables and Other Aids to Computation},
  volume  = {9},
  number  = {51},
  pages   = {118--120},
  year    = {1955},
  doi     = {10.1090/S0025-5718-1955-0071856-0}
}

@article{manteuffel1977tchebychev,
  title={The Tchebychev iteration for nonsymmetric linear systems},
  author={Manteuffel, Thomas A},
  journal={Numerische Mathematik},
  volume={28},
  number={3},
  pages={307--327},
  year={1977},
  publisher={Springer}
}

@article{rudikov2024neural,
  title={Neural operators meet conjugate gradients: The FCG-NO method for efficient PDE solving},
  author={Rudikov, Alexander and Fanaskov, Vladimir and Muravleva, Ekaterina and Laevsky, Yuri M and Oseledets, Ivan},
  journal={arXiv preprint arXiv:2402.05598},
  year={2024}
}

@article{prism_26_tr,
  title={{PRISM}: Distribution-free Adaptive Computation of Matrix Functions for Accelerating Neural Network Training},
  author={S. Yang and Z. Wang and O. Balabanov and N. B. Erichson and M. W. Mahoney},
  journal={arXiv preprint arXiv:2601.22137},
  year={2026}
}

@inproceedings{
luo2024neural,
title={Neural {K}rylov Iteration for Accelerating Linear System Solving},
author={Jian Luo and Jie Wang and Hong Wang and huanshuo dong and Zijie Geng and Hanzhu Chen and Yufei Kuang},
booktitle={The Thirty-eighth Annual Conference on Neural Information Processing Systems},
year={2024},
}

@book{saad2011eig,
  author    = {Saad, Yousef},
  title     = {Numerical Methods for Large Eigenvalue Problems},
  edition   = {Second},
  publisher = {Society for Industrial and Applied Mathematics},
  year      = {2011},
}

@article{saad1985practical,
  title={Practical use of polynomial preconditionings for the conjugate gradient method},
  author={Saad, Youcef},
  journal={SIAM Journal on Scientific and Statistical Computing},
  volume={6},
  number={4},
  pages={865--881},
  year={1985},
  publisher={SIAM}
}

@book{barrett1994templates,
  title={Templates for the solution of linear systems: building blocks for iterative methods},
  author={Barrett, Richard and Berry, Michael and Chan, Tony F and Demmel, James and Donato, June and Dongarra, Jack and Eijkhout, Victor and Pozo, Roldan and Romine, Charles and Van der Vorst, Henk},
  year={1994},
  publisher={SIAM}
}

@article{zhou2007chebdavidson,
  author  = {Zhou, Yunkai and Saad, Yousef},
  title   = {A {C}hebyshev--{D}avidson Algorithm for Large Symmetric Eigenproblems},
  journal = {SIAM Journal on Matrix Analysis and Applications},
  volume  = {29},
  number  = {3},
  pages   = {954--971},
  year    = {2007},
}

@article{banerjee2016chefsidg,
  author  = {Banerjee, Amartya S. and Lin, Lin and Hu, Wei and Yang, Chao and Pask, John E.},
  title   = {{C}hebyshev Polynomial Filtered Subspace Iteration in the Discontinuous {G}alerkin Method for Large-Scale Electronic Structure Calculations},
  journal = {The Journal of Chemical Physics},
  volume  = {145},
  number  = {15},
  pages   = {154101},
  year    = {2016},
}

@article{petersen2019deep,
  title={Deep symbolic regression: Recovering mathematical expressions from data via risk-seeking policy gradients},
  author={Petersen, Brenden K and Landajuela, Mikel and Mundhenk, T Nathan and Santiago, Claudio P and Kim, Soo K and Kim, Joanne T},
  journal={arXiv preprint arXiv:1912.04871},
  year={2019}
}

@article{hayes2025deep,
  title={Deep Symbolic Optimization: Reinforcement Learning for Symbolic Mathematics},
  author={Hayes, Conor F and Da Silva, Felipe Leno and Yang, Jiachen and Mundhenk, T Nathan and Lee, Chak Shing and Pettit, Jacob F and Santiago, Claudio and Kim, Sookyung and Kim, Joanne T and Solis, Ignacio Aravena and others},
  journal={arXiv preprint arXiv:2505.10762},
  year={2025}
}

@article{ashby1991minimax,
  title={Minimax polynomial preconditioning for {Hermitian} linear systems},
  author={Ashby, Steven F},
  journal={SIAM Journal on Matrix Analysis and Applications},
  volume={12},
  number={4},
  pages={766--789},
  year={1991},
  publisher={SIAM}
}

@book{young1971,
  title={Iterative solution of large linear systems},
  author={Young, David M},
  year={2014},
  publisher={Elsevier}
}

@book{varga2000mia,
  author    = {Varga, Richard S.},
  title     = {Matrix Iterative Analysis},
  edition   = {Second revised and expanded},
  series    = {Springer Series in Computational Mathematics},
  volume    = {27},
  publisher = {Springer},
  year      = {2000},
}

@book{axelsson1994iter,
  title={Iterative solution methods},
  author={Axelsson, Owe},
  year={1996},
  publisher={Cambridge university press}
}

@book{saad2003iter,
  author    = {Saad, Yousef},
  title     = {Iterative Methods for Sparse Linear Systems},
  edition   = {Second},
  publisher = {Society for Industrial and Applied Mathematics},
  year      = {2003},
}

@article{hestenesstiefel1952cg,
  title={Methods of conjugate gradients for solving linear systems},
  author={Hestenes, Magnus R and Stiefel, Eduard and others},
  journal={Journal of research of the National Bureau of Standards},
  volume={49},
  number={6},
  pages={409--436},
  year={1952}
}

@article{saadschultz1986gmres,
  author  = {Saad, Youcef and Schultz, Martin H.},
  title   = {{GMRES}: A Generalized Minimal Residual Algorithm for Solving Nonsymmetric Linear Systems},
  journal = {SIAM Journal on Scientific and Statistical Computing},
  volume  = {7},
  number  = {3},
  pages   = {856--869},
  year    = {1986},
}

@book{greenbaum1997,
  author    = {Greenbaum, Anne},
  title     = {Iterative Methods for Solving Linear Systems},
  publisher = {Society for Industrial and Applied Mathematics},
  year      = {1997},
}

@book{liesenstrakos2013,
  title={Krylov subspace methods: principles and analysis},
  author={Liesen, J{\"o}rg and Strakos, Zdenek},
  year={2013},
  publisher={Numerical Mathematics and Scientific Computation}
}

@article{johnson1983poly,
  author  = {Johnson, Olin G. and Micchelli, Charles A. and Paul, George},
  title   = {Polynomial Preconditioners for Conjugate Gradient Calculations},
  journal = {SIAM Journal on Numerical Analysis},
  volume  = {20},
  number  = {2},
  pages   = {362--376},
  year    = {1983},
}

@article{ashby1992ls,
  title={A comparison of adaptive {C}hebyshev and least squares polynomial preconditioning for {H}ermitian positive definite linear systems},
  author={Ashby, Steven F and Manteuffel, Thomas A and Otto, James S},
  journal={SIAM Journal on Scientific and Statistical Computing},
  volume={13},
  number={1},
  pages={1--29},
  year={1992},
  publisher={SIAM}
}

@article{benzi2002survey,
  author  = {Benzi, Michele},
  title   = {Preconditioning Techniques for Large Linear Systems: a Survey},
  journal = {Journal of Computational Physics},
  volume  = {182},
  number  = {2},
  pages   = {418--477},
  year    = {2002},
}

@book{briggs2000multigrid,
  author    = {Briggs, William L. and Henson, Van Emden and McCormick, Steve F.},
  title     = {A Multigrid Tutorial},
  edition   = {Second},
  publisher = {Society for Industrial and Applied Mathematics},
  year      = {2000},
}

@book{trottenberg2000multigrid,
  author    = {Trottenberg, Ulrich and Oosterlee, Cornelius W. and Sch{\"u}ller, Anton},
  title     = {Multigrid},
  publisher = {Academic Press},
  year      = {2000},
}

@book{higham2008fm,
  author    = {Higham, Nicholas J.},
  title     = {Functions of Matrices: Theory and Computation},
  publisher = {Society for Industrial and Applied Mathematics},
  year      = {2008},
}

@article{druskin1998extended,
  author  = {Druskin, Vladimir and Knizhnerman, Leonid},
  title   = {Extended {K}rylov Subspaces: Approximation of the Matrix Square Root and Related Functions},
  journal = {SIAM Journal on Matrix Analysis and Applications},
  volume  = {19},
  number  = {3},
  pages   = {755--771},
  year    = {1998},
}

@article{guettel2013rationalkrylov,
  author  = {G{\"u}ttel, Stefan},
  title   = {Rational {K}rylov Approximation of Matrix Functions: Numerical Methods and Optimal Pole Selection},
  journal = {GAMM-Mitteilungen},
  volume  = {36},
  number  = {1},
  pages   = {8--31},
  year    = {2013},
}

@article{nakatsukasa2018aaa,
  author  = {Nakatsukasa, Yuji and S{\`e}te, Olivier and Trefethen, Lloyd N.},
  title   = {The {AAA} Algorithm for Rational Approximation},
  journal = {SIAM Journal on Scientific Computing},
  volume  = {40},
  number  = {3},
  pages   = {A1494--A1522},
  year    = {2018},
}

@article{koza_genetic_1994,
	title = {Genetic programming as a means for programming computers by natural selection},
	volume = {4},
	number = {2},
	journal = {Statistics and Computing},
	author = {Koza, John R.},
	month = jun,
	year = {1994},
	pages = {87--112},
}

@article{
doi:10.1126/science.1165893,
author = {Michael Schmidt  and Hod Lipson },
title = {Distilling Free-Form Natural Laws from Experimental Data},
journal = {Science},
volume = {324},
number = {5923},
pages = {81-85},
year = {2009}}

@article{kim2020integration,
  title={Integration of neural network-based symbolic regression in deep learning for scientific discovery},
  author={Kim, Samuel and Lu, Peter Y and Mukherjee, Srijon and Gilbert, Michael and Jing, Li and {\v{C}}eperi{\'c}, Vladimir and Solja{\v{c}}i{\'c}, Marin},
  journal={IEEE transactions on neural networks and learning systems},
  volume={32},
  number={9},
  pages={4166--4177},
  year={2020},
  publisher={IEEE}
}

@article{mundhenk2021symbolic,
  title={Symbolic regression via deep reinforcement learning enhanced genetic programming seeding},
  author={Mundhenk, Terrell and Landajuela, Mikel and Glatt, Ruben and Santiago, Claudio P and Petersen, Brenden K and others},
  journal={Advances in Neural Information Processing Systems},
  volume={34},
  pages={24912--24923},
  year={2021}
}

@article{
doi:10.1126/sciadv.aay2631,
author = {Silviu-Marian Udrescu  and Max Tegmark },
title = {{AI Feynman}: A physics-inspired method for symbolic regression},
journal = {Science Advances},
volume = {6},
number = {16},
pages = {eaay2631},
year = {2020}}

@article{lample2019deep,
  title={Deep learning for symbolic mathematics},
  author={Lample, Guillaume and Charton, Fran{\c{c}}ois},
  journal={arXiv preprint arXiv:1912.01412},
  year={2019}
}

@article{fawzi_discovering_2022,
	title = {Discovering faster matrix multiplication algorithms with reinforcement learning},
	volume = {610},
	number = {7930},
	journal = {Nature},
	author = {Fawzi, Alhussein and Balog, Matej and Huang, Aja and Hubert, Thomas and Romera-Paredes, Bernardino and Barekatain, Mohammadamin and Novikov, Alexander and R. Ruiz, Francisco J. and Schrittwieser, Julian and Swirszcz, Grzegorz and Silver, David and Hassabis, Demis and Kohli, Pushmeet},
	month = oct,
	year = {2022},
	pages = {47--53},
}

@article{oh2020discovering,
  title={Discovering reinforcement learning algorithms},
  author={Oh, Junhyuk and Hessel, Matteo and Czarnecki, Wojciech M and Xu, Zhongwen and van Hasselt, Hado P and Singh, Satinder and Silver, David},
  journal={Advances in Neural Information Processing Systems},
  volume={33},
  pages={1060--1070},
  year={2020}
}

@article{mankowitz2023alphadev,
  author  = {Mankowitz, Daniel J. and Michi, Andrea and Zhernov, Anton and Gelmi, Marco and Selvi, Marco and Paduraru, Cosmin and Leurent, Edouard and Iqbal, Shariq and Lespiau, Jean-Baptiste and Ahern, Alex and K{\"o}ppe, Thomas and Millikin, Kevin and others},
  title   = {Faster sorting algorithms discovered using deep reinforcement learning},
  journal = {Nature},
  year    = {2023},
  volume  = {618},
  number  = {7964},
  pages   = {257--263},
}

@inproceedings{real2020automlzero,
  author    = {Real, Esteban and Liang, Chen and So, David and Le, Quoc V.},
  title     = {{AutoML-Zero}: Evolving Machine Learning Algorithms From Scratch},
  booktitle = {Proceedings of the 37th International Conference on Machine Learning (ICML)},
  series    = {Proceedings of Machine Learning Research},
  volume    = {119},
  pages     = {8007--8019},
  year      = {2020},
  publisher = {PMLR},
}

@inproceedings{ellis2021dreamcoder,
  author    = {Ellis, Kevin and Wong, Catherine and Nye, Maxwell and Sabl{\'e}-Meyer, Mathias and Morales, Lucas and Hewitt, Luke and Cary, Luc and Solar-Lezama, Armando and Tenenbaum, Joshua B.},
  title     = {DreamCoder: Bootstrapping Inductive Program Synthesis with Wake-Sleep Library Learning},
  booktitle = {Proceedings of the 42nd ACM SIGPLAN International Conference on Programming Language Design and Implementation (PLDI)},
  year      = {2021},
  publisher = {Association for Computing Machinery},
}

@article{oh_discovering_2025,
	title = {Discovering state-of-the-art reinforcement learning algorithms},
	volume = {648},
	number = {8093},
	journal = {Nature},
	author = {Oh, Junhyuk and Farquhar, Gregory and Kemaev, Iurii and Calian, Dan A. and Hessel, Matteo and Zintgraf, Luisa and Singh, Satinder and van Hasselt, Hado and Silver, David},
	month = dec,
	year = {2025},
	pages = {312--319},
}

@article{romera-paredes_mathematical_2024,
	title = {Mathematical discoveries from program search with large language models},
	volume = {625},
	number = {7995},
	journal = {Nature},
	author = {Romera-Paredes, Bernardino and Barekatain, Mohammadamin and Novikov, Alexander and Balog, Matej and Kumar, M. Pawan and Dupont, Emilien and Ruiz, Francisco J. R. and Ellenberg, Jordan S. and Wang, Pengming and Fawzi, Omar and Kohli, Pushmeet and Fawzi, Alhussein},
	month = jan,
	year = {2024},
	pages = {468--475},
}

@article{novikov2025alphaevolve,
  title={AlphaEvolve: A coding agent for scientific and algorithmic discovery},
  author={Novikov, Alexander and V{\~u}, Ng{\^a}n and Eisenberger, Marvin and Dupont, Emilien and Huang, Po-Sen and Wagner, Adam Zsolt and Shirobokov, Sergey and Kozlovskii, Borislav and Ruiz, Francisco JR and Mehrabian, Abbas and others},
  journal={arXiv preprint arXiv:2506.13131},
  year={2025}
}

@article{chervonyi2025gold,
  title={Gold-medalist performance in solving olympiad geometry with alphageometry2},
  author={Chervonyi, Yuri and Trinh, Trieu H and Ol{\v{s}}{\'a}k, Miroslav and Yang, Xiaomeng and Nguyen, Hoang H and Menegali, Marcelo and Jung, Junehyuk and Kim, Junsu and Verma, Vikas and Le, Quoc V and others},
  journal={Journal of Machine Learning Research},
  volume={26},
  number={241},
  pages={1--39},
  year={2025}
}

@article{andrychowicz2016learning,
  title={Learning to learn by gradient descent by gradient descent},
  author={Andrychowicz, Marcin and Denil, Misha and Gomez, Sergio and Hoffman, Matthew W and Pfau, David and Schaul, Tom and Shillingford, Brendan and De Freitas, Nando},
  journal={Advances in neural information processing systems},
  volume={29},
  year={2016}
}

@inproceedings{ravi2017optimization,
  title={Optimization as a model for few-shot learning},
  author={Ravi, Sachin and Larochelle, Hugo},
  booktitle={International conference on learning representations},
  year={2017}
}

@inproceedings{wichrowska2017learned,
  title={Learned optimizers that scale and generalize},
  author={Wichrowska, Olga and Maheswaranathan, Niru and Hoffman, Matthew W and Colmenarejo, Sergio Gomez and Denil, Misha and Freitas, Nando and Sohl-Dickstein, Jascha},
  booktitle={International conference on machine learning},
  pages={3751--3760},
  year={2017},
  organization={PMLR}
}

@article{li2016learning,
  title={Learning through dialogue interactions by asking questions},
  author={Li, Jiwei and Miller, Alexander H and Chopra, Sumit and Ranzato, Marc'Aurelio and Weston, Jason},
  journal={arXiv preprint arXiv:1612.04936},
  year={2016}
}

@inproceedings{bello2017neural,
  title={Neural optimizer search with reinforcement learning},
  author={Bello, Irwan and Zoph, Barret and Vasudevan, Vijay and Le, Quoc V},
  booktitle={International Conference on Machine Learning},
  pages={459--468},
  year={2017},
  organization={PMLR}
}

@article{chen2023symbolic,
  title={Symbolic discovery of optimization algorithms},
  author={Chen, Xiangning and Liang, Chen and Huang, Da and Real, Esteban and Wang, Kaiyuan and Pham, Hieu and Dong, Xuanyi and Luong, Thang and Hsieh, Cho-Jui and Lu, Yifeng and others},
  journal={Advances in neural information processing systems},
  volume={36},
  pages={49205--49233},
  year={2023}
}

@inproceedings{li2023learning,
  title={Learning preconditioners for conjugate gradient {PDE} solvers},
  author={Li, Yichen and Chen, Peter Yichen and Du, Tao and Matusik, Wojciech},
  booktitle={International Conference on Machine Learning},
  pages={19425--19439},
  year={2023},
  organization={PMLR}
}

@article{hausner2023neural,
  title={Neural incomplete factorization: learning preconditioners for the conjugate gradient method},
  author={H{\"a}usner, Paul and {\"O}ktem, Ozan and Sj{\"o}lund, Jens},
  journal={arXiv preprint arXiv:2305.16368},
  year={2023}
}

@article{lerer2024multigrid,
  title={Multigrid-augmented deep learning preconditioners for the Helmholtz equation using compact implicit layers},
  author={Lerer, Bar and Ben-Yair, Ido and Treister, Eran},
  journal={SIAM Journal on Scientific Computing},
  volume={46},
  number={5},
  pages={S123--S144},
  year={2024},
  publisher={SIAM}
}

@inproceedings{kaneda2023deep,
  title={A deep conjugate direction method for iteratively solving linear systems},
  author={Kaneda, Ayano and Akar, Osman and Chen, Jingyu and Kala, Victoria Alicia Trevino and Hyde, David and Teran, Joseph},
  booktitle={International Conference on Machine Learning},
  pages={15720--15736},
  year={2023},
  organization={PMLR}
}

@article{takabe2022convergence,
  title={Convergence acceleration via {C}hebyshev step: Plausible interpretation of deep-unfolded gradient descent},
  author={Takabe, Satoshi and Wadayama, Tadashi},
  journal={IEICE Transactions on Fundamentals of Electronics, Communications and Computer Sciences},
  volume={105},
  number={8},
  pages={1110--1120},
  year={2022},
  publisher={The Institute of Electronics, Information and Communication Engineers}
}

@inproceedings{gupta2018shampoo,
  title={Shampoo: Preconditioned stochastic tensor optimization},
  author={Gupta, Vineet and Koren, Tomer and Singer, Yoram},
  booktitle={International Conference on Machine Learning},
  pages={1842--1850},
  year={2018},
  organization={PMLR}
}

@article{jordan2024muon,
  title={Muon: An optimizer for hidden layers in neural networks},
  author={Jordan, Keller and Jin, Yuchen and Boza, Vlado and You, Jiacheng and Cesista, Franz and Newhouse, Laker and Bernstein, Jeremy},
  journal={Cited on},
  pages={10},
  year={2024},
  url = {https://kellerjordan.github.io/posts/muon/}
}

@article{amsel2025polarexpress,
  title = {The {P}olar {E}xpress: Optimal Matrix Sign Methods and Their Application to the {M}uon Algorithm},
  author = {Noah Amsel and David Persson and Christopher Musco and Robert M. Gower},
  journal = {arXiv preprint arXiv:2505.16932},
  year = {2025},
  url = {https://arxiv.org/abs/2505.16932}
}

@article{grishina2025cans,
  title   = {Accelerating {N}ewton--{S}chulz Iteration for Orthogonalization via {C}hebyshev-type Polynomials},
  author  = {Ekaterina Grishina and Matvey Smirnov and Maxim Rakhuba},
  journal = {arXiv preprint arXiv:2506.10935},
  year    = {2025},
}

@article{ahn2025dion,
  title={Dion: Distributed orthonormalized updates},
  author={Ahn, Kwangjun and Xu, Byron and Abreu, Natalie and Fan, Ying and Magakyan, Gagik and Sharma, Pratyusha and Zhan, Zheng and Langford, John},
  journal={arXiv preprint arXiv:2504.05295},
  year={2025}
}

@inproceedings{song2021approximate,
 author = {Song, Yue and Sebe, Nicu and Wang, Wei },
  booktitle = {Proceedings of the IEEE/CVF International Conference on Computer Vision (ICCV)},
  title = {Why Approximate Matrix Square Root Outperforms Accurate SVD in Global Covariance Pooling?},
  year = {2021},
  publisher = {IEEE Computer Society}
}

@article{song2023fast,
  author={Song, Yue and Sebe, Nicu and Wang, Wei},
  journal={IEEE Transactions on Pattern Analysis and Machine Intelligence (TPAMI)}, 
  title={Fast Differentiable Matrix Square Root and Inverse Square Root}, 
  year={2023},
  volume={45},
  number={6},
  pages={7367-7380},
}

@article{li2017universal,
  title={Universal style transfer via feature transforms},
  author={Li, Yijun and Fang, Chen and Yang, Jimei and Wang, Zhaowen and Lu, Xin and Yang, Ming-Hsuan},
  journal={Advances in Neural Information Processing Systems (NeurIPS)},
  volume={30},
  year={2017}
}

@inproceedings{huang2018decorrelated,
  title={Decorrelated batch normalization},
  author={Huang, Lei and Yang, Dawei and Lang, Bo and Deng, Jia},
  booktitle={Proceedings of the IEEE Conference on Computer Vision and Pattern Recognition (CVPR)},
  pages={791--800},
  year={2018}
}

@article{polyak1964,
  author  = {Polyak, B. T.},
  title   = {Some Methods of Speeding Up the Convergence of Iteration Methods},
  journal = {USSR Computational Mathematics and Mathematical Physics},
  volume  = {4},
  number  = {5},
  pages   = {1--17},
  year    = {1964},
}

@article{nesterov1983,
  author  = {Nesterov, Yurii E.},
  title   = {A Method of Solving a Convex Programming Problem with Convergence Rate {$O(1/k^2)$}},
  journal = {Soviet Mathematics Doklady},
  volume  = {27},
  number  = {2},
  pages   = {372--376},
  year    = {1983}
}

@book{nesterov2004,
  author    = {Nesterov, Yurii},
  title     = {Introductory Lectures on Convex Optimization: A Basic Course},
  series    = {Applied Optimization},
  volume    = {87},
  publisher = {Springer},
  address   = {New York, NY},
  year      = {2004},
}

@article{DavisHu2011,
  author  = {Davis, Timothy A. and Hu, Yifan},
  title   = {The University of Florida Sparse Matrix Collection},
  journal = {ACM Transactions on Mathematical Software},
  volume  = {38},
  number  = {1},
  articleno = {1},
  pages   = {1:1--1:25},
  year    = {2011},
  doi     = {10.1145/2049662.2049663}
}

@article{Kolodziej2019,
  author  = {Kolodziej, Scott P. and Aznaveh, Mohsen and Bullock, Matthew and David, Jarrett and Davis, Timothy A. and Henderson, Matthew and Hu, Yifan and Sandstrom, Read},
  title   = {The SuiteSparse Matrix Collection Website Interface},
  journal = {Journal of Open Source Software},
  volume  = {4},
  number  = {35},
  pages   = {1244},
  year    = {2019},
  doi     = {10.21105/joss.01244}
}
